\documentclass[10pt,journal,compsoc]{IEEEtran}
\usepackage[compress]{cite}
\usepackage{graphicx}
\usepackage{array}
\usepackage[inline]{enumitem}
\usepackage[style=default]{caption}
\usepackage{subcaption}
\usepackage{multirow}
\usepackage[para,online,flushleft]{threeparttable}
\usepackage{booktabs} %
\usepackage{comment}
\usepackage{xstring} %

\usepackage{amsmath}
\usepackage{amssymb} %
\usepackage{etoolbox} %
\usepackage[colorlinks]{hyperref}
\usepackage[usenames,dvipsnames]{xcolor}

\usepackage{soul}
\setstcolor{blue}

\renewcommand{\figureautorefname}{Fig.} %
\renewcommand{\tableautorefname}{Tab.}%
\def\equationautorefname~#1\null{Eq.~(#1)\null} %
\renewcommand{\sectionautorefname}{Sect.} %
\renewcommand{\subsectionautorefname}{\sectionautorefname} %

\providetoggle{includesuppmat}
\providetoggle{showcomments}
\settoggle{includesuppmat}{true}
\settoggle{showcomments}{false}

\iftoggle{showcomments}{%
    \newcommand{\changed}[1]{\textcolor{blue}{#1}}
    \newcommand{\numberchanged}[1]{\textcolor{green}{#1}}
    \newcommand{\todo}[1]{\textcolor{blue}{\textbf{TODO:} #1}}
    
    \newcommand{\resolved}[3][]{\ifstrequal{#1}{resolved}{\textcolor{blue}{RESOLVED:}~\textbf{{\MakeUppercase #2:}}~{#3}}{\textbf{\MakeUppercase #2:}~#3}}
    \newcommand{\footcomment}[1]{{#1}}
    \newcommand{\michael}[2][]{\textcolor{ForestGreen}{\resolved[#1]{michael}{#2}}}
    \newcommand{\alina}[2][]{\textcolor{NavyBlue}{\resolved[#1]{alina}{#2}}}
    \newcommand{\vitto}[2][]{\textcolor{red}{\resolved[#1]{vitto}{#2}}}
    \newcommand{\jasper}[2][]{\textcolor{violet}{\resolved[#1]{jasper}{#2}}}
    \newcommand{\tm}[2][]{\textcolor{magenta}{\resolved[#1]{TM}{#2}}}
    
    \newcommand{\draft}[1]{\textcolor{red}{#1}}
}{%
    \newcommand{\changed}[1]{#1}
    \newcommand{\numberchanged}[1]{#1}
    \newcommand{\todo}[1]{}
    
    \newcommand{\resolved}[1]{}
    \newcommand{\footcomment}[1]{}
    \newcommand{\michael}[2][]{}
    \newcommand{\alina}[2][]{}
    \newcommand{\vitto}[2][]{}
    \newcommand{\jasper}[2][]{}
    \newcommand{\tm}[2][]{}
    
    \newcommand{\draft}[1]{}
}

\newcommand{\consumer}{consumer photos}
\newcommand{\driving}{driving}
\newcommand{\indoor}{indoor}
\newcommand{\underwater}{underwater}
\newcommand{\aerial}{aerial}
\newcommand{\virtual}{synthetic}

\newcommand{\para}[1]{
    \par\noindent\textbf{#1}
}
\newcommand{\ie}{\textit{i.e.} }
\newcommand{\eg}{\textit{e.g.} }

\newcommand{\wrt}{\textit{w.r.t.}}
\newcommand{\cf}{\textit{c.f.} }

\usepackage{xspace}
\newcommand{\refappendix}[1]{\hyperref[#1]{Appendix~\ref*{#1}}}

\newcommand{\thisappendix}{%
    \iftoggle{includesuppmat}{appendix}{supplementary material}\xspace
}

\def\ifUnDefinedCs#1{\expandafter\ifx\csname#1\endcsname\relax} 
\newcommand{\appendixlabel}[1]{%
    {\color{RubineRed}\IfStrEqCase{#1}{%
    {sec:appendix_networks}{Sect. A}%
    {sec:appendix_results}{Sect. C}%
    {sec:detailed_generalization_experiments}{Sect. B}%
    {tab:tt_segmentation_raw_within_fixed_crop}{Table A.1}%
    {tab:tt_segmentation_raw_within_resnet_full}{Table A.2a}%
    {tab:tt_segmentation_raw_within_resnet_ss}{Table A.2b}%
    {tab:tt_segmentation_raw_within_resnet_ss_chain}{Table A.4}%
    }[LABEL NOT DEFINED: $#1$]%
    }
}
\newcommand{\appendixref}[1]{\ifUnDefinedCs{r@#1}{\appendixlabel{#1}~(in supplementary material)}\else%
	\iftoggle{includesuppmat}{%
	\autoref{#1}~(see Appendix)%
	}{\autoref{#1}}%
\fi}

\newcommand{\appendixsecref}[1]{%
	\iftoggle{includesuppmat}{\renewcommand{\sectionautorefname}{Appendix}\autoref{#1}\renewcommand{\sectionautorefname}{Section}}{\autoref{#1}}%
}

\begin{document}
\title{Factors of Influence for Transfer Learning across Diverse Appearance Domains and Task Types}
\author{Thomas~Mensink*,
        Jasper~Uijlings*,
        Alina~Kuznetsova,
        Michael~Gygli,
        and~Vittorio~Ferrari%
    \IEEEcompsocitemizethanks{
        \IEEEcompsocthanksitem Google Research
        \IEEEcompsocthanksitem Primary contacts:  \href{mailto:mensink@google.com}{mensink@google.com} and \href{mailto:jrru@google.com}{jrru@google.com}.
    }
\thanks{* Equal contribution.}
\thanks{Manuscript submitted in March, 2021, revision submitted in September, 2021, and accepted in November 2021 for future publication in TPAMI. Copyright may be transferred without notice, after which this version may no longer be accessible.} %
}

\markboth{Factors of Influence for Transfer Learning}{Mensink \MakeLowercase{\textit{et al.}}: Factors of Influence for Transfer Learning}

\IEEEtitleabstractindextext{%
    \begin{abstract}
Transfer learning enables to re-use knowledge learned on a source task to help learning a target task.
A simple form of transfer learning is common in current state-of-the-art computer vision models, i.e. pre-training a model for image classification on the ILSVRC dataset, and then fine-tune on any target task.
However, previous systematic studies of transfer learning have been limited and the circumstances in which it is expected to work are not fully understood.
In this paper we carry out an extensive experimental exploration of transfer learning across vastly different image domains (consumer photos, autonomous driving, aerial imagery, underwater, indoor scenes, synthetic, close-ups) and task types (semantic segmentation, object detection, depth estimation, keypoint detection). Importantly, these are all complex, structured output tasks types relevant to modern computer vision applications.
In total we carry out \changed{over 2000 transfer learning experiments}, including many where the source and target come from different image domains, task types, or both.
We systematically analyze these experiments to understand the impact of image domain, task type, and dataset size on transfer learning performance. 
\changed{%
Our study leads to several insights and concrete recommendations:
(1) for most tasks there exists a source which significantly outperforms ILSVRC'12 pre-training;
(2) the image domain is the most important factor for achieving positive transfer;
(3) the source dataset should \emph{include} the image domain of the target dataset to achieve best results;
(4) at the same time, we observe only small negative effects when the image domain of the source task is much broader than that of the target;
(5) transfer across task types can be beneficial, but its success is heavily dependent on both the source and target task types.
}
\end{abstract}


}
\maketitle

\IEEEdisplaynontitleabstractindextext
\IEEEpeerreviewmaketitle

\IEEEraisesectionheading{\section{Introduction}\label{sec:introduction}}

\IEEEPARstart{T}{} ranfer learning is omnipresent in computer vision.
The common practice is transfer learning through ILSVRC'12 pre-training: train on the ILSVRC'12 image classification task~\cite{ilsvrc12}, copy the resulting weights to a target model, then fine-tune for the target task at hand. 
This strategy was shown to be effective on a wide variety of datasets and task types, including image classification~\cite{azizpour15pami,chu16eccv,huh16nips,kornblith19cvpr,gygli21aaai},
object detection~\cite{girshick15iccv}, semantic segmentation~\cite{shelhamer16pami}, human pose estimation~\cite{he2017iccv,zhou19arxiv,zhang19arxiv}, and depth estimation~\cite{fu18cvpr, chen193dv}.
Intuitively, the reason for this success is that the network learns a strong generic visual representation, providing a better starting point for learning a new task than training from scratch.

But can we do better than ILSVRC'12 pre-training?
And what factors make a source task good for transferring to a given target task?
Some previous works aim to demonstrate that a generic representation trained on a single large source dataset works well for a variety of classification tasks~\cite{zhai19arxiv,kolesnikov20eccv,mustafa21arxiv}.
Others instead try to automatically find a subset of the source dataset that transfers best to a given target~\cite{puigcerver20iclr,ngiam18arxiv,yan20cvpr,ge17cvpr}.
By virtue of dataset collection suites such as VTAB~\cite{zhai19arxiv} and VisDA~\cite{peng18cvprw}, recently several works in transfer learning and domain adaptation experiment with target datasets spanning a variety of image domains~\cite{zhai19arxiv,kolesnikov20eccv,dosovitskiy20arxiv,mustafa21arxiv,peng18cvprw,kundu20cvpr,zhao17arxiv}.
However, most previous work focuses solely on image classification~\cite{zhai19arxiv,puigcerver20iclr,ge17cvpr,dosovitskiy20arxiv,long15icml,saenko10eccv,kundu20cvpr} or a single structured prediction task~\cite{yan20cvpr,uijlings18cvpr,hoffman16arXiv,tsai18cvpr,lee21aaai}.

In this paper, we go beyond previous works by providing a large-scale exploration of transfer learning across a wide variety of image domains and task types.
In particular, we perform over \changed{2000 transfer learning experiments} across 20 datasets spanning seven diverse image domains
(consumer, driving, aerial, underwater, indoor, synthetic, close-ups)
and four task types
(semantic segmentation, object detection, depth estimation, keypoint detection).
In many of our experiments the source and target come from different image domains, task types, or both.
For example, we find that semantic segmentation on COCO~\cite{lin14eccv} is a good source for depth estimation on SUN RGB-D~\cite{song15cvpr} (\autoref{tab:few_shot_depth});
and even that keypoint detection on Stanford Dogs~\cite{biggs20eccv,khosla11fgvc} helps object detection on the Underwater Trash~\cite{underwatertrash20} dataset (\autoref{tab:full_target_detection}).
We then do a systematic meta-analysis of these experiments, relating transfer learning performance to three underlying factors of variation:
the difference in image domain between source and target tasks,
their difference in task type,
and the size of their training sets.
This yields new insights into when transfer learning brings benefits and which source works best for a given target.

At a high level, our main conclusions are:
(1) for most target tasks we are able to find sources that significantly outperform ILSVRC'12 pre-training;
(2) the image domain is the most important factor for achieving positive transfer;
(3) the source dataset should \emph{include} the image domain of the target dataset to achieve best results;
(4) at the same time, we observe \changed{only small negative effects} when the image domain of the source task is much broader than that of the target;
(5) transfer across task types can be beneficial, but its success is heavily dependent on both the source and target task types.

The rest of our paper is organized as follows: 
\autoref{sec:related_work} discusses related work. \autoref{sec:factors} discusses the three factors of variation. \autoref{sec:networks} details and validates the network architectures that we use. \autoref{sec:transfer_experiments} presents our transfer learning experiments. \autoref{sec:analysis} presents a detailed analysis of our results. \autoref{sec:conclusion} concludes our paper.

\newcommand{\myfigspacing}{-8pt}

\begin{figure*}[pt]
    {\setlength{\fboxrule}{4pt}
    \resizebox{\textwidth}{!}{
        \fcolorbox{BurntOrange}{BurntOrange}{
            \includegraphics{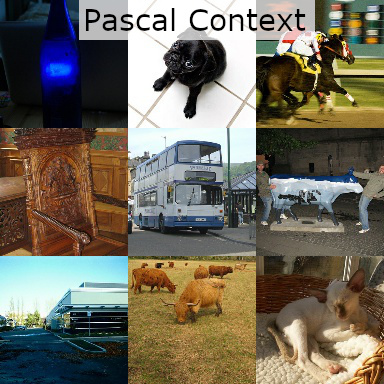}
        }
        \hspace{\myfigspacing}
        \fcolorbox{BurntOrange}{BurntOrange}{
            \includegraphics{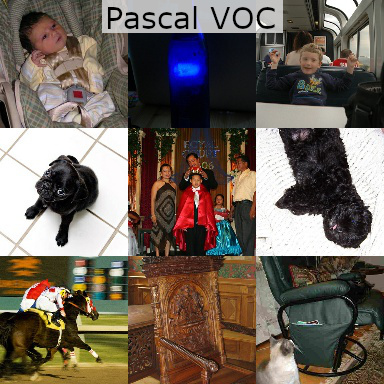}
        }
        \hspace{\myfigspacing}
        \fcolorbox{BurntOrange}{BurntOrange}{
            \includegraphics{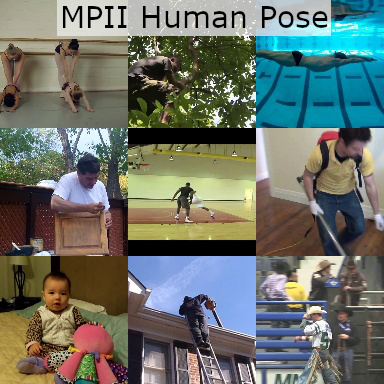}
        }
        \hspace{\myfigspacing}
        \fcolorbox{BurntOrange}{BurntOrange}{
            \includegraphics{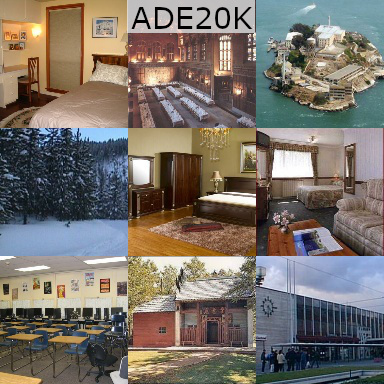}
        }
        \hspace{\myfigspacing}
        \fcolorbox{BurntOrange}{BurntOrange}{
            \includegraphics{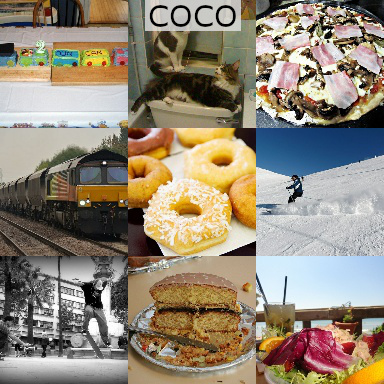}
        }
    }
    \\
    \resizebox{\textwidth}{!}{
        \fcolorbox{LimeGreen}{LimeGreen}{
            \includegraphics{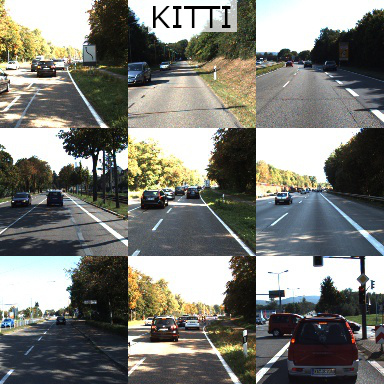}
        }
        \hspace{\myfigspacing}
        \fcolorbox{LimeGreen}{LimeGreen}{
            \includegraphics{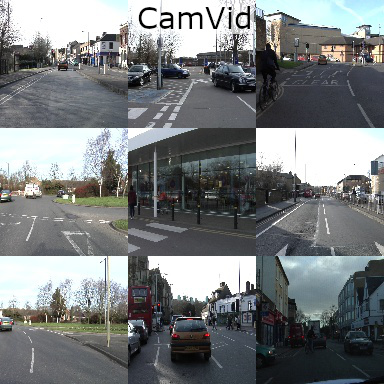}
        }
        \hspace{\myfigspacing}
        \fcolorbox{LimeGreen}{LimeGreen}{
            \includegraphics{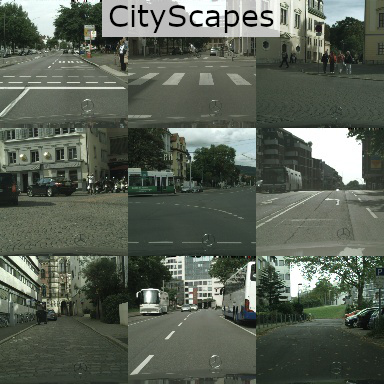}
        }
        \hspace{\myfigspacing}
        \fcolorbox{Maroon}{Maroon}{
            \includegraphics{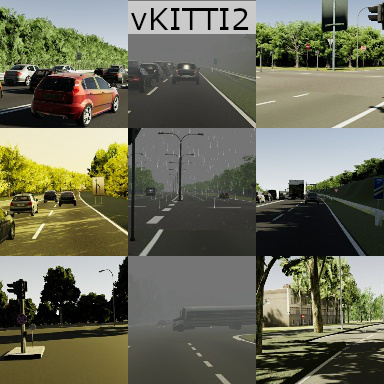}
        }
        \hspace{\myfigspacing}
        \fcolorbox{Maroon}{Maroon}{
            \includegraphics{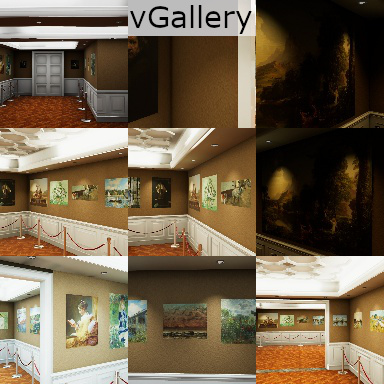}
        }
    }
    \\
    \resizebox{\textwidth}{!}{
        \fcolorbox{LimeGreen}{LimeGreen}{
            \includegraphics{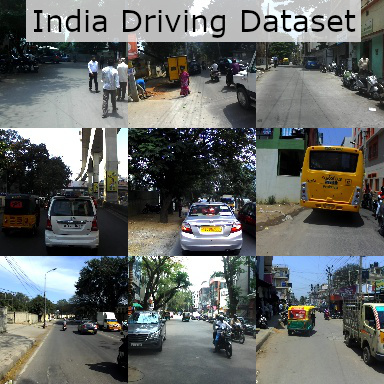}
        }
        \hspace{\myfigspacing}
        \fcolorbox{LimeGreen}{LimeGreen}{
            \includegraphics{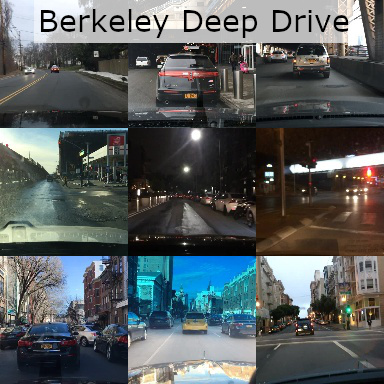}
        }
        \hspace{\myfigspacing}
        \fcolorbox{LimeGreen}{LimeGreen}{
            \includegraphics{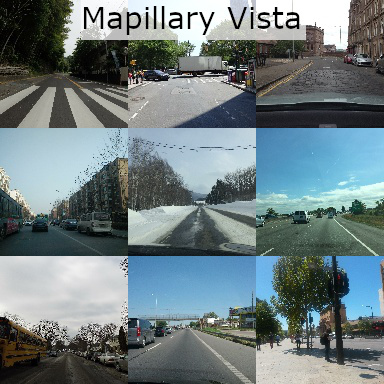}
        }
        \hspace{\myfigspacing}
        \fcolorbox{Violet}{Violet}{
            \includegraphics{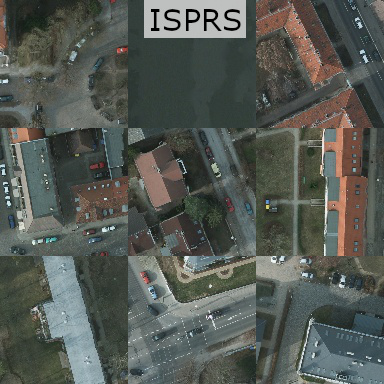}
        }
        \hspace{\myfigspacing}
        \fcolorbox{Violet}{Violet}{
            \includegraphics{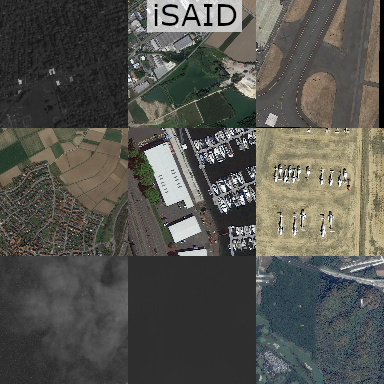}
        }
    }
    \\
    \resizebox{\textwidth}{!}{
        \fcolorbox{NavyBlue}{NavyBlue}{
            \includegraphics{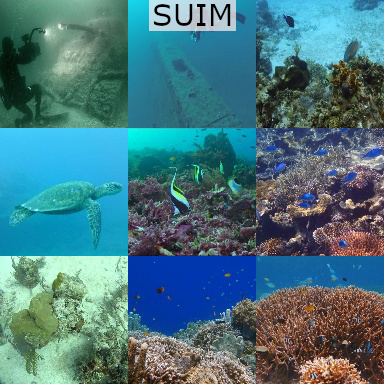}
        }
        \hspace{\myfigspacing}
        \fcolorbox{NavyBlue}{NavyBlue}{
            \includegraphics{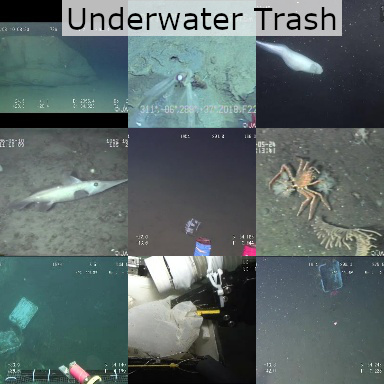}
        }
        \hspace{\myfigspacing}
        \fcolorbox{Yellow}{Yellow}{
            \includegraphics{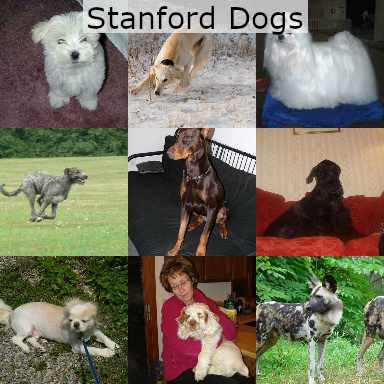}
        }
        \hspace{\myfigspacing}
        \fcolorbox{Magenta}{Magenta}{
            \includegraphics{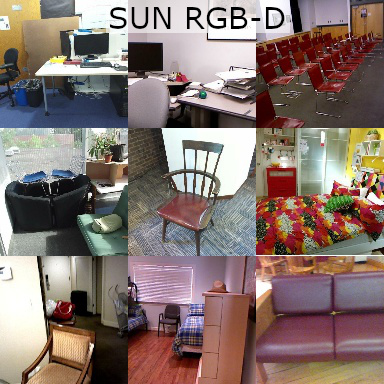}
        }
        \hspace{\myfigspacing}
        \fcolorbox{Magenta}{Magenta}{
            \includegraphics{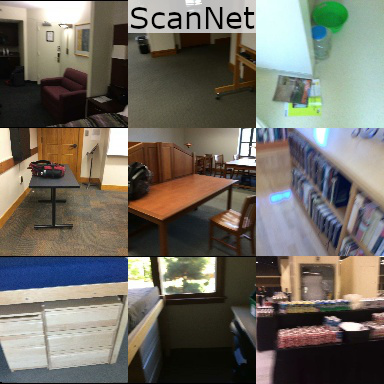}
        }
    }
    }
    \caption{We explore transfer learning across a wide variety of image domain and task types. We show here example images for the 20 datasets we consider, highlighting their visual diversity. We grouped them into manually defined image domains: 
    consumer in orange, driving in green, synthetic in red, aerial in purple, underwater in blue, close-ups in yellow, and indoor in magenta.
    }
    \label{fig:total_transfer_datasets_illustration}
\end{figure*}

\section[Related Work]{Related work}
\label{sec:related_work}

We review here related work on transfer learning in computer vision.
We take a rather broad definition of this term, and include several families of works that transfer knowledge from a task to another, even though they are not explicitly positioned as `transfer learning'.
We pay particular attention to their experimental setting and analyses, as this is the aspect most related to our work.

\para{Domain adaptation.}
This family of works adapts image classifiers from a source domain to a target domain, typically containing the same classes but appearing in different kind of images.
Some of the most common techniques are
minimizing the discrepancy in feature distributions
\cite{long15icml,ganin15icml,sun16eccvw,tzeng17cvpr,venkateswara17cvpr,saenko10eccv,zhao17arxiv},
embedding domain alignment layers into a network
\cite{carlucci17iccv,chang19cvpr,roy19cvpr},
directly transform images from the target to the source domain
\cite{bousmalis17cvpr,russo18cvpr,sankaranarayanan18cvpr}.
Earlier works consider relatively small domain shifts, e.g. where the source and target datasets are captured by different cameras~\cite{saenko10eccv}. %
Recent works explored larger domain shifts,
e.g. across clipart, web-shopping products, consumer photos and artist paintings~\cite{venkateswara17cvpr}, %
or across simple synthetic renderings and real consumer photos~\cite{peng18cvprw,kundu20cvpr},
and even fusing multiple datasets spanning several modalities such as hand-drawn sketches, synthetic renderings and consumer photos~\cite{zhao17arxiv}. %
Despite this substantial variation in domain, most works consider only object classification tasks, %
with relatively few papers tackling semantic segmentation~\cite{hoffman16arXiv,peng18cvprw,tsai18cvpr, sun19arxiv_a,yu21aaai,lee21aaai}.

\para{Few-shot learning.}
Works in this area aim to learn to classify new target classes from very few examples, typically 1-10, by transferring knowledge from source classes with many training samples. 
Typically the source and target classes come from the same dataset, hence there is no domain shift.
There are two main approaches: metric-based and optimization-based.
Optimization-based methods employ meta-learning~\cite{ravi16iclr,finn17icml,weng18blog,nichol18arxiv, raghu2020iclr}.
These methods train a model such that it can be later adapted to new classes with few update steps. MAML~\cite{finn17icml} is the most prominent example of this line of work. Despite its success,~\cite{raghu2020iclr} later showed that its performance largely stems from learning a generic feature embedding, rather than a model that can adapt to new data faster.
Metric-based methods~\cite{mensink13pami,koch15icmlw,vinyals16nips, snell17nips, sung18cvpr} aim at learning an embedding space which allows to classify examples of any class using a distance-based classifier,~\eg nearest neighbor~\cite{snell17nips}.
Overall, the community seems to be reaching a consensus~\cite{qi18cvpr_lowshotlearning,gidaris18cvpr, huang19arxiv,chen2019iclr,su20eccv, tian20eccv}: the key ingredient to high-performing few-shot classification is learning a general representation, rather than sophisticated algorithms for adapting to the new classes.
In line with these works, we study what representation is suitable for solving a target task.
In contrast with these works, our focus is on more complex structured prediction.

While most few-shot learning works focus on classification, there are a few exceptions, \eg for object detection~\cite{pmlr-v119-wang20j,10.1007/978-3-030-58517-4_27,Wang_2019_ICCV} and semantic segmentation~\cite{shaban17arxiv}. These works follow a similar setup with source and targets coming from the same dataset.
We also focus on structured prediction tasks, however, we
tackle more realistic scenarios, where a model is adapted to new datasets, appearance domains, and task types.

\para{Transfer learning.}
A common practice in computer vision is to start from a neural network trained for image classification on ILSVRC'12~\cite{ilsvrc12} as a generic source (\autoref{sec:introduction}).
Because of its success, this strategy is the starting point in all our experiments, and should be seen as the baseline to beat for any transfer learning method.

While many works simply apply this strategy as a practical `trick of the trade', several papers explore transfer learning in more depth, attempting to understand when it works, to find even better source datasets than ILSVRC'12, or to propose more sophisticated transfer techniques. 
The typical setting considers a very large source dataset:
besides ILSVRC'12~\cite{kolesnikov20eccv,ngiam18arxiv,ge17cvpr,dosovitskiy20arxiv,mustafa21arxiv},
also ImageNet21k with 9M images~\cite{puigcerver20iclr,zhai19arxiv,kolesnikov20eccv,dosovitskiy20arxiv,mustafa21arxiv},
Open Images with 1.7M images~\cite{yan20cvpr},
Places-205 with 2.5M images~\cite{ge17cvpr},
and even JFT-300M with 300M images~\cite{puigcerver20iclr,kolesnikov20eccv,ngiam18arxiv,dosovitskiy20arxiv,mustafa21arxiv}.
Several works~\cite{zhai19arxiv,dosovitskiy20arxiv,kolesnikov20eccv,mustafa21arxiv} report experiments on target datasets spanning different image domains, especially since the advent of the VTAB suite~\cite{zhai19arxiv} which assembles datasets captured using a standard camera, as well as remote sensing, medical, and synthetic ones.
Yet, each paper considers little variation in the source dataset, experimenting with 1-3 sources overall, and sometimes picking just one for each target dataset~\cite{ge17cvpr}.
More importantly, the vast majority of reported experiments are on image classification, for both source and target datasets. %
Note how VTAB downgrades some structured tasks to simpler versions that can be expressed as classification, e.g. predict the depth of the closest object to the camera, as opposed to a depth value for each object or pixel. %

In terms of method, most works follow the classical protocol of learning a feature representation from the source dataset, then fine-tune with a new task head on each target dataset in turn.
However, they differ in how they select source images for a given target.
Some aim to demonstrate that a single generic representation trained on the {\em whole} large source dataset works well for all target tasks~\cite{zhai19arxiv,kolesnikov20eccv}. 
Others instead try to automatically find a subset of the source dataset that transfers best to a given target~\cite{puigcerver20iclr,ngiam18arxiv,yan20cvpr,ge17cvpr}. The subsets correspond either to subtrees of the source class hierarchy~\cite{puigcerver20iclr,ngiam18arxiv}, or are derived by clustering source images by appearance~\cite{yan20cvpr}. After selecting a subset, they retrieve its corresponding pre-trained source model and fine-tune only that one on the target dataset.

Taskonomy~\cite{zamir18cvpr} explores a converse scenario.
They perform transfer learning across many task types, including several dense prediction ones, e.g. surface normal estimation and semantic segmentation.
However, there is only one image domain (indoor scenes) as all the tasks are defined on the same images (one dataset).
This convenient setting enables them to study which source task type helps the most which target task type, by exhaustively trying out all pairs. As an powerful outcome, they derive a taxonomy of task types, with directed edges weighted by how much a task helps another.

Finally, several other papers perform transfer learning experiments on more complex tasks than image classification as a way to validate their proposed method, but typically only for a few source-target dataset pairs, and only within the same task type
(e.g.
from Open Images~\cite{kuznetsova20ijcv} to CityScapes~\cite{cordts16cvpr} for instance segmentation~\cite{yan20cvpr}; %
and across pairs of ILSVRC'12~\cite{ilsvrc12}, COCO~\cite{lin14eccv}, Pascal VOC~\cite{pascal-challenge} for object detection~\cite{uijlings18cvpr}).

\para{Other related work.}
Finally we discuss two other directions of related work.
Experimenting over a collection of datasets is getting more common, \eg for \emph{robust vision} approaches~\cite{lambert20cvpr, rebuffi17nips,robustvisionchallenge}.
However, the general aim for robust methods is to learn a \emph{single} model which performs well across several datasets for one task type.

A difficulty for sequential learning of neural networks, is the tendency of \emph{catastrophic forgetting}~\cite{kirkpatrick17pnas, shmelkov17iccv, davidson20cvpr}.
In our transfer learning setup, the new target model might have forgotten the old source tasks. 
However, forgetting is not relevant in our study, since we are interested in performance on the target task only.
Moreover, we analyse in which conditions the target task can successfully re-use the knowledge from the source task.

\section[Factors of Influence]{Factors of Influence}
\label{sec:factors}

We study transfer learning at scale across three factors of influence:
the difference in image domain between source and target tasks (\autoref{sec:image_domain}),
their difference in task type (\autoref{sec:task_type}),
and the size of the source and target training sets (\autoref{sec:dataset_size}).
To study these factors of influence, we introduce a collection of 20 datasets in~\autoref{sec:total_transfer_datasets}, which we have chosen to cover these factors well.

\subsection{Transfer Learning through pre-training}

This paper explores transfer learning from a source task to a target task. We define a task as the combination of a task type (\eg object detection, semantic segmentation) and a dataset (\eg COCO). 
We follow the widespread practice of initializing the backbones of our networks with the weights obtained from image classification pre-training on ILSVRC'12~\cite{azizpour15pami,chu16eccv,huh16nips,kornblith19cvpr,girshick15iccv,shelhamer16pami,he2017iccv,zhou19arxiv}.
This leads to the following process:
\begin{enumerate}
\item we train a model on ILSVRC'12 classification.
\item we copy the weights of the backbone of the ILSVRC'12 classification model to the source model. We randomly initialize the head of the source model, which is specific to the task type.
\item we train on the source task.
\item we copy the weights of the backbone of the source model to the target model. Again, we randomly initialize its head.
\item we train on the target training set.
\item we evaluate on the target validation set.
\end{enumerate}

This protocol essentially defines a \emph{transfer chain}: 
\textbf{ILSVRC'12} $\rightarrow$ \textbf{source task} $\rightarrow$ \textbf{target task}. 
We compare these \emph{transfer chains} to the default practice:  \textbf{ILSVRC'12} $\rightarrow$ \textbf{target task}.
To ensure fair comparisons, we have one set of ILSVRC'12 pre-trained weights which are use throughout all experiments in this paper. Analogously, for each source task we create one set of pre-trained weights used throughout all our experiments.

\subsection{Image Domain}\label{sec:image_domain}
We want to study transfer learning across a wide range of different image domains.
For this we considered many publicly available datasets and manually selected the following domain types: \emph{\consumer{}}, \emph{\driving{}}, \emph{\indoor{}}, \emph{\aerial{}}, \emph{\underwater{}}, \emph{close-ups}, and \emph{\virtual{}}
We deliberately only use domains from RGB image sensors, and exclude imagery from other sensors, such as CT scans, multi-spectral imagery, or lidar,
because it is unclear how knowledge could be re-used when transferring across datasets acquired by different sensors.

To measure the difference between source and target domains, one way is to use these manually defined domain types. This results in a simple binary {\em same} or {\em different} measure.
We also use a continuous similarity metric obtained directly from the visual appearance of the source and target datasets, to offer a more fine grained metric.
In particular, we extract image features from a backbone based on our multi-source semantic segmentation model (\autoref{sec:transfer_experiments_setup}), where we attach a spatial average pooling layer to the backbone, resulting in an 720 dimensional image vector.
The domain difference is then the average distance of a target image to its closest source image:
\begin{equation}
    D(T|S) = \frac{1}{|T|}\sum_t \left( \min_s d(f_t, f_s) \right)
    \label{eq:domain_difference}
\end{equation}
where, $|\cdot|$ denotes the cardinality, $f$ denotes an image feature vector, and for $d(\cdot, \cdot)$ we use the Euclidean distance. 
From each dataset we sample 1000 images to compute $D(T|S)$. The same images are used irrespective of whether the dataset is used as target or source.
Due to the $\min$ operation in~\autoref{eq:domain_difference} this is an asymmetric measure, \ie $D(A|B) \neq D(B|A)$. 
This measure enables more fine-grained analysis than using our manually defined domains.

\subsection{Task type}\label{sec:task_type} 
In this study we focus on structured prediction tasks, which all involve some form of spatial localization.
Since training on these tasks yields spatially sensitive features, we hypothesise that these tasks could benefit each other.
We consider four task types:
\begin{enumerate}
    \item %
    \emph{semantic segmentation}: predict the class label for each pixel in the image.
    \item %
    \emph{object detection}: predict tight bounding boxes around objects and predict their class labels.
    \item %
    \emph{keypoint detection}: detect the image location of body joints or parts, for the human and dog classes in the datasets we consider.
    \item %
    \emph{depth estimation}: predict for each pixel the distance from the camera to the physical surface (from a single image).
\end{enumerate}
These four task types significantly vary in nature: semantic segmentation and depth estimation are pixel-wise prediction tasks, whereas keypoint detection requires identifying sparse (but related) points in an image, and object detection requires predicting bounding boxes.
Hence, while these tasks are all about spatial localisation, their different nature makes it interesting to study the influence of transferring from one task type to another.

\subsection{Dataset Size}\label{sec:dataset_size}

The size of the target training set is important (\eg~\cite{kolesnikov20eccv}).
When a target dataset is very large, the effect of transfer learning is likely to be minimal: all the required visual knowledge can be gathered directly from this target dataset. 
Therefore we consider two transfer learning settings, one using only a small number of target images and one using the full target dataset.
We believe that using a small target training set is most relevant in practice, given that we often want to train strong models from a small set of annotated images.

We also perform some experiments varying the size of the source training set.
A source model trained on a larger dataset is likely to be more beneficial for transfer learning~\cite{mahajan18eccv,sun17iccv,huh16nips}.
Hence, in some of our experiments we limit all source sets to have a maximum number of images, sampled uniformly from the dataset. This allows to study the influence of source domain versus source size.

\subsection{Dataset collection}
\label{sec:total_transfer_datasets}
\newcommand{\present}{$\checkmark$}
\newcommand{\derived}{$\rightsquigarrow$}
\newcommand{\nota}{-}  %
\newcommand{\maybe}{\nota}  %
\newcolumntype{H}{>{\setbox0=\hbox\bgroup}c<{\egroup}@{}}
\newcommand{\scalefactor}{1}

\renewcommand{\TPTtagStyle}[1]{{\color{red}\textbf{#1}}} 

\begin{table*}[pt]
    \centering
    \begin{threeparttable}[t]
    \resizebox{\scalefactor\textwidth}{!}{
        \begin{tabular}{lllcccccccl}\toprule
        Nr  & Name                      & Reference             & Domain                & Semantic  & Detection & Keypoints & Depth     & \# Train  & \# Val& \\\midrule
        1   & Pascal Context            &\cite{mottaghi14cvpr}  & Consumer              & 60        & \nota     & \nota     & \nota     & 5K        & 5K    & \\
        2   & Pascal VOC                &\cite{pascal-voc-2012} & Consumer              & 22        & 20        & \nota     & \nota     & 10K       & 1449  & \tnote{a}\\
        3   & MPII Human Pose           &\cite{andriluka14cvpr} & Consumer              & \nota     & \nota     & 16        & \nota     & 17K       & 7K    & \\
        4   & ADE20K                    &\cite{zhou17cvpr}      & Consumer              & 150       & \nota     & \nota     & \nota     & 20K       & 2K    & \\
        5   & COCO Panoptic             &\cite{caesar18cvpr,lin14eccv,kirillov19cvpr}       & Consumer              & 134       & 80        & 17        & \nota     & 118K      & 5K    & \\\cmidrule(r){2-10}        
        6   & KITTI                     &\cite{alhaija18ijcv}   & Driving               & 30        & \maybe    & \nota     & \maybe    & 150       & 50    & \\
        7   & CamVid                    &\cite{brostow09prl}    & Driving               & 23        & \nota     & \nota     & \nota     & 367       & 101   & \\        
        8   & CityScapes                &\cite{cordts16cvpr}    & Driving               & 33        & \nota     & \nota     & \nota     & 3K        & 500   & \\        
        9   & India Driving Dataset (IDD)    &\cite{idd2019}         & Driving               & 35        & \nota     & \nota     & \nota     & 7K        & 981   & \\                        
        10   & Berkeley Deep Drive (BDD)      &\cite{bdd100k}         & Driving               & 20        & 10        & \nota     & \nota     & 7K        & 3K    & \tnote{b}\\ 
        11  & Mapillary Vista Dataset           &\cite{mvd2017}         & Driving               & 66        & \nota     & \nota     & \nota     & 18K       & 2K    & \\
        \cmidrule(r){2-10}
        12  & ISPRS                     &\cite{isprs14dataset}  & Aerial                & 6         & \nota     & \nota     & \maybe    & 4K        & 437   & \tnote{c}\\
        13  & iSAID                     &\cite{waqas2019isaid, xia18cvpr}   &Aerial     & 16        & \maybe 	& \nota     & \nota     & 27K       & 9K    & \tnote{c}\\
        \cmidrule(r){2-10}
        14  & SUN RGB-D                  &\cite{song15cvpr}      & Indoor                & 37        & \nota     & \nota     & \present  & 5K        & 425   & \tnote{d} \\
        15  & ScanNet                  &\cite{dai17cvpr}       & Indoor                & 41        & \nota     & \nota     & \nota     & 19K       & 5K    & \\
        \cmidrule(r){2-10} %
        16  & SUIM                      &\cite{islam2020suim}   & Underwater            & 8         & \nota     & \nota     & \nota     & 1525      & 110 \\
        17  & Underwater Trash (UWT)    &\cite{underwatertrash20}& Underwater           & \nota     & 3         & \nota     & \nota     & 6K        & 1144\\        
        \cmidrule(r){2-10}
        18  & Stanford Dogs             &\cite{biggs20eccv, khosla11fgvc} &Close-ups    & \nota     & \nota     & 20        & \nota     & 7K        & 5K    & \\
        \cmidrule(r){2-10}
        19  & vKITTI2                 &\cite{cabon20vkitti2, gaidon16cvpr}& Synthetic (driving) & 9  & \maybe    & \nota     & \present  & 43K       & 7K \\        
        20  & vGallery                  &\cite{weinzaepfel19cvpr}& Synthetic (indoor)     & 8         & \maybe    & \nota     & \present  & 44K       & 12K \\
        \cmidrule(r){1-10}
            & ILSVRC'12/ImageNet        &\cite{ILSVRC15, deng09cvpr} & Websearch		& \multicolumn{4}{c}{1,000 classes for classification}	& 1.2M	& 50K & \\\bottomrule
        \end{tabular}
    }
    \end{threeparttable}
    \begin{tablenotes}
    \item[a] From the Pascal VOC datasets we use VOC2012 for semantic segmentation and VOC2007 for object detection. %
    \item[b] For object detection BDD100K is used. %
    \item[c] iSAID and ISPRS consist of few very high resolution images. Using the procedure of the \href{https://github.com/CAPTAIN-WHU/iSAID_Devkit}{iSAID devkit}~\cite{waqas2019isaid}, we split each high-res image into multiple partially overlapping low-res images ($800\times800$), which we use for training and evaluation. Train and validation images originate from different high-res images and hence are not overlapping in content. %
    \item[d] The SUN RBG-D dataset contains imagery from NYU depth v2 \cite{silberman12eccv}, Berkeley B3DO \cite{janoch11iccvw}, and SUN3D \cite{xiao13iccv}. %
    \end{tablenotes}    
    \caption{
        Overview of the 20 datasets used in this paper. They vary in size, task types (we consider semantic segmentation, objct detection, keypoint detection, depth estimation), number of classes, and image domain. We included datasets from consumer imagery, driving, aerial, indoor, underwater, close-ups, and synthetic domains.
        Some datasets (such as KTTI~\cite{alhaija18ijcv}) have more annotation types available, but are not used in this study.
    }
    \label{tab:total_transfer_datasets_overview}
\end{table*}

To study these factors of influence on transfer learning we selected 20 public available datasets annotated for one (or more) of our task types, ensuring to span various very different image domains and with a large variety in dataset size.
An overview of the datasets and their task types which we use is given in~\autoref{tab:total_transfer_datasets_overview}. 
The variety in visual appearance is illustrated in ~\autoref{fig:total_transfer_datasets_illustration}.

While we consider the annotations listed in~\autoref{tab:total_transfer_datasets_overview}, some datasets have additional annotation types available. For example, CityScapes~\cite{cordts16cvpr} also provides disparity and Berkeley Deep Drive ~\cite{bdd100k} also includes road segmentation annotations.

In our experiments, each dataset plays both roles of source and target, in different experiments in turn.
This allows us to extensively study the aforementioned factors of influence for transfer learning
over a wide range of image domains, both within task types and across task types, and for training sets of various size.
\section{Setting up Network Architectures for Transfer Learning}
\label{sec:networks}
\begin{figure*}[t]
    \centering
    \begin{subfigure}[t]{\columnwidth}
        \centering
        \includegraphics[width=\columnwidth]{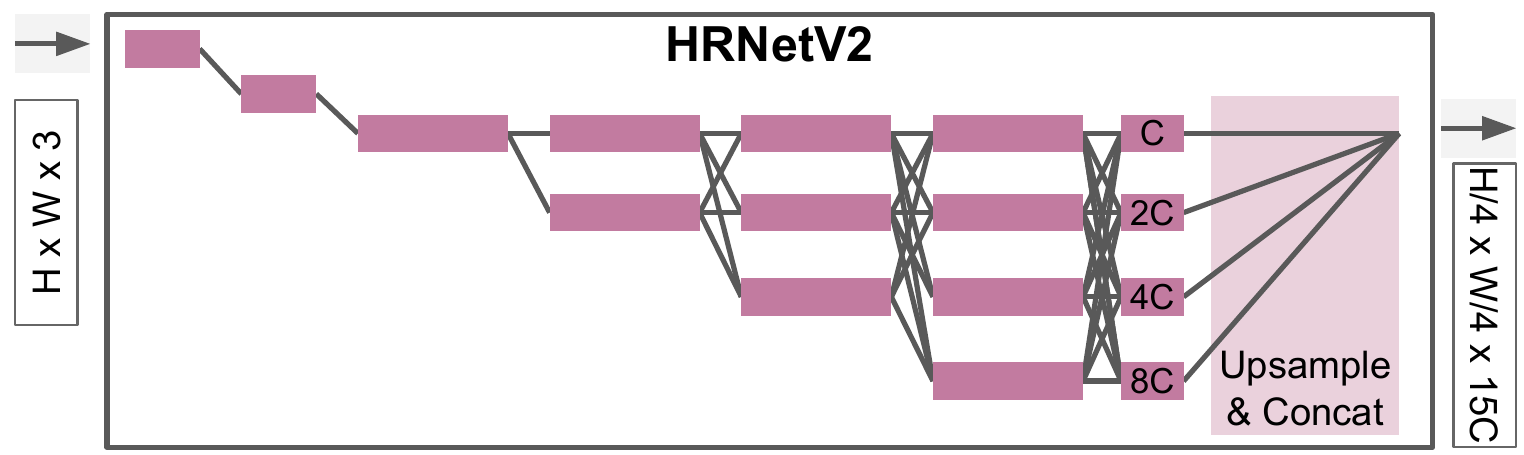}    
        \caption{}
        \label{fig:network_backbone}
    \end{subfigure}
    \hfill
    \begin{subfigure}[t]{\columnwidth}
        \includegraphics[width=\columnwidth]{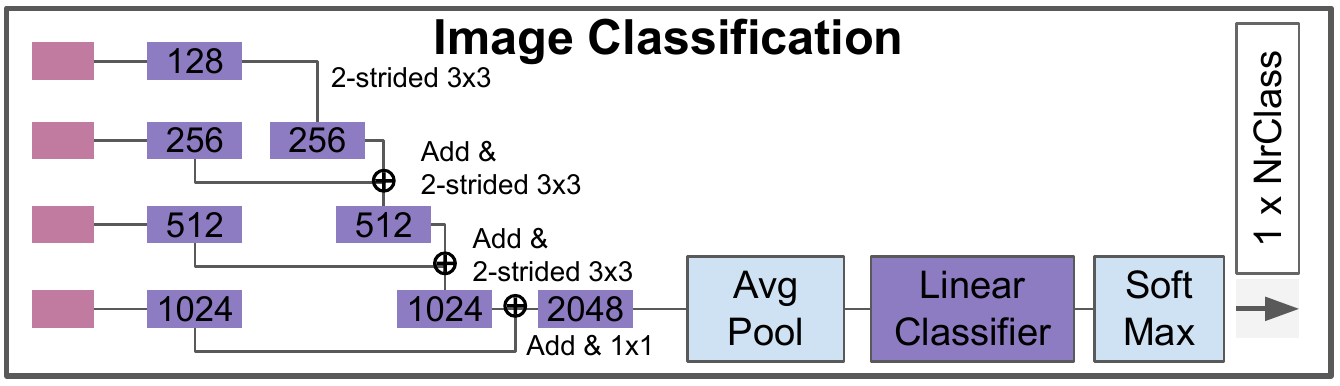}
        \caption{}
        \label{fig:network_classification_head}
    \end{subfigure}
    
    \begin{minipage}[b][3.7cm][t]{.59\columnwidth}
        \centering
        \includegraphics[width=\textwidth]{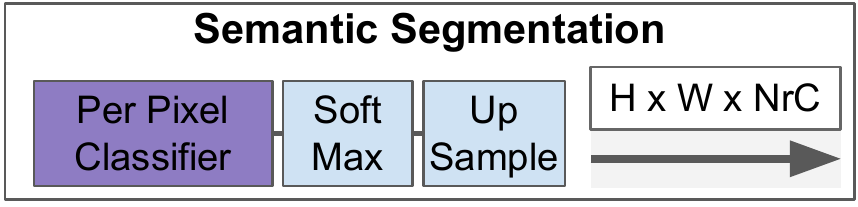}
        \subcaption{}\label{fig:network_semseg_head}
        
        \includegraphics[width=\textwidth]{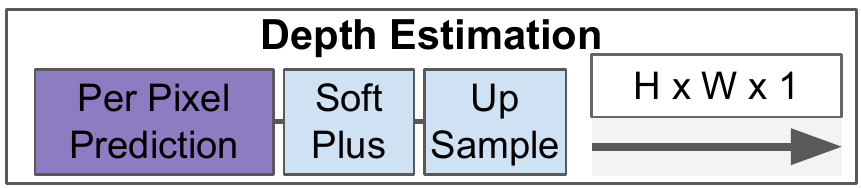}
        \subcaption{}\label{fig:network_depth_head}
    \end{minipage}    
    \hfill
    \begin{minipage}[b][3.7cm][t]{.73\columnwidth}
        \includegraphics[width=\textwidth]{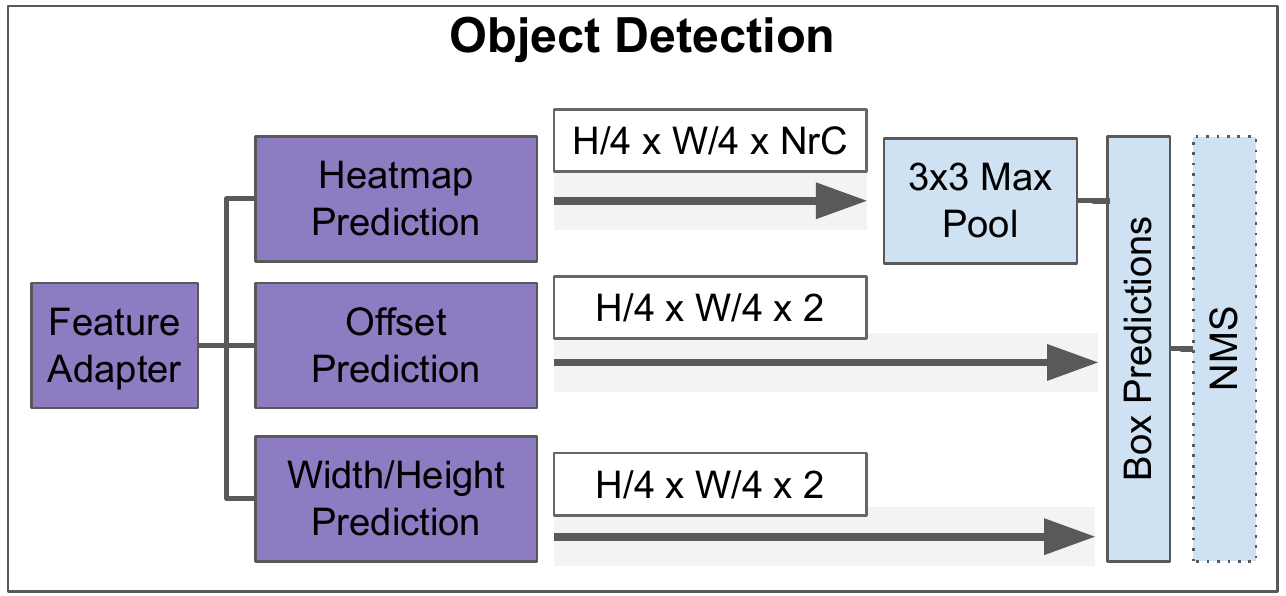}
        \subcaption{}\label{fig:network_detection_head}
    \end{minipage}
    \hfill
    \begin{minipage}[b][3.7cm][t]{.65\columnwidth}
        \includegraphics[width=\textwidth]{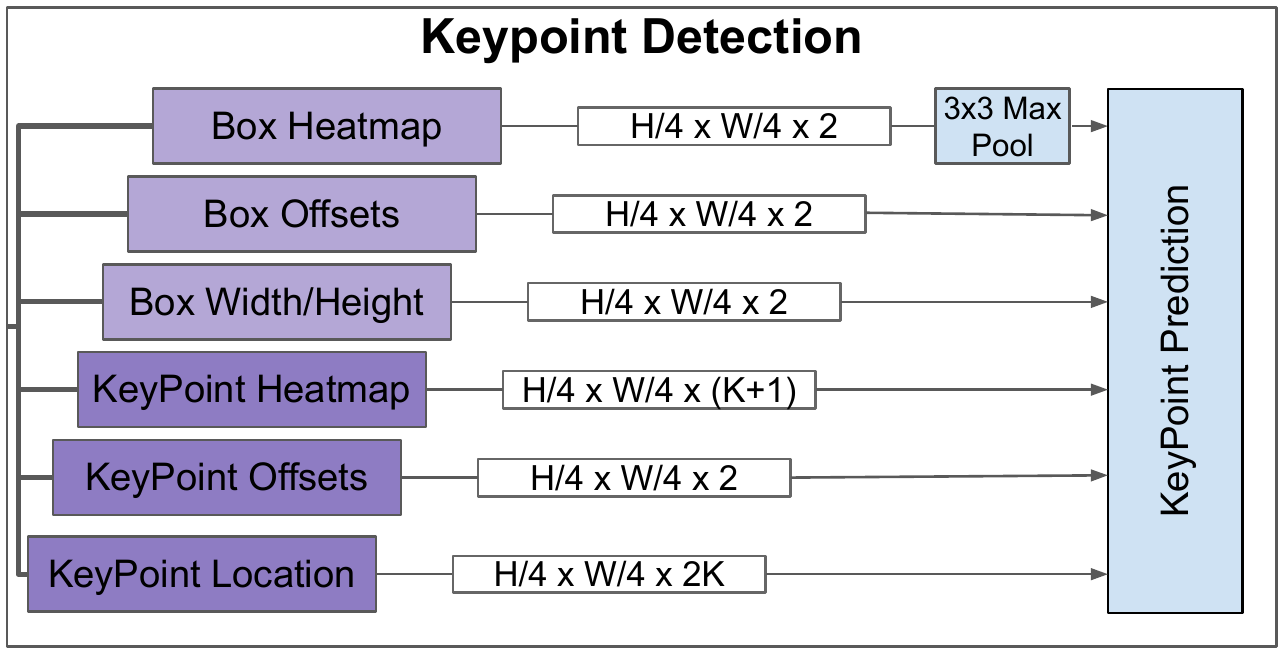}
        \subcaption{}\label{fig:network_keypoints_head}
    \end{minipage}
    
    \caption{\changed{Schematic illustrations of the backbone (a) and task type specific network architectures (b -- f). Red and purple blocks are trainable (backbone and task type specific networks respectively), while blue blocks have no trainable parameters. }
    }
    \label{fig:network_architectures}
\end{figure*}

This section describes our network setup.
Three things are important for our study: a common framework with shared data augmentation, a common backbone, and high quality models specific to each task type.

\subsection[Data normalization]{Data normalization and augmentation}\label{sec:data_normalization}

During training various data normalization and augmentation techniques are typically used~\cite{krizhevsky09,cubuk19cvpr} which have significant impact on model performance.
We unify the data normalization and augmentation step, because it changes the input of the network, \ie it influences the low-level statistics of the input imagery. 
This allows to prevent variations in data normalization and augmentation from affecting transfer learning performance.

Data augmentation also influences what the model learns.
For example, applying large rotation transformations or anisotropic scaling
makes the network invariant to those aspects, which may be beneficial for some tasks but detrimental for others, \eg full rotation invariant networks are not able to distinguish \emph{6} from \emph{9}.
In this paper we do not want data normalization and augmentation to be confounding factors in transfer learning experiments.
Hence we aim to keep data normalization and augmentation as simple as possible without compromising on accuracy, and apply the same protocol to all experiments.

\para{Illumination normalization.}
For each dataset we normalize the images so that each color channel has zero mean and standard deviation $1$.
This form of whitening minimizes illumination biases from the different datasets as much as possible.

\para{Data augmentation.}
To unify the data augmentation across all four task types considered, we ran many experiments to estimate the importance of common data augmentation techniques.
We found the following augmentations to always have a positive (or neutral) effect and thus use them in all experiments:
(1) random horizontal flipping of the image;
(2) random rescaling of the image; and
(3) taking a random crop.
For object detection and keypoint detection, we consider only random crops fully containing at least one object. 
As in~\cite{lambert20cvpr}, we found that the image scales that lead to the best performance are intrinsic to the dataset, and hence dataset-dependent. Since we see this (partly) as a property of the image domain, we optimized for each task the input resolution of the network.
Overall, the input resolutions during training range from $420 \times 420$ for Pascal Context to $713 \times 713$ for SUN RGB-D to $512 \times 1024$ for CityScapes, where the latter is a landscape format common to most driving datasets.
We always evaluate at a single scale and resize each image such that one side matches the input resolution used for that task. %

After careful consideration, we decided not to use image rotation, varying image aspect ratio, or any form of color augmentation.
\emph{Image rotation} is incompatible with object detection which (generally) assumes axis-aligned ground-truth boxes.
Moreover, we were able to reproduce the semantic segmentation performance of~\cite{lambert20cvpr} even without rotation augmentation (see~\autoref{sec:single_task}).
\emph{Varying image aspect ratio} did not yield positive effects on semantic segmentation, and also such augmentation is uncommon for object detection and keypoint detection.
\emph{Color augmentation} random changes in hue, contrast, saturation, and brightness did not yield positive effects for us on semantic segmentation and object detection, while substantially slowing down training. 

\para{Our Setup.}
To conclude, we consistently apply a limited number of data augmentation techniques across all experiments. The only difference across experiments is the input resolution, which is fixed per-dataset and is thus consistent with the image domain factor of variation we want to study (\autoref{sec:image_domain}).
This uniform protocol effectively cancels out varying data augmentation as a potential causal factor influencing the performance of transfer learning.

\subsection{\changed{Network Architectures}}\label{sec:network_architectures}
\begin{table}[t]
    \centering
    \begin{tabular}{c|cccc}\toprule
         Backbone   & Depth     & Semseg    & Detection         & Keypoints  \\
                    &           & 20        & 20                & 17         \\\midrule
         69M        & 721       & 14K       & 2.4M              & 10M\\
                    & $<1\%$    & $<1\%$    & $3.5\%$           & $14\%$\\
        \bottomrule
    \end{tabular}
    \caption{Number of parameters in the backbone and the respective task heads. 
    The backbone forms the majority of the neural network and contains visual knowledge which could be shared to new target tasks.
}
    \label{tab:num_parameter_per_task}
\end{table}
\begin{table*}
    \centering
    
    \begin{subtable}[t]{.54\columnwidth}
        \begin{tabular}{lc}\toprule
            Description                 & Top-1 Acc\\\midrule
            ResNet-50\cite{wang20pami}  & 76.9      \\
            HRNetV2-W44\cite{wang20pami}& 77.0      \\\midrule
            \textbf{ours} HRNetV2-W48   & 79.5      \\\bottomrule
        \end{tabular}
        \caption{}
        \label{tab:results_classification_head}
    \end{subtable}
    ~
    \begin{subtable}[t]{.64\columnwidth}
        \begin{tabular}{lcc}\toprule
            Description                         & RMSE ($\downarrow$)   & $\delta_1 (\uparrow)$\\\midrule
            ResNet DORN \cite{fu18cvpr}         & .509                  & 82.8                      \\
            ResNet D-DFN\cite{chen193dv}        & .528                  & 86.6                      \\\midrule
            \textbf{ours} HRNetV2-W48           & .493                  & 82.5                      \\\bottomrule
        \end{tabular}
        \caption{}
        \label{tab:results_depth_head}
    \end{subtable}
    ~
    \begin{subtable}[t]{.775\columnwidth}
        \centering
        \begin{tabular}{lc}\toprule
            Description                                             & AP50\\\midrule
            Hourglass-104 - CenterNet - 1 stage~\cite{zhou19arxiv} & 84.2\\ %
            HRNetV1-W44 - 2 stage cascade~\cite{wang20pami}& 90.8\\\midrule
            \textbf{ours} HRNetV2-W48 - CenterNet - 1 stage    & 81.1\\\bottomrule
        \end{tabular}
        \caption{}
        \label{tab:results_keypoints_head}
    \end{subtable}
    
    \begin{subtable}[t]{.9\columnwidth}
        \centering
        \begin{tabular}{lc}\toprule
            Description                                         & MAP\\\midrule 
            ResNet-101 - CenterNet~\cite{zhou19arxiv}           & 34.6 \\
            Hourglass ResNet-104 - CenterNet~\cite{zhou19arxiv} & 40.3 \\
            Hourglass ResNet-104 MS+flip - CenterNet~\cite{zhou19arxiv} & 45.1 \\
            HRNetV2-W48-pyramid - CenterNet~\cite{wang20pami}       & 43.4\\\midrule
            \textbf{ours} HRNetV2-W48 - CenterNet               & 39.0\\\bottomrule
        \end{tabular}
        \caption{}
        \label{tab:results_detection_head}
    \end{subtable}\vspace{-2mm}
    \quad
    \begin{subtable}[t]{.6\columnwidth}
        \centering
        \begin{tabular}{lcc}\toprule
            Dataset         & MSEG \cite{lambert20cvpr}  & Ours\\\midrule
            BDD             & 63.2          & 65.0\\
            CityScapes      & 77.6          & 75.7\\
            COCO Panoptic   & 52.6          & 54.0\\
            Pascal Context  & 46.0          & 47.1\\
            ScanNet         & 62.2          & 61.9\\\bottomrule
        \end{tabular}
        \caption{}
        \label{tab:results_semseg_head}
    \end{subtable}\vspace{-2mm}
    
    \caption{\changed{Comparison of our task type specific networks to recent networks on standard benchmarks (see text for details).}  
    }
\end{table*}

Transfer learning through pre-training is only possible when the models for all task types share the same backbone architecture. To have meaningful results, we need to choose a backbone which works well across all task types we explore. 
In this section we outline the used backbone architecture and the task type specific networks, these are illustrated in~\autoref{fig:network_architectures}.
More details are in~\appendixsecref{sec:appendix_networks}.

\para{Backbone Architecture.}
For the backbone the recent high-resolution HRNetV2~\cite{wang20pami} architecture was chosen.
It extends the ResNet architecture to preserve high-resolution spatial features.
Where a regular ResNet reduces the image resolution at every stage, HRNetV2 \emph{also} keeps a parallel high-resolution branch, as illustrated in~\autoref{fig:network_backbone}.
The resulting four output maps are combined into a single feature map by upscaling and concatenation. The backbone is pre-trained with supervised classification on ILSVRC'12 using the architecture shown in~\autoref{fig:network_classification_head}.

HRNetV2 has two main advantages.
First, it outperforms ResNet~\cite{he16cvpr}, ResNeXt~\cite{xie16cvpr}, Wide ResNet~\cite{zagoruyko16bmvc}, and Stacked Hourglass Networks~\cite{newell16eccv} on three of the tasks we explore: semantic segmentation, keypoint detection, and object detection.
Second, the high-resolution output feature maps allow to use relative shallow task type specific heads, see~\autoref{tab:num_parameter_per_task} for the number of trainable weights.

\para{Semantic Segmentation.}
For semantic segmentation, the task type head is shown in~\autoref{fig:network_semseg_head}, it is an adoption of the network head proposed in~\cite{wang20pami}.
It consists of a linear classifier, a softmax layer, and a bi-linear up-sampling layer to produce the final predicted segmentation map in the input image resolution.

\para{Object Detection.}
For object detection we follow the CenterNet\footnote{With CenterNet we denote a family of detection models which predicts boxes via their center points~\cite{duan19iccv,zhou19arxiv}.}
approach of Zhou et al.~\cite{zhou19arxiv} as shown in~\autoref{fig:network_detection_head}.
Each pixel is classified as being the center point of a bounding box of a specific class (\ie the \textit{heatmap}); additionally the bounding box size and offsets are predicted. Using center points to predict bounding boxes results in a simpler model than a two-stage architecture like Faster-RCNN~\cite{ren15nips}, while being about as accurate~\cite{duan19iccv,zhou19arxiv}.

\para{Keypoint Detection.}
For keypoint detection we follow (again) the CenterNet approach of~\cite{zhou19arxiv} as illustrated in~\autoref{fig:network_keypoints_head}. 
This essentially uses the object detection architecture to predict a bounding box for each person/dog, and then predicts the location of the keypoints within this bounding box. 

\para{Depth Estimation.}
For monocular depth estimation we mimic the architecture of the semantic segmentation head. 
We use a single regression layer, followed by a soft-plus layer and a bi-linear upsampling layer, see \autoref{fig:network_depth_head}.
The soft-plus activation: $\log(\exp(x) + 1)$ is a differential clipping to convert the logit values to depth ensuring that the predictions are positive, it is also used in~\cite{gordon19iccv, zhu19iccv}.

\subsection{\changed{Single Task Performances}}
\label{sec:single_task}

In order to validate our setup, we compare the performance of the used networks for each task type to a set of baselines.
More details are in~\appendixsecref{sec:appendix_networks}.

\para{Backbone.}
In order to validate our backbone architecture we use image classification on ILSVRC'12 and measure Top-1 accuracy.
We compare our setup to a ResNet50 and HRNetV2-W44~\cite{wang20pami} network. 
The results are in~\autoref{tab:results_classification_head}. 
Our model performs best, reaching $79.5\%$ Top-1 accuracy, validating our choice of architecture.

\para{Semantic Segmentation.}
To validate our setup, we compare on a subset of the MSEG dataset collection~\cite{lambert20cvpr}.
For this experiment we use the annotation provided in MSEG, which uses fewer classes than our transfer setting setup. 
Training starts from a ILSVRC'12 pre-trained backbone. %

Performance is evaluated by the Intersection-over-Union averaged over classes (mIoU)~\cite{everingham15ijcv}.
During evaluation we process complete images at a single resolution only. 
The results are shown in~\autoref{tab:results_semseg_head}. 
Our models perform on par with those of~\cite{lambert20cvpr}, which uses a similar architecture, but with different normalisation and data augmentation steps and a different implementation.

\para{Object Detection.}
We validate our setup by training on COCO17 and evaluate  performance on its 5K validation set, without using Non-Maximum Suppression (NMS). We compare to results reported in~\cite{zhou19arxiv} and~\cite{wang20pami}.

The evaluation is based on the COCO definition of mean Average Precision (mAP)~\cite{lin14eccv}. 
The results are in~\autoref{tab:results_detection_head}.
We observe that the best results are obtained by using data augmentation during evaluation or by using a feature pyramid in the detection phase.
Without these enhancements, our implementation performs close to the Hourglass ResNet-104 and outperforms the ResNet-101.
Hence we conclude that we can base our transfer learning experiments on a strong, modern object detection framework.

\para{Keypoint Detection.}
We validate our setup on the keypoint detection task on the COCO dataset
and report the mean Average Precision at 0.5 Object Keypoint Similarity (AP50)~\cite{coco-challenge}. 

Results are in~\autoref{tab:results_keypoints_head}, where we compare to the Hourglass-104 CenterNet~\cite{zhou19arxiv} which yields 84.2 AP50\footnote{
Result obtained via correspondence with the authors: The public available model in the zoo (see \href{https://github.com/xingyizhou/CenterNet}{repo}) obtaining 64.0 AP50 on \texttt{coco-test}, obtains 84.2 AP50 on \texttt{coco-val}, without flipping. %
}
and to the HRNetV1 Cascade~\cite{wang20pami} which yields 90.8. Our setup uses a much simpler single-stage CenterNet head. With a performance of 81.1 AP50, it is near~\cite{zhou19arxiv} and thus strong enough for our transfer learning exploration.

\para{Depth Estimation.}
To validate our setup we compare monocular depth estimation on the NYUDepthV2~\cite{silberman12eccv} dataset, using the root mean squared error (RMSE) metric and the $\delta < 1.25$ accuracy, where $\delta = \max(\tfrac{\hat{z}}{z},\tfrac{z}{\hat{z}})$ is a measure of relative accuracy defined in~\cite{ladicky14cpr}.

The results are in~\autoref{tab:results_depth_head}, where we compare to two recent ResNet based models: \cite{fu18cvpr} uses depth-specific losses based on ordinal regression; and
\cite{chen193dv} uses a depth-specific network architecture, but with the same loss function as we do. 
From the results.
Our proposed light-weight depth prediction model outperforms both depth specific models on RMSE and hence, it is well suited for our monocular depth estimation transfer learning experiments. 
\section{Transfer learning experiments}
\label{sec:transfer_experiments}
In this section we describe our transfer learning experiments. We mainly conduct experiments in two settings: transfer learning with a small target training set and with the full target set.
The analysis of the results across all experiments will be discussed in~\autoref{sec:analysis}.

\subsection{Setup}
\label{sec:transfer_experiments_setup}
\para{Transfer Chains.}
\changed{
In our experimental setup we consider \emph{transfer chains}: ILSVRC'12 $\rightarrow$ source $\rightarrow$ target. Specifically, 
we first train a single classification model on ILSVRC'12, whose weights we reuse in all our experiments.
To train a source model $S$, we first copy the ILSVRC'12 backbone weights into it. We randomly initialize the task head of $S$. Then we fine-tune until convergence on the source training set. This results in a single set of backbone weights per source task which we reuse in all our experiments. Analogously, we continue and copy the backbone weights of $S$ into the backbone of $T$, randomly initialize the task head of $T$, and fine-tune until convergence on the target training set.
}

\changed{
\para{Baseline.} As baseline we consider the default in the community, which is starting from ILSVRC'12 pretrained backbone weights: ILSVRC'12 $\rightarrow$ target.
}

\para{Evaluation.} 
\changed{Our core question is: can we get additional gains over our baseline by picking a good source?
We therefore measure improvements w.r.t. the baseline, which we call \emph{relative transfer gain}:}
\begin{equation}
   r(T | S) = \left(\frac{m(T | S)}{m(T | \textrm{ILSVRC})}-1 \right) * 100
   \label{eq:relative_transfer_gain}
\end{equation}
where $m$ denotes a metric specific for the task type (\eg mean-IoU for semantic segmentation), which is evaluated for all tasks on complete images at a single resolution only.
Since for depth estimation lower values of $m$ mean better performance, we multiply $r$ by $-1$ in that case.

This notion of gain is similar in spirit to the one defined in Taskonomy~\cite{zamir18cvpr}.
However their gain is the percentage of test images for which the transfer model outperforms a model trained from scratch. 
Instead, our \emph{relative transfer gain} metric refers to a much stronger baseline: a model fine-tuned from ILSVRC'12. Moreover, we evaluate \emph{how much better} the target model becomes in terms of a standard metric specific to each task type, averaged over all test images.

\para{Multi-Source Models.} 
Inspired by the success of using generic visual representations for various computer vision tasks~\cite{kolesnikov20eccv,kokkinos17cvpr,lambert20cvpr}, we include a multi-source model in our experiments. 

We train a multi-source model for a specific task type based on several datasets. For each dataset a separate head is attached to a single, common backbone:
\begin{itemize}
    \item semantic segmentation is trained across iSAID, COCO, Mapillary, ScanNet, SUIM, vGallery, and vKITTI2;
    \item depth estimation is trained across all three depth datasets we consider: SUN RGB-D, vGallery, and vKITTI2;
    \item object detection is trained across all four object detection datasets: COCO, BDD, Pascal VOC, and Underwater Trash.
\end{itemize}
The multi-source models are trained using dataset interleaving at the batch level, \ie each batch is sampled from another dataset, and each head classifies only the images from that particular dataset. 
When doing transfer learning, only the weights of the common backbone are used as a source model.

\changed{\para{Training.} We determined the number of training steps per dataset for all source models in preliminary experiments. 
We use these for source model training and in the full target training setting. 
We lower the number of steps for the small target training setting. 
For each experiment in this paper, we selected the best model from three learning rates. More training details are in \appendixsecref{sec:appendix_networks}.
}

\subsection[Small Target Set Transfer]{Transfer learning with a small target training set}\label{sec:limited_target}
\begin{table}[tp]
    \begin{subtable}[t]{\columnwidth}
        \includegraphics[width=\columnwidth]{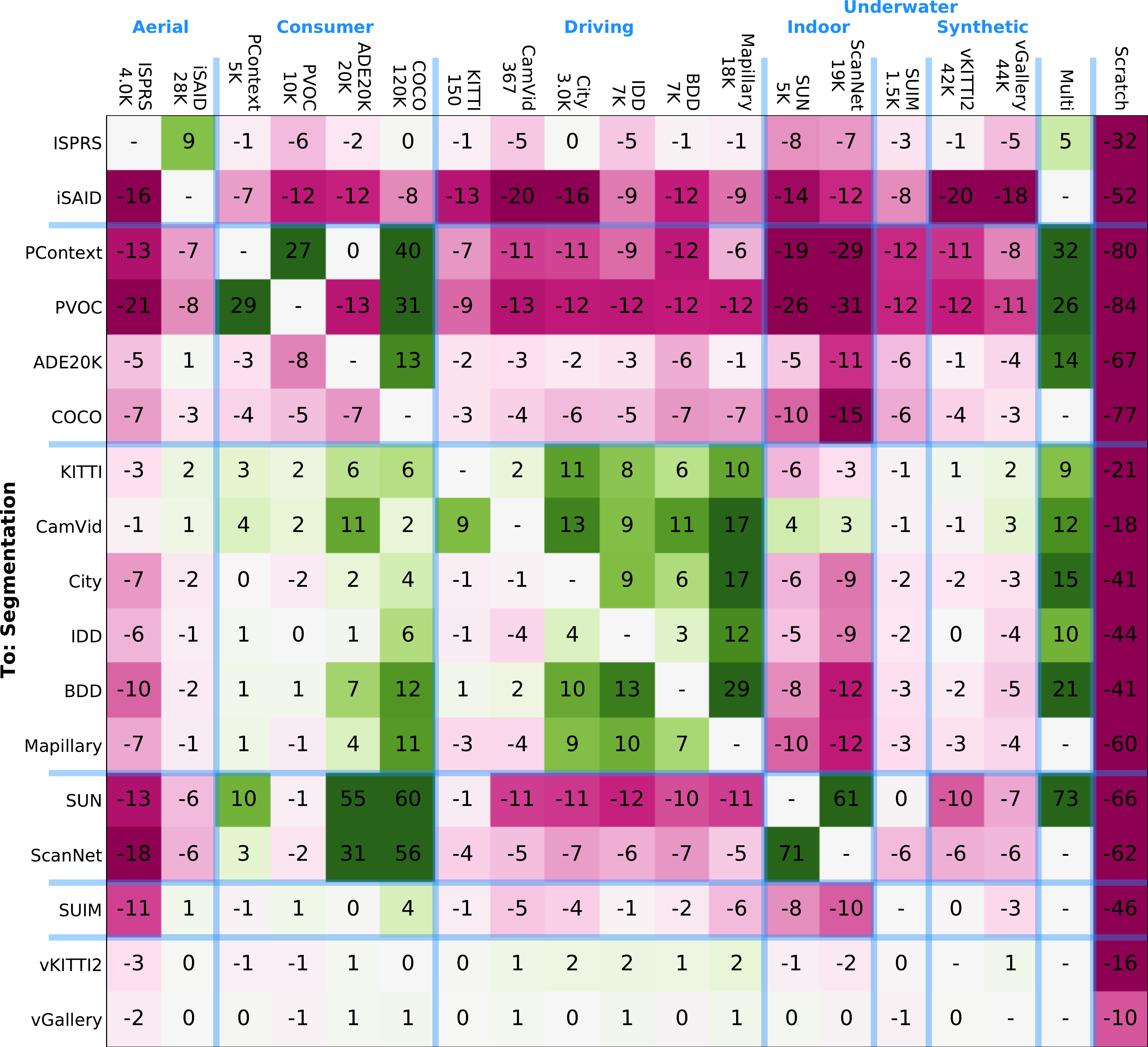}
        \caption{}\label{tab:few_shot_semseg_within}
    \end{subtable}
    \begin{subtable}[t]{\columnwidth}
        \centering
        \includegraphics[width=.85\columnwidth]{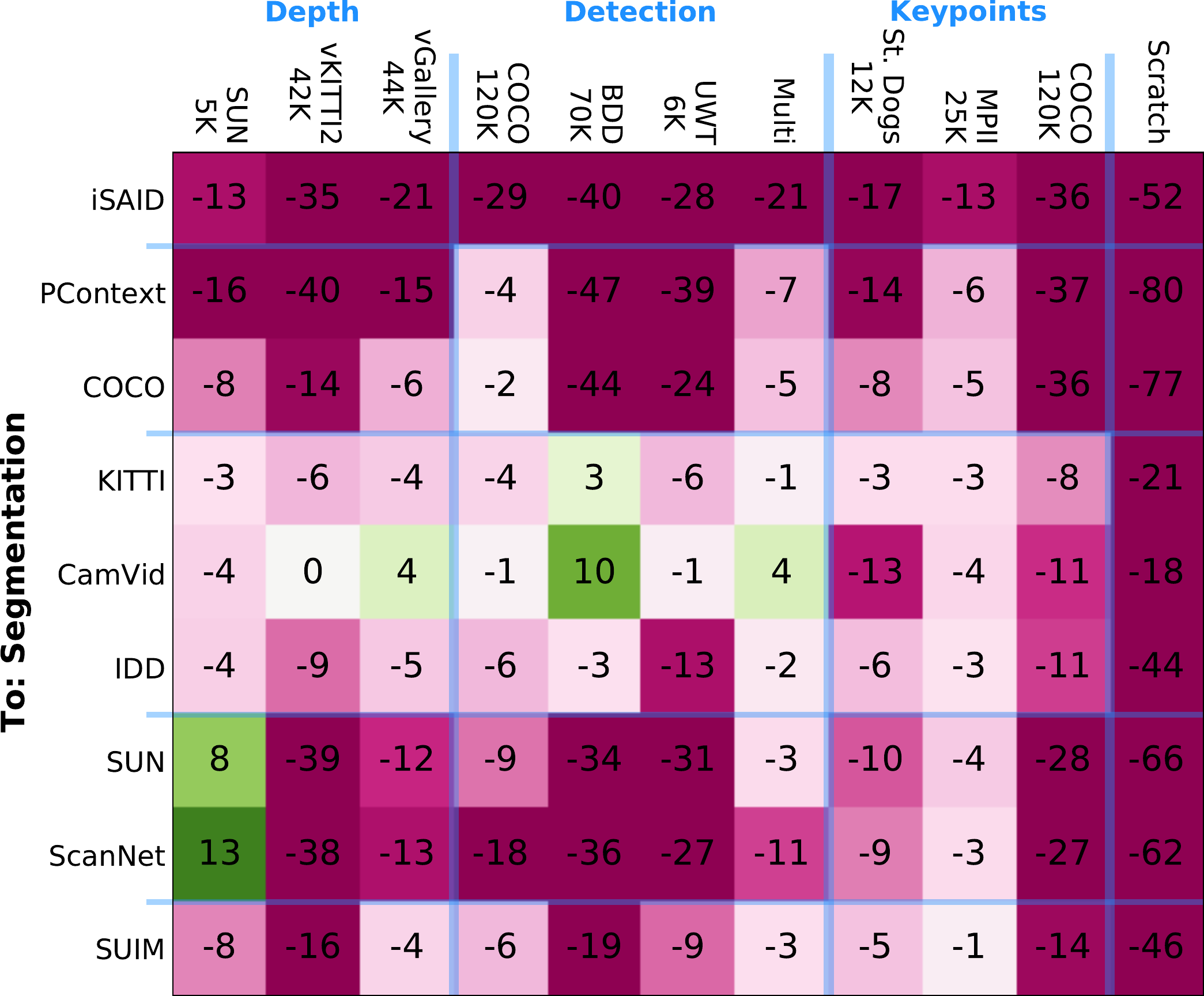}
        \caption{}\label{tab:few_shot_semseg_cross}
    \end{subtable}
    \begin{subtable}[t]{\columnwidth}
        \centering
        \includegraphics[width=.95\columnwidth]{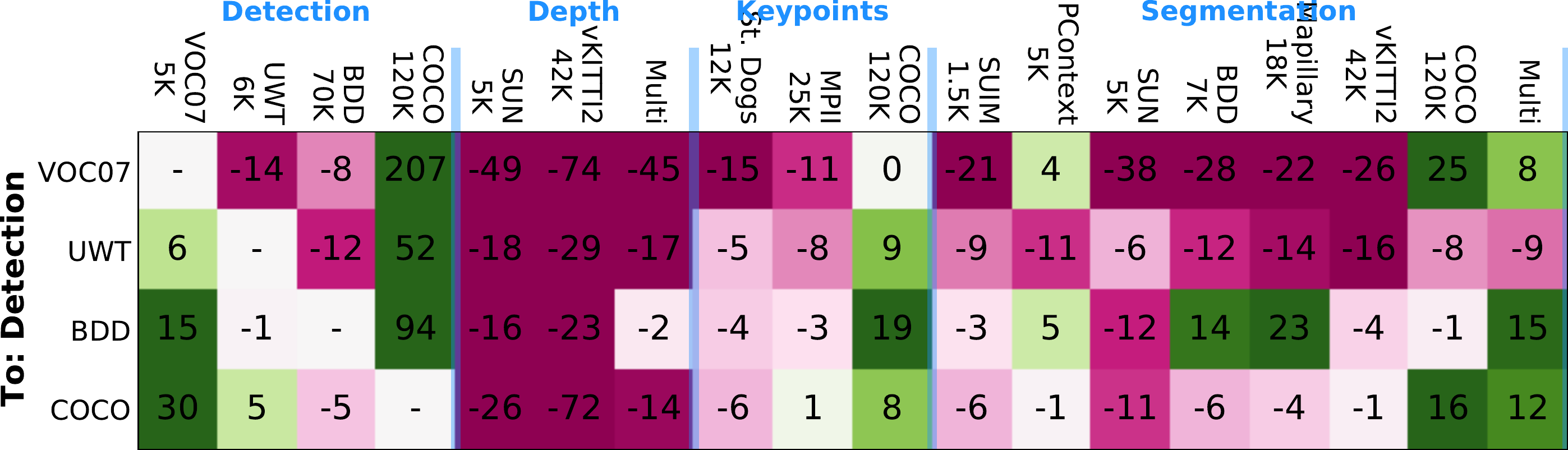}
        \caption{}\label{tab:few_shot_detection}
    \end{subtable}
    \begin{subtable}[t]{\columnwidth}
        \centering
        \includegraphics[width=.5\columnwidth]{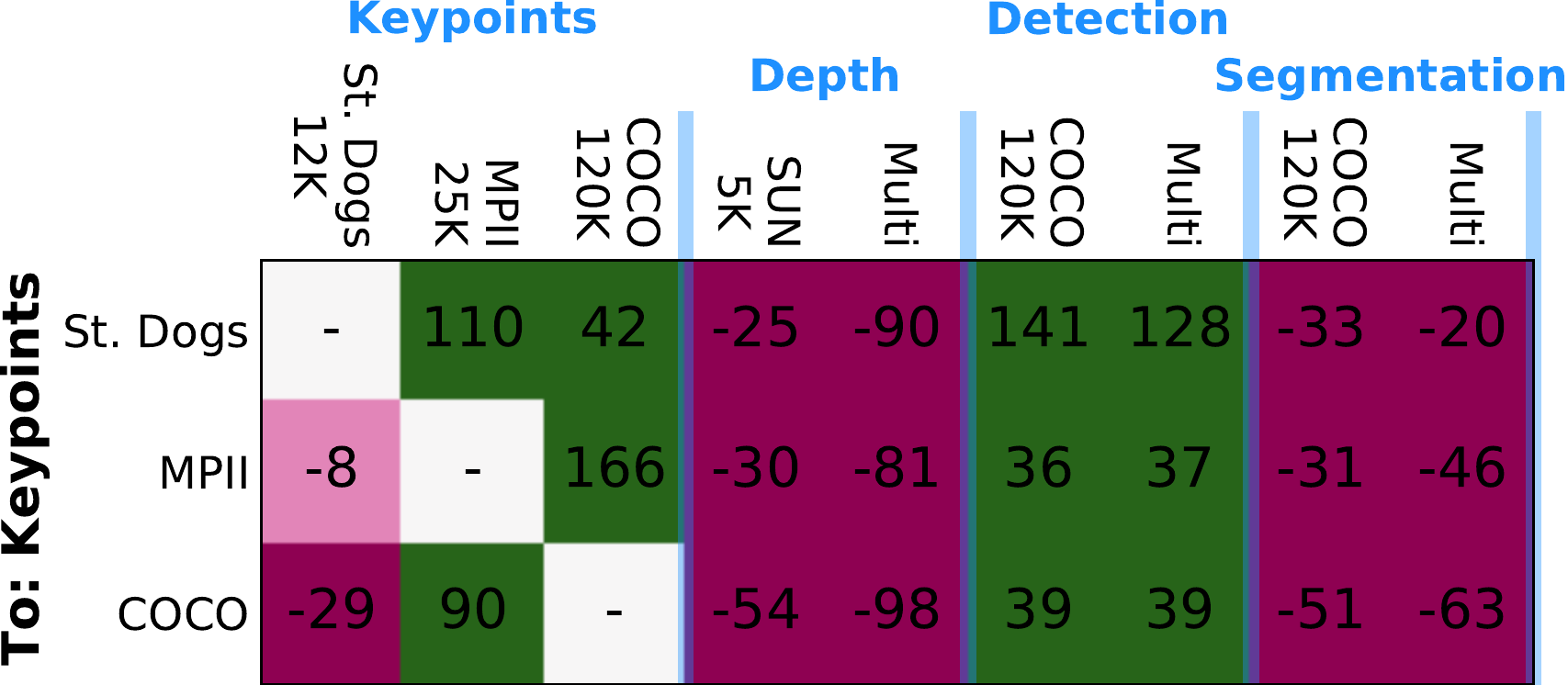}
        \caption{}\label{tab:few_shot_keypoints}
    \end{subtable}
    \begin{subtable}[t]{\columnwidth}
        \centering
        \includegraphics[width=.95\columnwidth]{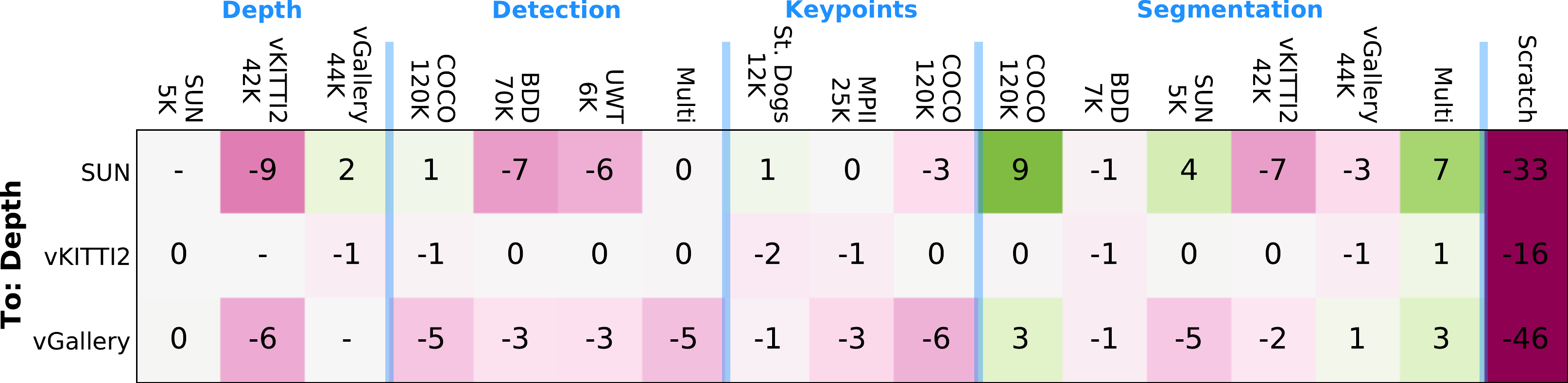}
        \caption{}\label{tab:few_shot_depth}
    \end{subtable}
    \caption{
    Relative improvement gains for the small target training setting.
    In \autoref{tab:few_shot_semseg_within} both the source and target task types are semantic segmentation.
    See main text for details.}
    \label{tab:few_shot}
\end{table}

The first setting we study is transfer learning with a small target training set, in which we limit the number of annotated examples are available for training.
We deem this setting as the most challenging and practically relevant: obtain good performance for a structured prediction task by annotating only a few images. Transfer learning is particularly relevant in such a low-data regime.

Concretely, we limit the number of target training images to 150 per dataset (this is less than 3\% of the available train data for 14 out of the 20 datasets).
For COCO and ADE20k we make an exception and use 1000 images, which is still less than 1\% of the available train data for COCO and 5\% for ADE20K.
The reason is that these datasets have a large number of classes following a long-tail distribution (COCO has 134 classes for segmentation, and ADE20K has 150). Just 150 images did not properly cover all classes: 
In all experiments we use a seeded selection, such that for each dataset all models are fine-tuned using the exact same training samples.

\para{Experiment.}
Our results are shown in~\autoref{tab:few_shot} (all relative transfer gains). Positive transfer gains are in green, negative transfer gains are in purple, no gains are represented by white. For ease of presentation, we group results by the target task type, which leads to multiple tables. Then we group by the source task type (two separate tables for semantic segmentation, blue vertical line in other experiments). Then we group by image domain (marked in blue above in \autoref{tab:few_shot_semseg_within}). Finally, within each image domain we order sources by their size. For completeness, absolute performance table are available in~\appendixsecref{sec:appendix_results}.
We tried to keep all experiments as homogenous as possible, while maintaining a wide scope of transfer across many diverse domains and task types. This leads to two peculiarities, which we clarify below.

First, in most experiments, the source training set and the target training set are disjoint (i.e. always when transferring within a task type, and most of the time across task types).
However, when transferring across task types, a few of the source-target pairs use (partly) the same training set, \eg COCO Object Detection as source for COCO Keypoint Detection in~\autoref{tab:few_shot_keypoints} or BDD Semantic Segmentation as source for BDD Object Detection in~\autoref{tab:few_shot_detection}.
In these experiments the source models have been trained on \emph{all} images of the training dataset for the source task type, while the target model is fine-tuned on only a small part of that training set, and for a different task type.
This setting is relevant if one is interested in extending the annotation of a dataset from one task type to another (e.g. one has a lot of object bounding boxes, but very few depth masks for a dataset).

Second, when transferring within a task type, the multi-source models are only applied for target datasets which were not part of their training (hence the blank entries for this column in~\autoref{tab:few_shot_semseg_within}). This is  necessary for meaningful experiments, as otherwise some target training sets would simply be a subset of the images and annotations present in the multi-source.

\subsection{Transfer learning with full target training set}

\begin{table}[tp]
    \centering
    \begin{subtable}[t]{\columnwidth}
        \includegraphics[width=\columnwidth]{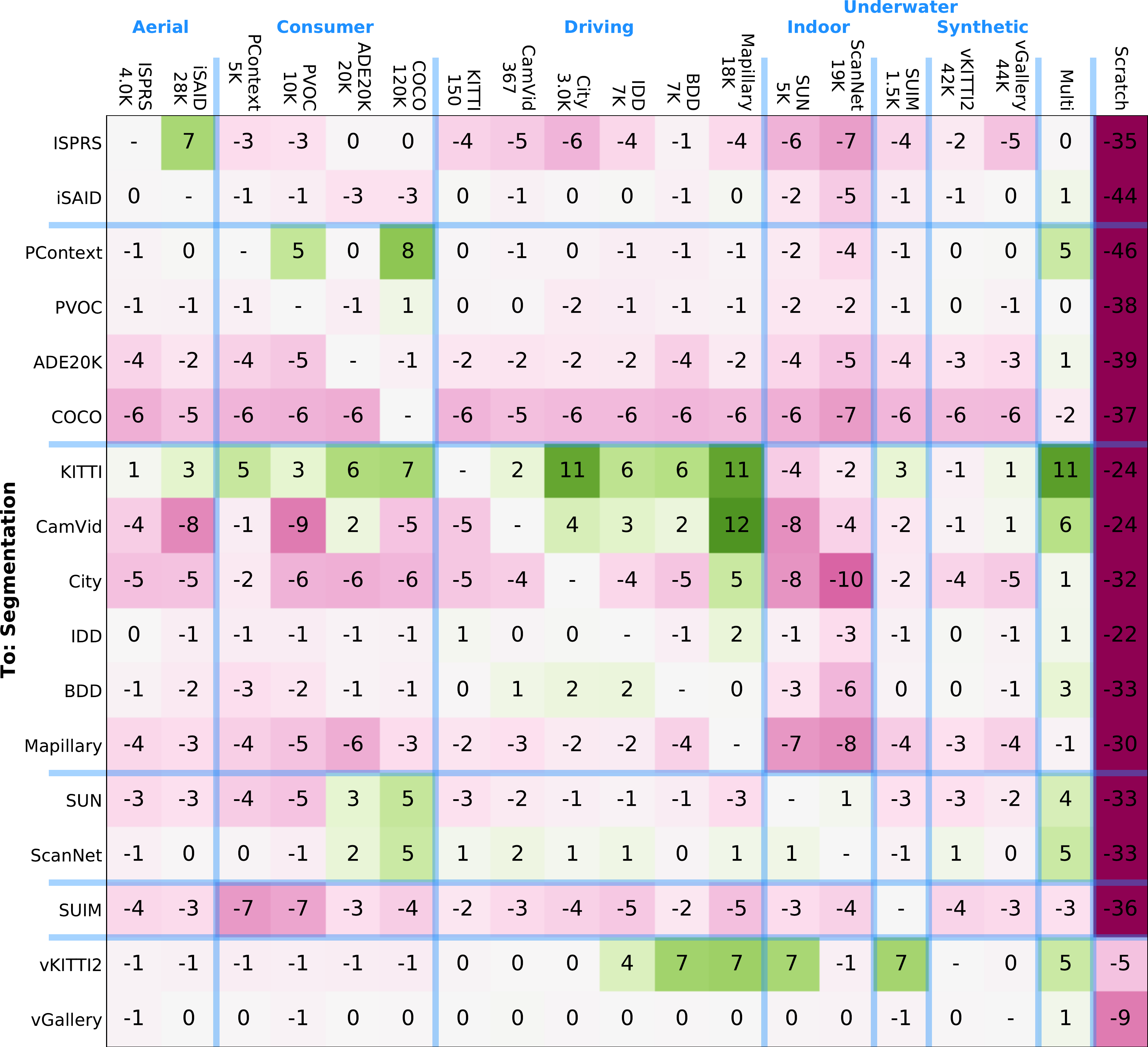}
        \caption{}\label{tab:full_target_semseg_within}
    \end{subtable}
    \begin{subtable}[t]{\columnwidth}
        \centering
        \includegraphics[width=.9\columnwidth]{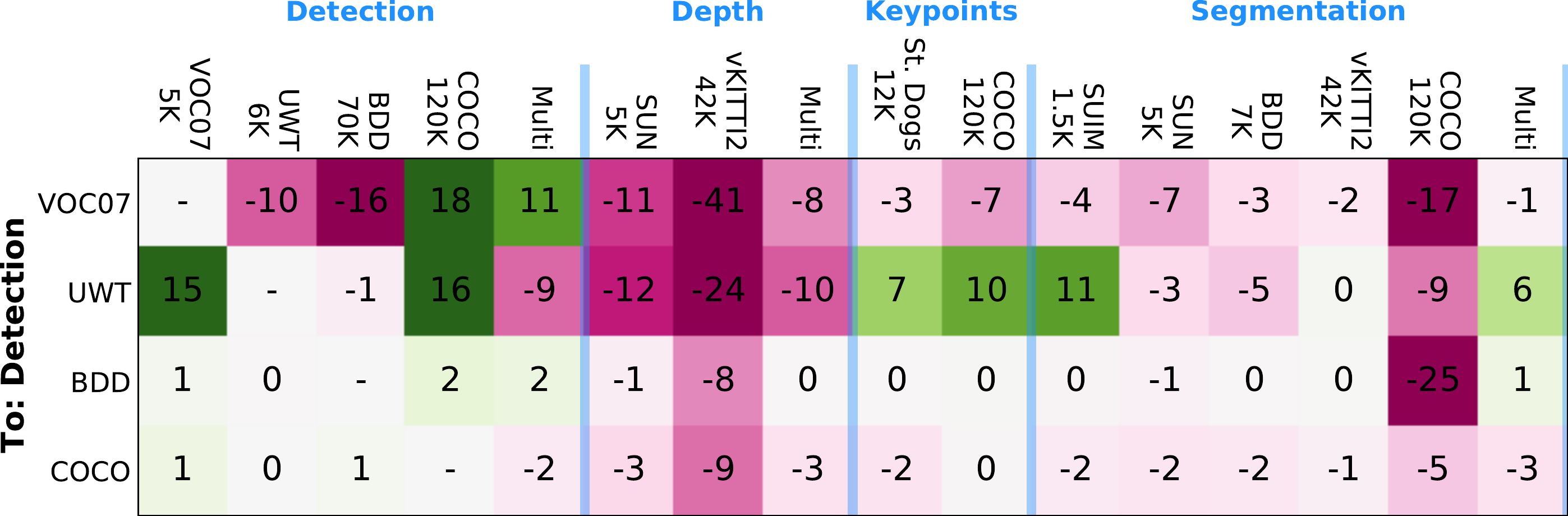}
        \caption{}\label{tab:full_target_detection}
    \end{subtable}
    \begin{subtable}[t]{.59\columnwidth}
        \centering
        \includegraphics[height=24mm]{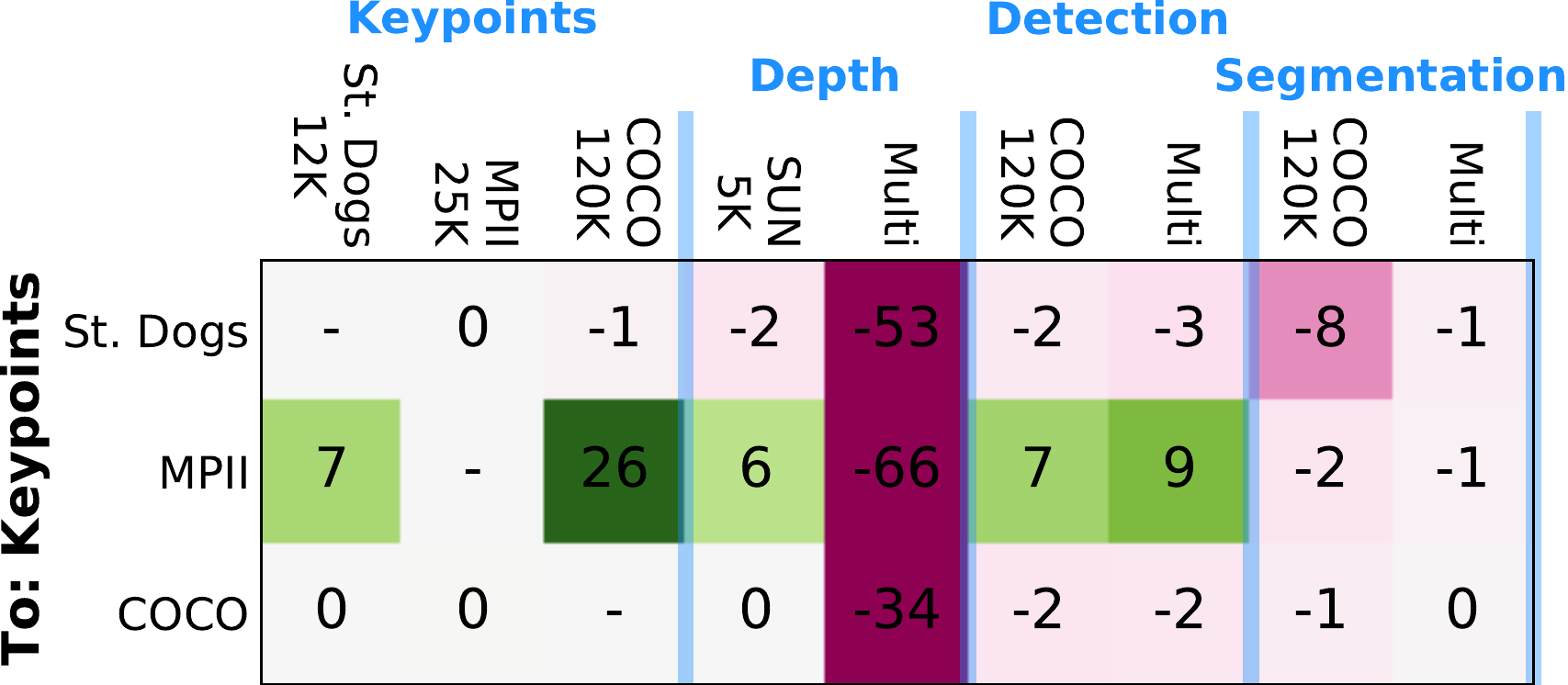}
        \caption{}\label{tab:full_target_keypoints}
    \end{subtable}
    \begin{subtable}[t]{.39\columnwidth}
        \centering
        \includegraphics[height=24mm]{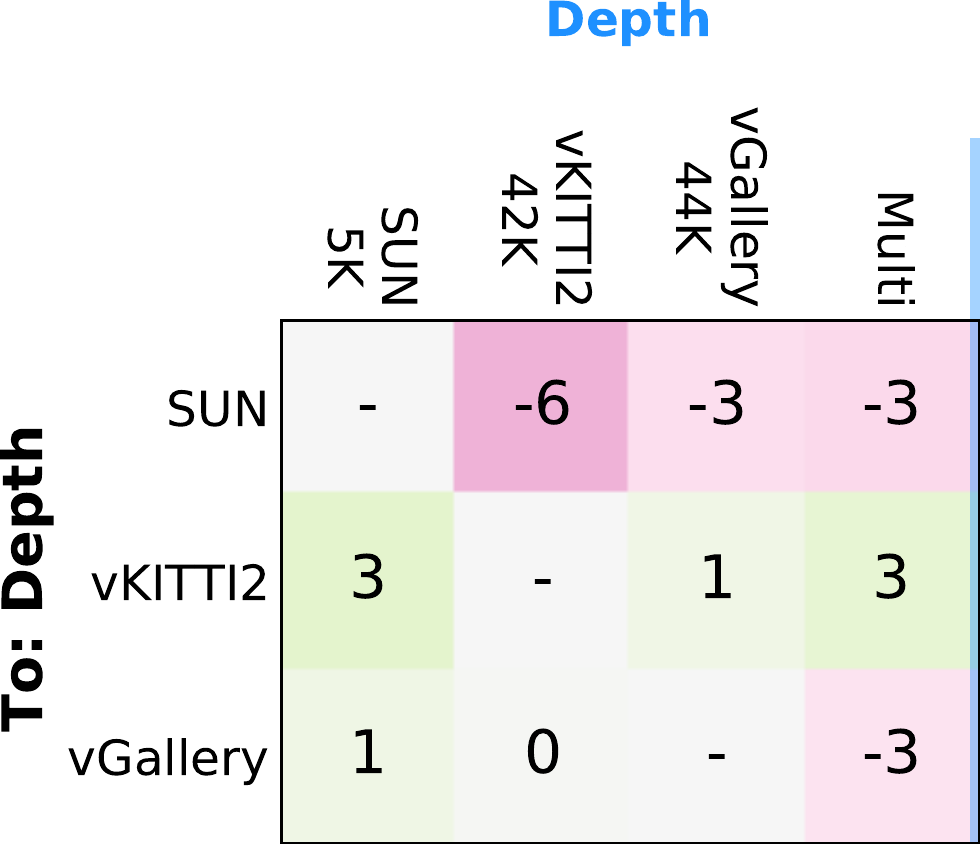}
        \caption{}\label{tab:full_target_depth}
    \end{subtable}
    \caption{Results for transfer learning in the full target training setting.
    Performance measured in \emph{relative improvement gain} over a model fine-tuned from ILSVRC'12.
    In Table (a) both the source and target task types are semantic segmentation.
    See main text for details.
    }    
    \label{tab:full_target}
\end{table}

In this setting we use the full available target training set. We expect transfer learning to have a lesser effect, given that the target training set contains more information.

We perform a study similar to~\autoref{sec:limited_target}, but focus more on transfer within a task type.
The results are shown in~\autoref{tab:full_target}, again grouped by respectively target task type, source task type, image domain, and source size. Absolute numbers can be found in~\appendixsecref{sec:appendix_results}.

\subsection{Small source and small target training set}\label{sec:limited_source}
\begin{table}[tp]
    \includegraphics[width=\columnwidth]{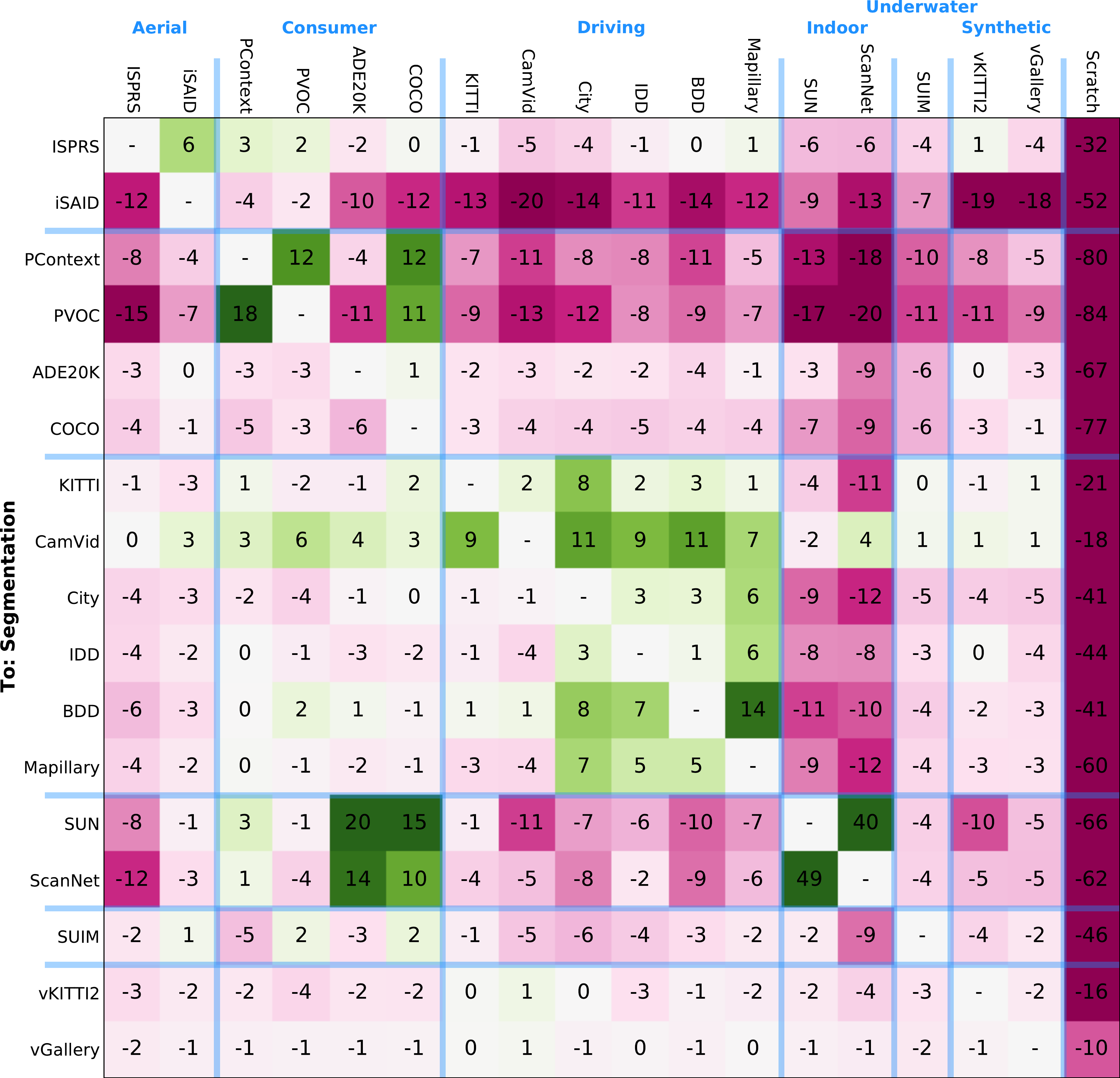}
    \caption{
    Results for transfer learning in the small source and small target setting. All results transferring within semantic segmentation.
    Performance measured in \emph{relative improvement gain} over a model fine-tuned from ILSVRC'12.
    See main text for details.
    }
    \label{tab:limited_source_semseg_within}
\end{table}

In this last setting we limit the size of the target training set as in~\autoref{sec:limited_target}, and also limit the size of each source training set to 1500 samples. This enables to study transfer learning effects where sources differ in appearance domain, but not in the amount of labeled images available. For all except two datasets,
this implies a reduction to the number of images available (KITTI and CamVid have only respectively 150 and 367  training images overall).

We study transfer learning in this setting only for the semantic segmentation task type and only for transfer within-task.
The results are shown in~\autoref{tab:limited_source_semseg_within}.

\section[Analysis]{Analysis}
\label{sec:analysis}

In this section we analyse our results across different settings, task types and domains.
\changed{
To facilitate the discussion, we distinguish four levels of relative transfer gains (as defined in~\autoref{eq:relative_transfer_gain}):
\begin{enumerate}
    \item[VP] Very positive transfer effect, when $r > 10$;
    \item[P] Positive transfer effect, when $r > 2$;
    \item[I] Insignificant transfer effect, when $2 \leq r > -2$;
    \item[N] Negative transfer effect when $r < -2$.
\end{enumerate}
We report the percentage of experiments for each level, but we do not report insignificant transfer (I).}
\autoref{tab:meta_analysis} shows the main results, split into the small target training set and full target training set settings, and filtered by whether we transfer within/across image domain and within/across task type.

\para{A1: Classic ILSVRC'12 transfer learning always outperforms training a model from scratch.}
For \emph{all} our experiments starting from ILSVRC'12 outperforms training from scratch. And by a large margin: even when using the full target training set, transferring from ILSVRC'12 improves performance by \numberchanged{$5\%-46\%$} (see \autoref{tab:full_target_semseg_within}, right-most column).
This confirms that ILSVRC'12 pre-training is a solid way of (starting) transfer learning, which explains why this practice is widespread.

\para{A2: For most target tasks there exists a source task which brings further benefits on top of ILSVCR'12 pre-training.}
This can be seen in~\autoref{tab:few_shot} and~\autoref{tab:full_target}, by noticing that for almost any row there are green entries, indicating positive relative transfer gain over ILSVRC'12 for that source model.

To quantify this observation, we have computed the number of target tasks for which there is a source leading to positive (P) or very positive (VP) transfer effect on top of ILSVRC'12 pre-training.
To do this, for each target task we take the transfer gain brought by the best source (\autoref{tab:meta_analysis_argmax}).
In the small target training set regime, there is a positive effect for \numberchanged{85\%} of the target tasks, and very positive for \numberchanged{67\%}.
Even in the full target training set regime, for \numberchanged{56\%} of the target tasks there is a positive effect, and very positive for \numberchanged{19\%}.

So although ILSVRC'12 pre-training is the de-facto standard way to do transfer learning, there is surprisingly much to gain from an additional transfer step, and this for all task types. Next, we analyze what are the factors that influence these benefits.

\begin{table*}[ht]
\centering
\resizebox{\linewidth}{!}{
\begin{tabular}{ll|cccc|cccc}
\toprule
\midrule
& & \multicolumn{4}{c|}{Small target training set} & \multicolumn{4}{c}{Full target training set} \\
\midrule
image domain & task type & P & VP & N & \#  & P & VP & N & \#   \\
transfer & transfer & $r >2\%$ &  $r >10\%$ & $r < -2\%$ & & $r >2\%$ & $r >10\%$ & $r < -2\%$\\
\midrule
all & all & 21\% & 11\% & 56\% & 509 & 14\% & 3\% & 41\% & 382 \\
within & within & 69\% & 44\% & 17\% & 64 & 33\% & 9\% & 22\% & 78 \\
within & cross & 43\% & 22\% & 37\% & 49 & 22\% & 6\% & 39\% & 18 \\
cross & within & 14\% & 5\% & 55\% & 242 & 8\% & 1\% & 43\% & 242 \\
cross & cross & 5\% & 2\% & 79\% & 154 & 7\% & 2\% & 64\% & 44 \\
\bottomrule
\end{tabular}
}
\caption{Percentage of experiments for which we measured positive/negative transfer effects of a certain magnitude.
We show these for the small target training set setting (i.e. measured over all source-target pairs from all tables in \autoref{tab:few_shot}) and full target training setting (measured over tables in \autoref{tab:full_target}).
We aggregate these under different image domain and task type transfer constraints. $\#$ denotes the number of experiments considered according to the constraint.}
\label{tab:meta_analysis}
\end{table*}

\begin{table*}[ht]
\centering
\resizebox{\linewidth}{!}{
\begin{tabular}{ll|cccc|cccc}
\toprule
\midrule
& & \multicolumn{4}{c|}{Small target training set} & \multicolumn{4}{c}{Full target training set} \\
\midrule
image domain & task type & P & VP & N & \#  & P & VP & N & \#   \\
transfer & transfer & $r >2\%$ &  $r >10\%$ & $r < -2\%$ & & $r >2\%$ & $r >10\%$ & $r < -2\%$\\
\midrule
all & all & 85\% & 67\% & 4\% & 27 & 56\% & 19\% & 7\% & 27 \\
within & within & 73\% & 64\% & 14\% & 22 & 50\% & 15\% & 12\% & 26 \\
within & cross & 65\% & 35\% & 18\% & 17 & 33\% & 17\% & 0\% & 6 \\
cross & within & 56\% & 30\% & 26\% & 27 & 30\% & 4\% & 19\% & 27 \\
cross & cross & 32\% & 11\% & 53\% & 19 & 29\% & 14\% & 14\% & 7 \\
\midrule
\bottomrule
\end{tabular}
}
\caption{Transfer effects induced by the best available source for each target task. We show these for the small target training setting (i.e. max over each row of the tables in \autoref{tab:few_shot}) and full target training setting (i.e. max over rows in \autoref{tab:full_target}).
We aggregated these under different image domain and task type transfer constraints. $\#$ denotes the number of target tasks considered according to the constraints.}
\label{tab:meta_analysis_argmax}
\end{table*}

\para{A3: The image domain strongly affects transfer gains.}
From~\autoref{tab:few_shot_semseg_within} we see that most positive gains occur when the source and target tasks are in the same image domain (within-domain transfer). Conversely, often transfer across image domains yields negative gains. For example, for the consumer datasets as target, all other domains are bad sources. This makes the image domain an important factor.
As can be seen, the size of the source dataset also plays a role: for example, for \emph{driving} as a target domain, the larger driving and consumer datasets are generally better sources than the smaller ones. However, this effect  is far less important than the image domain.
Finally,~\autoref{tab:few_shot_semseg_cross} shows transfer from other task types to segmentation (cross-task-type transfer). Here, most effects are negative.

To quantify, we first compare the individual effects of image domain and task type in~\autoref{tab:meta_analysis}. 
Out of all experiments with small target training set, within-domain, cross-task-type setting, \numberchanged{43\%} yields positive transfer gains and \numberchanged{37\%} negative ones.
Conversely, out of all experiments in the cross-domain, within-task-type setting, only \numberchanged{14\%} yields positive transfer gains while \numberchanged{55\%} negative ones.
This pattern is repeated when using the full target training set.
Hence we conclude that, given a target task, the image domain of the source is more important to achieve good transfer gains than its task type.

\begin{table}
    \centering
    \includegraphics[width=\columnwidth]{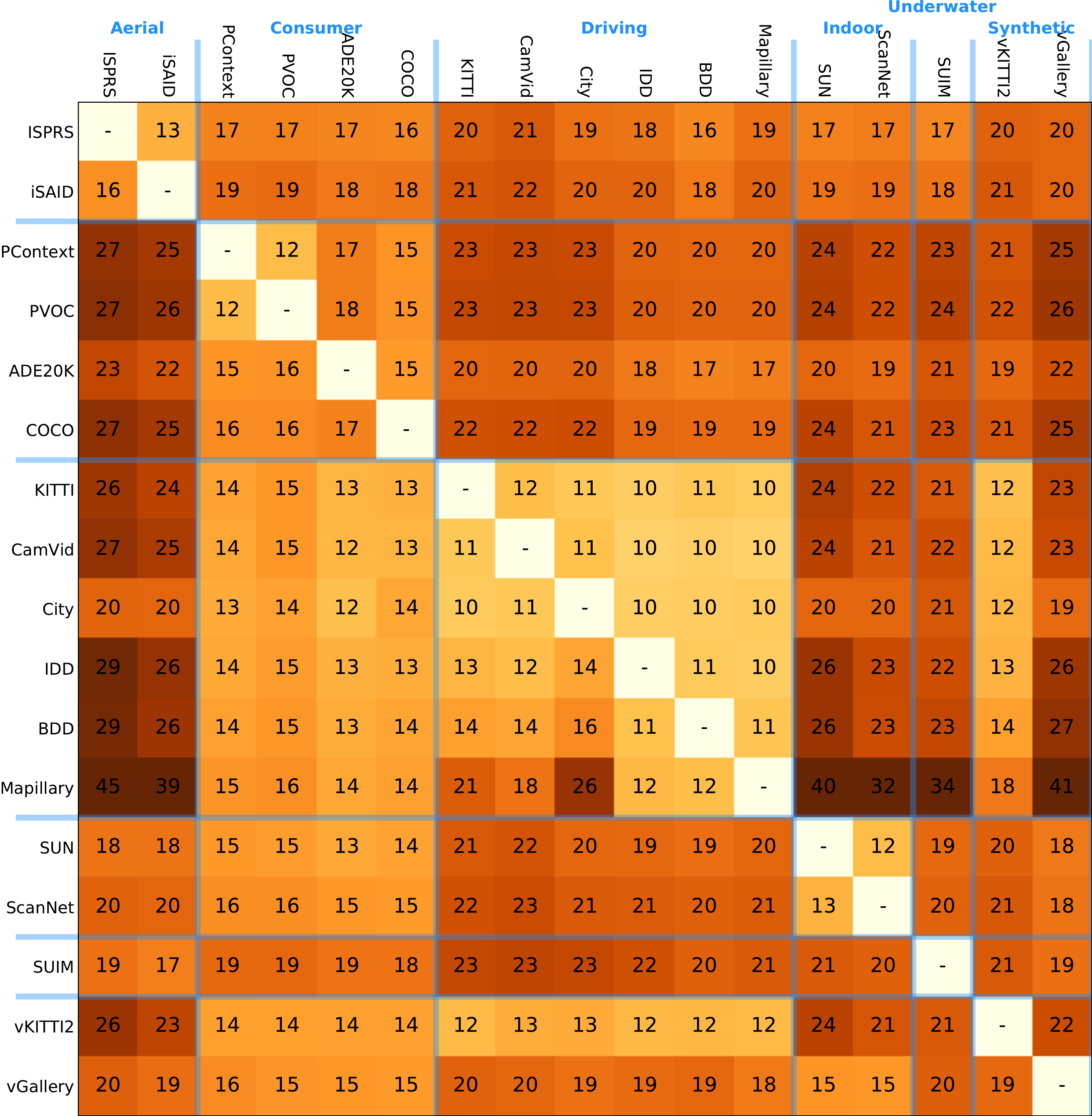}
    \caption{Domain distance between datasets (\autoref{eq:domain_difference}, asymmetric). Rows are targets, columns are sources. Lowest distance is within a manually defined image domain, as expected. Consumer to driving and indoor domains yield close distance values, but not vice versa. This shows that consumer images \emph{include} both the driving and indoor domains.}
    \label{tab:domain_distance_segmentation}
\end{table}

Above we considered image domains as manually defined, intuitive types.
As presented in \autoref{sec:image_domain}, we also consider a continuous measure of domain distance based on image appearance.
~\autoref{tab:domain_distance_segmentation} visualizes the domain distance for semantic segmentation datasets, with lighter colors indicating a smaller distance. This appearance distance correlates with the manually defined domains. One exception is the vKITTI2 dataset, which seems to be closer to other driving datasets than to the other synthetic dataset vGallery.

With such a continuous measure of domain distance we can compare its influence on transfer gains to the influence of source size.
We first measure the rank correlation between domain distance and transfer gains using Kendall $\tau$~\cite{kendall38biometrika}. For the small target training set experiments this yields a correlation of \numberchanged{0.40},
while the correlation between source size and transfer gains is \numberchanged{0.12}. Therefore the image domain is a more important factor of influence than source size.

\para{A4: For positive transfer, the source image domain should include the target domain.}
In most of our experiments a source from a broader domain helps a target from a more specific domain contained within it.
This can be qualitatively observed from~\autoref{tab:few_shot_semseg_within}.
For example, sources from the consumer domain achieve positive transfer not only on consumer targets, but also on driving or indoor. 
Indeed, the consumer datasets also contain images of street views and indoor scenes (\autoref{fig:total_transfer_datasets_illustration}). 
Conversely, if we look at the consumer domain as a target, \emph{none} of the driving or indoor sources yield any positive transfer.
The same pattern can also explain the success of the multi-source model: this model always yields positive transfer, and by design it contains images from each of the manually defined target domains. 

To visualize domain inclusion, we created a t-SNE plot of the activations of 150 images from each semantic segmentation dataset (\autoref{fig:tsne_semseg}; same features as~\autoref{sec:image_domain}).
The visualization show that the consumer domain points are indeed scattered around the plot, covering almost all other domains. Other domains form more compact clusters, \eg aerial in green, and driving in yellow/orange. 
\newcommand{\thightfbox}[1]{{
    \setlength{\fboxsep}{0pt}
    \fbox{#1}
    }}
\begin{figure*}
    \centering
    \begin{subfigure}[T]{.66\textwidth}
        \thightfbox{\includegraphics[height=112mm]{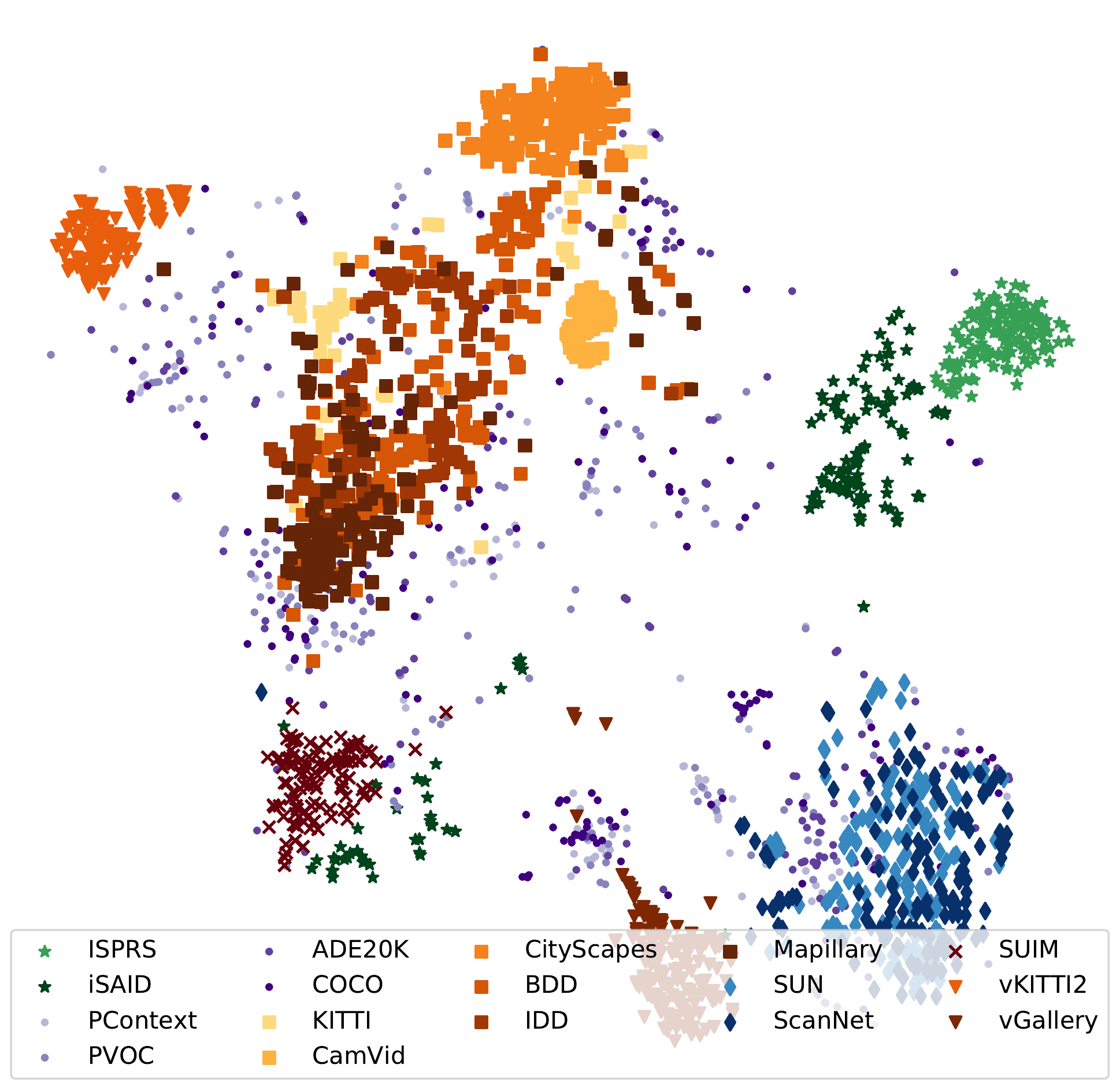}}
        \caption{Multi source network features for all semantic segmentation datasets.}
        \label{fig:tsne_semseg}
    \end{subfigure}
    \hspace{-1em}
    \begin{subfigure}[T]{.32\textwidth}
        \begin{subfigure}[T]{\textwidth}
            \thightfbox{\includegraphics[height=53mm]{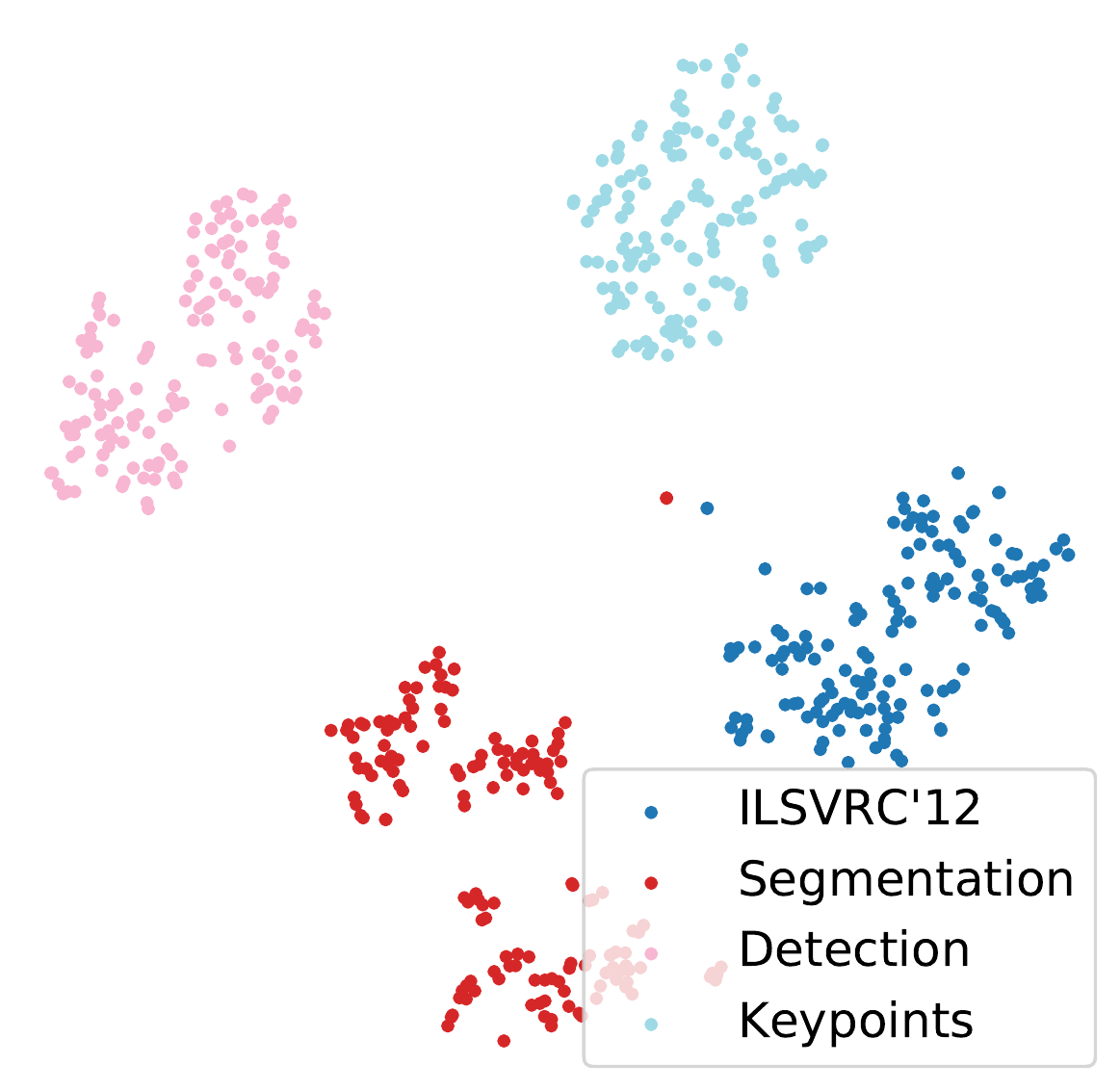}}
            \caption{Different networks on COCO.}
            \label{fig:tsne_coco}    
        \end{subfigure}
        
        \begin{subfigure}[T]{\textwidth}
            \thightfbox{\includegraphics[height=53mm]{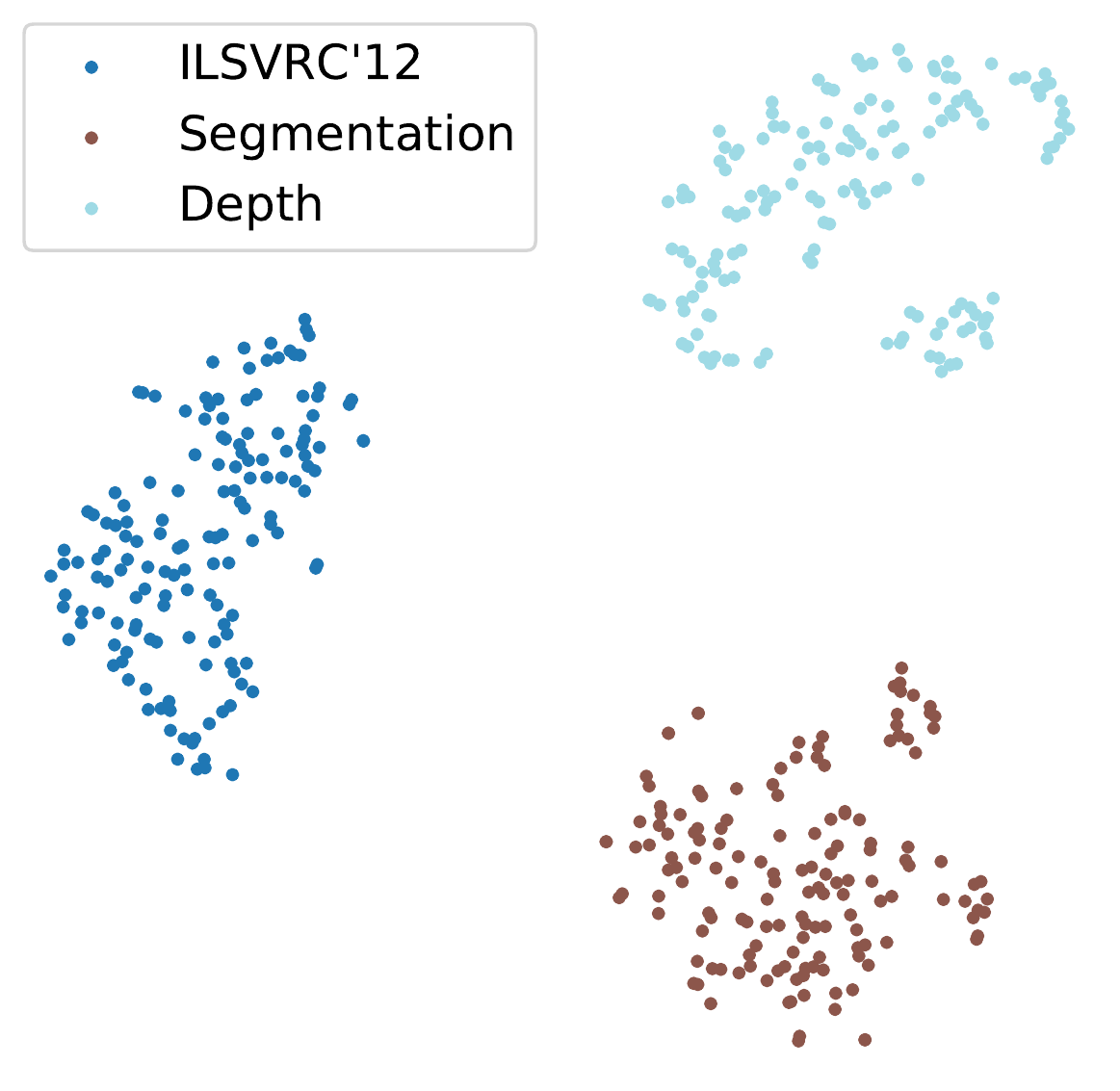}}
            \caption{Different networks on SUN RGB-D.}
            \label{fig:tsne_sunrgbd}    
        \end{subfigure}
    \end{subfigure}
    
    \caption{%
    Illustration of the apperance distribution of the datasets, according to different feature encoding networks, using t-SNE. 
    \autoref{fig:tsne_semseg} visualizes the semantic segmentation datasets using features extracted from the multi-source network. 
    \autoref{fig:tsne_coco} and~\autoref{fig:tsne_sunrgbd} visualizes the COCO and SUN RGB-D datasets using networks trained on different task types (ILSVRC'12 indicates classification).
    }
    \label{fig:tsne}
\end{figure*}

\begin{table}
\centering
\resizebox{0.8\columnwidth}{!}{
\begin{tabular}{lc}
Assignment method & $\tau$ \\
\midrule
(1) Earth Mover's Distance & 0.20 \\
(2) Target image $\rightarrow$ closest source image & 0.42  \\
(3) Source image $\rightarrow$ closest target image & 0.06  \\
(4) Source $\rightarrow$ Target \& Target $\rightarrow$ Source & 0.27\\
\bottomrule
\end{tabular}
}
\caption{Kendall-$\tau$ correlation between transfer gains and dataset distance for semantic segmentation in the small source small target setting (\autoref{sec:limited_source}). As distance we take the average distance between images from the source and target datasets, while varying the assignment method (\ie $\mathrm{min}$ function in \autoref{eq:domain_difference}). The assignment with the highest correlation measures \emph{inclusion} of the target by the source.}
\label{tab:correlation_image_measures}
\end{table}

Finally, we also quantitatively demonstrate that domain inclusion matters. 
To do so, we change the function used in~\autoref{eq:domain_difference} to match target images to source images. We have four strategies:
(1) Assign one-to-one each target image to a source image such that the overall Earth Mover's Distance is minimized (\ie matching is done through the Hungarian algorithm). This is a measure of overlap of the distributions of the two image domains.
(2) Assign each target image to its closest source. This measures \emph{inclusion} of the target dataset in the source dataset and is identical to \autoref{eq:domain_difference}.
(3) Assign each source image to its closest target. A large distance here means that many source images are far from the target.
(4) Take the average of (2) and (3), yielding a symmetric measure.

We correlate each of the four measures to transfer gains using Kendall-$\tau$ rank correlation. 
To remove any influence of task type and dataset size (both source and target), we do this in the small source and small target set setting (\autoref{sec:limited_source}).
The results are shown in~\autoref{tab:correlation_image_measures}.
The assignment measure capturing inclusion (2) has the highest correlation with transfer gains.
Interestingly, measure (3) has (very) low correlation. This suggests that even if the source domain has images far from the target domain, this has only a small influence on transfer gains.
This confirms that the most important aspect for a source to yield positive transfer is that is should \emph{include} the target image domain.

\para{A5: Multi-source models yield good transfer, but are outperformed by the largest within-domain source.} 

\autoref{tab:few_shot_semseg_within} shows that our multi-source semantic segmentation model yields positive transfer gains for all targets. This is in line with our previous observations that the source domain should include the target, while it is less important that it spans a much broader range. Indeed, by construction the source data of this multi-source model spans all domains. However, we also observe that the largest within-domain source almost always yields better transfer gains: In \numberchanged{7} out of these 10 experiments the largest within-domain source yields significantly better transfer (i.e. $>2\%$). Only for the target SUN RGB-D the multi-source model is better. This is explainable since the multi-source model includes both COCO (consumer) and ScanNet (indoor) which are both good sources for SUN RGB-D. Hence the multi-source model arguably covers the domain of this dataset better than the largest single source model.

\begin{table}
\centering
\resizebox{\columnwidth}{!}{
\begin{tabular}{l|ccc|ccc}
\toprule
& \multicolumn{3}{c|}{same dataset} & \multicolumn{3}{c}{same domain} \\
\cmidrule{2-7}
& P & N & \# & P & N & \# \\
\midrule
seg $\rightarrow$ obj det & 100\% & 0\% & 2 & 63\% & 25\% & 8 \\
seg $\rightarrow$ kp det & 0\% & 100\% & 1 & 0\% & 100\% & 2 \\
seg $\rightarrow$ depth & 33\% & 0\% & 3 & 33\% & 33\% & 3 \\
obj det $\rightarrow$ seg & 0\% & 0\% & 1 & 30\% & 50\% & 10 \\
obj det $\rightarrow$ kp det & 100\% & 0\% & 1 & 100\% & 0\% & 2 \\
kp det $\rightarrow$ seg & 0\% & 100\% & 1 & 0\% & 100\% & 3 \\
kp det $\rightarrow$ obj det & 100\% & 0\% & 1 & 0\% & 33\% & 3 \\
depth $\rightarrow$ seg & 100\% & 0\% & 1 &  100\% & 0\% & 1 \\
\bottomrule
\end{tabular}
}
\caption{Percentage of positive and negative transfer experiments in the cross-task-type, small target setting. Semantic segmentation is denoted as `seg', object detection as `obj det', keypoint detection as `kp det', depth estimation as `depth'.
We consider the setting where the images of the source are \emph{the same} (dataset) as the target, and when the source and target share the \emph{same domain} but are different datasets.
Note that the two experiments where segmentation to object detection fails for the same domain, are for Underwater Trash as a target. Arguably, this is a rather different underwater domain then that of SUIM which acts as the source (once alone, once as a multi-source model).
}
\label{tab:cross_task_transfer}
\end{table}

\para{A6: Transfer across task types can bring positive transfer gains.}

Depending on the choice of task types for the source and target, cross-task-type transfer can be beneficial: for \numberchanged{65\%} of the targets within the same image domain as the source, cross-task-type transfer results in positive transfer gains (\autoref{tab:meta_analysis_argmax}).
To study this effect more precisely, we split the results over different cross-task-type pairs in~\autoref{tab:cross_task_transfer}.
The left column looks at cross-task-type transfer \emph{on the same dataset} (the specific setting studied in detail in Taskonomy~\cite{zamir18cvpr}). Here we can see that object detection and keypoint detection help each other. Segmentation helps object detection but not vice versa. Semantic segmentation and keypoint detection hurt each other. When transferring to a different dataset, but still saying in the same image domain (right column), results show similar effects (except that keypoint detection stops helping object detection).

Our results confirm some of the observations in~\cite{zamir18cvpr} as we also observe that some task types are more easily to transfer from / to than others, and that transfer across task types is asymmetric, \ie if task type A is beneficial for task type B, the reverse might not hold.

However, even when we observe gains through cross-task-type transfer within \emph{the same dataset}, if there is another within-domain source with the same task type, it yields even better gains. Two examples are the object detection results for BDD and COCO as targets in \autoref{tab:few_shot_detection}. This suggests that having a source with the same image domain and task type as the target is better than having annotations for another task type on the target dataset.

\autoref{tab:meta_analysis} shows that for cross-task-type transfer to work, the image domain of source and target should be the same: cross-domain, cross-task transfer yields negative transfer for \numberchanged{79\%} of the experiments, and positive for \numberchanged{5\%}. Moreover, out of those \numberchanged{5\%}, all except one experiment have consumer images as a source which includes the target domain of driving, indoor, and (arguably) underwater. Hence cross-task-type transfer can only be expected to work if the image domain from the source includes that of the target.

To understand how the task type influences the features learned by the source models, we visualize the features learned on the same dataset but for different task types using t-SNE (cosine distance, which removes arbitrary scaling effects). We do this for the exact same set of images for COCO (\autoref{fig:tsne_coco}) and SUN RGB-D (\autoref{fig:tsne_sunrgbd}). We observe that each model creates compact cluster representations, showing there is a larger distance between an identical image in the feature spaces of two different models than between two different images in the feature space of one model.

\para{A7: Transfer within-task-type and within-domain yields very positive effects.} 
Combining the observations A3 and A6, our recommendation to obtain positive transfer is to use a source model trained on the same domain and for the same task type as the target task. 
In this setting, a total of \numberchanged{69\%} of all source-target pairs exhibits positive transfer (\autoref{tab:meta_analysis}).
Moreover, the best available source in this setting leads to positive transfer for \numberchanged{73\%} of the target tasks (and \numberchanged{64\%} very positive).

\para{A8: Transfer naturally flows from larger to smaller datasets.}
To study the effect of dataset size, we look at transfer learning within-domain and within-task-type. In this setting, the transfer effect is best visibile in~\autoref{tab:few_shot_semseg_within} and ~\autoref{tab:full_target_semseg_within}, when looking at the blocks around the diagonal. Within each block of datasets of the same domain, they are ordered by training set size. We observe a clear \emph{upper-triangle} pattern, where the larger datasets positively transfer towards the smaller datasets. The reverse is much less present.

Quantitatively, we measure a positive transfer in \numberchanged{60\%} of within-domain and within-task-type experiments where the source training set is larger than the  target dataset. 
Conversely, when the source training set is smaller than the target training set, only \numberchanged{5\%} of our experiments result in positive transfer.

\para{A9: Transfer learning effects are larger for small target training sets.}
We expect this to hold because when the target task already offers a large training set, the target model can learn the required visual knowledge from the target task directly. We can clearly see such effects by comparing the within-domain segmentation results of the small and full target training set setting, \ie \autoref{tab:few_shot_semseg_within} vs \autoref{tab:full_target_semseg_within}. 
The absolute transfer gains are higher in the small target training set setting than in the full one, while overall patterns remain roughly the same. 

We now compare \autoref{tab:few_shot_semseg_within} and \autoref{tab:full_target_semseg_within} quantitatively. We first examine positive transfer gains: 
out of all experiments with positive transfer gains in the full target setting, \numberchanged{80\%} also have a positive transfer in the small target setting.
Vice versa, out of all experiments with positive transfer gains in the small target setting, 43\% also have significant gains in the full target setting (for most of these experiments the gains become insignificant).
When looking at negative transfer effects: if transfer effects are negative on a small target training set, these effects remain negative or become neutral in the full target set in 100\% of these experiments.

Practically, this suggests a quick test for whether a source dataset may be beneficial for a target.
First train on a small subset of the target training set. 
Then, if transfer effects are negative, discard this source for this target.
If transfer effects are positive, explore this source further, since there is a good chance its benefits are kept when using the full target training set.

\para{A10: The source domain including the target is more important than the number of source samples.}
Here we aim to disentangle the size of the source training set from the breadth of the image domain it spans (\eg COCO covers a broad image domain as visualized in \autoref{fig:tsne_coco}).
The influence of source size is eliminated in our experiment in \autoref{tab:limited_source_semseg_within}, where we use a small training set for all sources (and a small target training set too). It is instructive to compare this \autoref{tab:few_shot_semseg_within}, where full source sets are used.

Even after fixing source size, most patterns stay roughly the same. For example, COCO remains a good source for other consumer and indoor datasets. Indeed, COCO is not only the largest but also the most diverse consumer dataset.
Similarly, Mapillary remains a good source for most other driving datasets, while other driving sources are generally worse. Mapillary was designed to capture driving conditions around all continents, whereas the India Driving Dataset covers only Indian roads, Berkeley Deep Drive is US only, while CityScapes is (mostly) Germany. Camvid consists of frames of four videos, and therefore has a narrow image domain.
This suggests that much of the effects in \autoref{tab:few_shot_semseg_within} which we attributed to dataset size in A8, can in fact be attributed to the property of the source domain \emph{including} the target domain.
Of course, source size is generally correlated to the breadth of its domain (e.g. COCO and Mapillary were both designed to span broader domains), which is why care needs to be taken to disentangle their effects.

\section{Additional Scenarios}\label{sec:generalization_experiments}

In this section we perform several additional experiments to verify whether our work generalizes to other scenarios. 
We only report the main observations while details are provided in~\appendixsecref{sec:detailed_generalization_experiments}.

\para{Fixed image resolution.} 
We redo the segmentation experiments in the small target training setting (\ie~\autoref{tab:few_shot_semseg_within}) but now fixing the image resolution to be $713 \times 713$ pixels across all datasets. Results in~\appendixref{tab:tt_segmentation_raw_within_fixed_crop} show highly similar transfer patterns and all our previous conclusions hold.

\para{ResNet50.} We change the backbone to ResNet50~\cite{he16cvpr} and redo the experiments in the small target training setting for segmentation  and partially for detection (\ie redo~\autoref{tab:few_shot_semseg_within} partially~\autoref{tab:few_shot_detection}). Results in~\appendixref{tab:tt_segmentation_raw_within_resnet_full} suggest that ResNet50 benefits more from transfer learning, possibly due to improving its ability for localized predictions. But again, we observe similar transfer patterns and all our previous conclusions hold.

\para{Self-supervised ILSVRC'12 training.} We now redo the experiments with ResNet50, but instead we start from a checkpoint which was obtained by using self-supervised learning on ILSVRC'12 using the publicly available SimCLR V2 implementation~\cite{chen20icml}. We then train our sources fully supervised as before. Results in~\appendixref{tab:tt_segmentation_raw_within_resnet_ss} show primarily changes in the baseline: directly training from self-supervised ILSVRC'12 weights is better than directly training from ILSVRC'12 fully supervised classification weights, as also found in~\cite{ericsson21cvpr,zoph20nips}. At the same time patterns are again very similar and our main conclusions remain unaltered.

\para{Self-supervised Transfer Chain.}
We do a single experiment where we use SimCLR~\cite{chen20icml} to create a self-supervised transfer chain: ILSVRC'12 self-supervised $\rightarrow$ COCO self-supervised $\rightarrow$ target. We find that this transfer chain is mildly beneficial for COCO \emph{image classification}. However, for segmentation our results in~\appendixref{tab:tt_segmentation_raw_within_resnet_ss_chain} show that that this transfer chain is worse than directly training from ILSVRC'12 self-supervised weights for \emph{all} target datasets. On average it is 0.07 IoU worse, while even for COCO as a target results are 0.02 IoU worse.
This result suggests that self-supervised pre-training is biased towards image classification. Furthermore, image classification results on self-supervised models are not very predictive of performance on other tasks, as also shown in the dedicated study of~\cite{ericsson21cvpr}.

\section{Conclusion}
\label{sec:conclusion}

In this paper we performed over \changed{2000 transfer learning experiments} across a wide variety of image domains and task types. Our systematic analysis of these experiments lead to the following conclusions:
(1) for most tasks there exists a source which significantly outperforms ILSVRC'12 pre-training;
(2) the image domain is the most important factor for achieving positive transfer;
(3) the source task should \emph{include} the image domain of the target task to achieve best result;
(4) at the same time, we observe only small negative effects when the image domain of the source task is much broader than that of the target;
(5) transfer across task types can be beneficial, but its success is heavily dependent on both the source and target task types.

Our findings provide support for the success of large-scale pre-training for transfer learning, with a single very large source spanning a mixture of domains~\cite{joulin16eccv,kolesnikov20eccv,mahajan18eccv,sun17iccv,zhai19arxiv}. If a good source set should include the target but can be arbitrarily broad, training on a massive dataset which covers most of the visual domains should yield a good source model for most target tasks. However, this form of pre-training inevitably requires transfer across task types, whose success depends on both the source and target task types.
This suggests that future works exploring pre-training should focus also on structured prediction tasks.

{\small
\bibliographystyle{abbrv}
\bibliography{shortstrings,loco,loco_extra}
}

\begin{IEEEbiography}[{\includegraphics[width=1in,height=1.25in,clip,keepaspectratio]{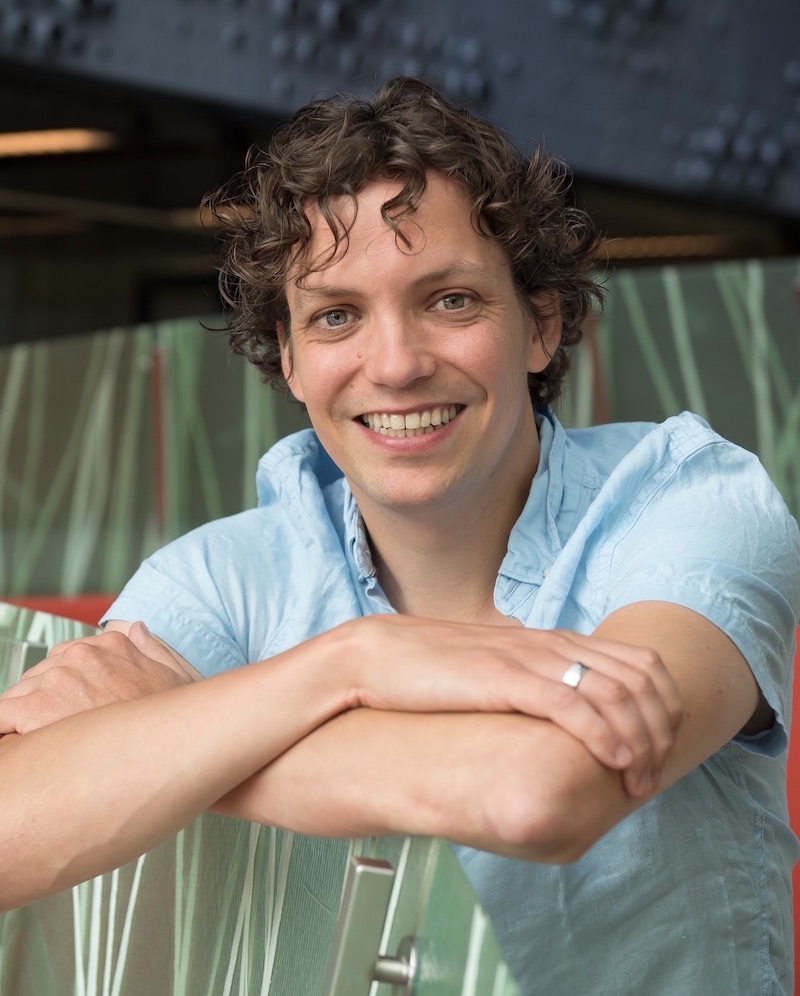}}]{Thomas Mensink}
is a Research Scientist at Google since 2019. 
He obtained his PhD from the University of Grenoble in 2012, working at the LEAR team of INRIA and at Xerox Research Centre Europe. 
He worked as Assistant Professor (2017-2019) and post-doc (2012-2017) at University of Amsterdam.
His research interest include transfer learning for dense prediction tasks and learning visual representations for zero-shot reasoning. 
He has received multiple best paper awards (ACM Multimedia 2014, ICMR 2016), a VENI grant (NWO 2015), and the Koenderink Prize for fundamental contributions in computer vision (ECCV 2020). 

\end{IEEEbiography}

\begin{IEEEbiography}[{\includegraphics[width=1in,height=1.25in,clip,keepaspectratio]{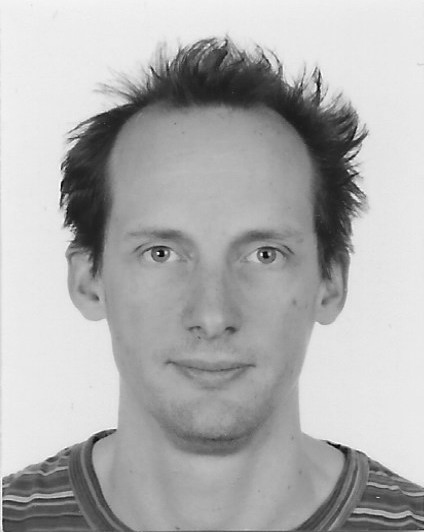}}]{Jasper Uijlings}
is a Research Scientist at Google since 2016, working in a team led by Vittorio Ferrari. His research topics are on lifelong learning and human-machine collaboration for image annotation. Before that, he worked as a post-doc at the University of Edinburgh with Prof. Vittorio Ferrari and at the University of Trento with Prof. Nicu Sebe. He obtained his PhD in 2011 at the University of Amsterdam under supervision of Prof. Dr. Ir. Remko Scha and Prof. Dr. Ir. Arnold Smeulders.
\end{IEEEbiography}

\begin{IEEEbiography}[{\includegraphics[width=1in,height=1.25in,clip,keepaspectratio]{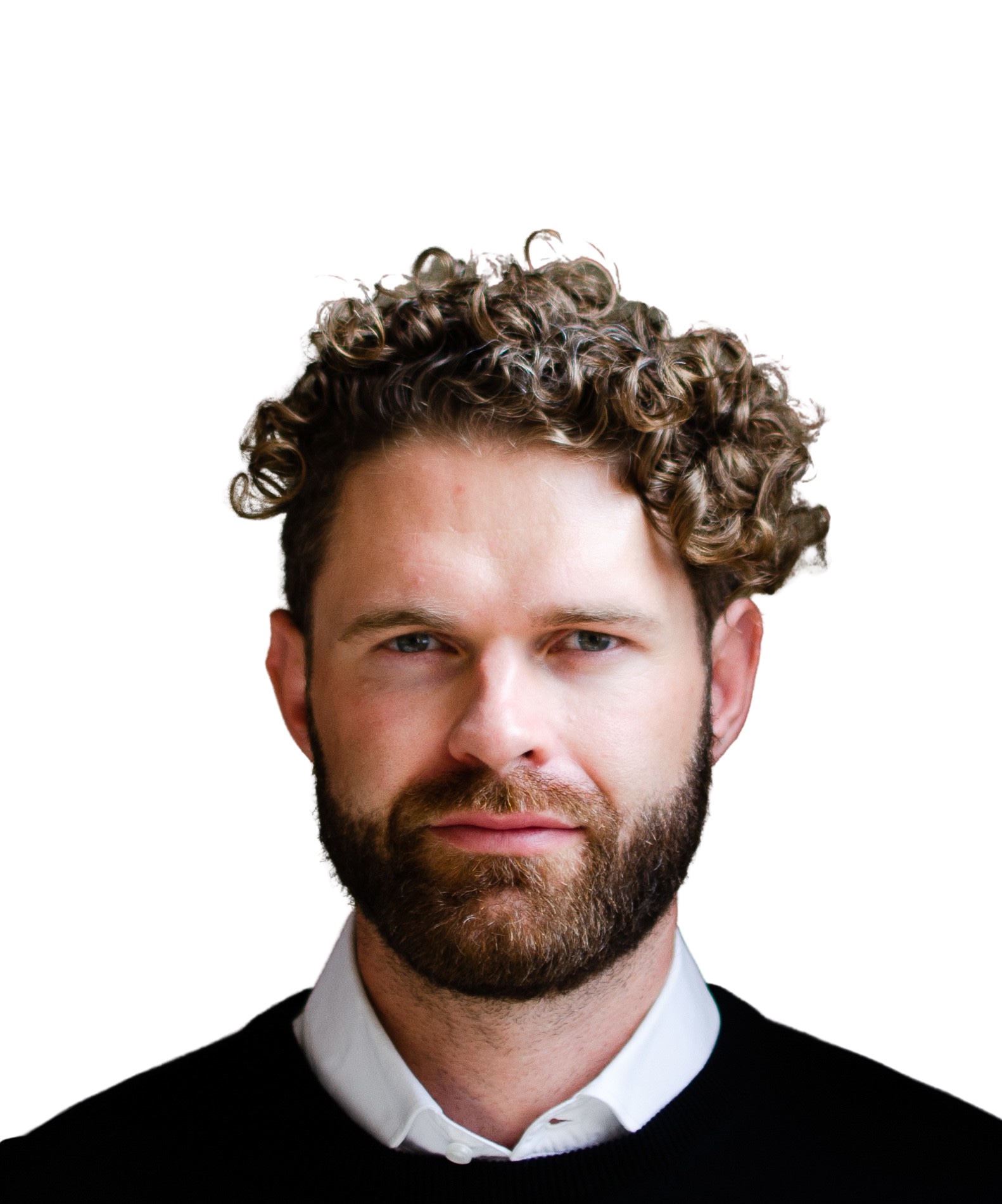}}]{Michael Gygli}
received his PhD from ETH Zurich in 2017. He is a technical co-founder of Cerrion, which uses Computer Vision to increase the stability of manufacturing processes. He is also an experienced computer vision researcher. During this PhD, his time as head of AI at gifs.com and as a Research Scientist at Google he has published over 20 research papers at venues such as CVPR, ICML, ICCV, IJCV, AAAI and ACM Multimedia. His main interests are real-time video analysis, image segmentation and transfer learning.
\end{IEEEbiography}

\begin{IEEEbiography}[{\includegraphics[width=1in,height=1.25in,clip,keepaspectratio]{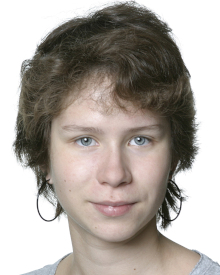}}]{Alina Kuznetsova}
received her Ph.D. from Leibniz University of Hannover in 2016. During her studies, she interned at Disney Research, working with Dr. Sung Ju Hwang and Dr. Leonid Sigal, and at Microsoft Research. She has co-authored over 15 technical publications in various fields of computer vision during her PhD and time in Google Research. Her research interests are large-scale annotation and transfer learning.
\end{IEEEbiography}

\begin{IEEEbiography}[{\includegraphics[width=1in,height=1.25in,clip,keepaspectratio]{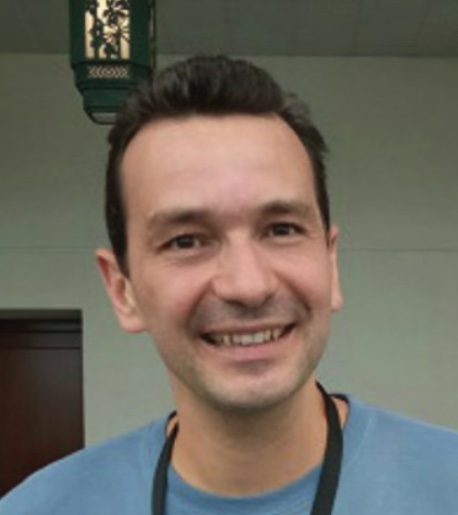}}]{Vittorio Ferrari}
is a Senior Staff Research Scientist at Google, where he leads a research group on visual learning. He received his PhD from ETH Zurich in 2004, then was a post-doc at INRIA Grenoble (2006-2007) and at the University of Oxford (2007-2008). Between 2008 and 2012 he was an Assistant Professor at ETH Zurich, funded by a Swiss National Science Foundation Professorship grant. In 2012-2018 he was faculty at the University of Edinburgh, where he became a Full Professor in 2016 (now a Honorary Professor). In 2012 he received the prestigious ERC Starting Grant, and the best paper award from the European Conference in Computer Vision. He is the author of over 130 technical publications. He regularly serves as an Area Chair for the major computer vision conferences, he was a Program Chair for ECCV 2018 and a General Chair for ECCV 2020. He is an Associate Editor of the International Journal of Computer Vision, and formerly of IEEE Pattern Analysis and Machine Intelligence. His current research interests are in learning visual models with minimal human supervision, human-machine collaboration, and 3D Deep Learning.
\end{IEEEbiography}
\vspace*{\fill}

\iftoggle{includesuppmat}{
\clearpage
\appendices
\newcommand{\appendixtitleauthor}[3]{
\bgroup\par\vskip\IEEEtitletopspace\vskip\IEEEtitletopspaceextra
\begin{center}
    \vskip1.0em{}
    {\Huge\sffamily {#1}\par}\relax
   \vskip1.0em\par
   {\lineskip.5em\sffamily\sublargesize {#2}\par
   }
\end{center}
\egroup}

\newcommand{\appendixabstract}[1]{
\bgroup\par\vskip\IEEEtitletopspace\vskip\IEEEtitletopspaceextra
\begin{center}
    {\vskip 1.5em\relax
        \parbox{0.922\columnwidth}{\footnotesize
        \textbf{Abstract ---} #1}\par\noindent\hfill
        \IEEEcompsocdiamondline\hfill\hbox{}\par
    }
\end{center}
\egroup}

\newcommand{\mainpaperlabel}[1]{%
    {\color{RubineRed}\IfStrEqCase{#1}{%
    {eq:relative_transfer_gain}{Eq 2.}%
    {tab:num_parameter_per_task}{Table 2}%
    {sec:analysis}{Section 6.}%
    {sec:data_normalization}{Section 4.1}%
    {sec:transfer_experiments}{Section 5.}%
    {tab:few_shot_semseg_within}{Table 4.a}%
    {tab:few_shot_detection}{Table 4.c}%
    }[LABEL NOT DEFINED: $#1$]%
    }
}
\def\ifUnDefinedCs#1{\expandafter\ifx\csname#1\endcsname\relax} 
\newcommand{\mainpaperref}[1]{\ifUnDefinedCs{r@#1}{\mainpaperlabel{#1}}\else\autoref{#1}\fi~(in main paper)}

\renewcommand{\figureautorefname}{Figure}%
\renewcommand{\tableautorefname}{Table}%
\def\equationautorefname~#1\null{Eq.~(#1)\null} %
\renewcommand{\sectionautorefname}{Section} %
\renewcommand{\subsectionautorefname}{\sectionautorefname} %

\renewcommand{\thesection}{\Alph{section}}
\renewcommand{\thefigure}{A.\arabic{figure}}
\renewcommand{\theequation}{A.\arabic{equation}}
\renewcommand{\thetable}{A.\arabic{table}}

\iftoggle{includesuppmat}{
    \setcounter{table}{0}
    \setcounter{figure}{0}
    \setcounter{equation}{0}
    \appendixtitleauthor{Appendices}{}
    \appendixabstract{
In these appendices we mainly provide more details on the network architectures in \appendixsecref{sec:appendix_networks}, generalization experiments in \appendixsecref{sec:appendix_generalization}, and more extensive experimental results in \appendixsecref{sec:appendix_results}. For all experiments not only the relative transfer gain is shown, but also the absolute value of the task-specific performance metric is shown. In \appendixsecref{sec:computational_cost} we discuss the additional computation costs of transfer chains. Finally, in~\appendixsecref{sec:data_overlap} we discuss potential data overlap in our collection of datasets.}
}{}

\section{Backbone and Task Type Specific Network Architectures}
\label{sec:appendix_networks}
In this section we describe in detail the network architectures used throughout our study. 

\subsection{Backbone Architecture}
Transfer learning through pre-training is only possible when the models for all task types share the same backbone architecture. To have meaningful results, we need to choose a backbone which works well across all task types we explore. Therefore
we choose the recent high-resolution backbone HRNetV2~\cite{wang20pami} (illustrated in~\autoref{fig:network_backbone_app}).
It has two main advantages.
First, HRNetV2 was shown to outperform ResNet~\cite{he16cvpr}, ResNeXt~\cite{xie16cvpr}, Wide ResNet~\cite{zagoruyko16bmvc}, and stacked hourglass networks~\cite{newell16eccv} on three of the tasks we explore: semantic segmentation, keypoint detection, and object detection.
Second, its design allows to use relative shallow task type specific heads compared to the number of parameters in the backbone. In ~\mainpaperref{tab:num_parameter_per_task} we provide an overview of the number of trainable parameters. The backbone consists of 69M parameters, making up 87\%-99.99\% of all parameters depending on the task type and the number of classes.

\begin{figure}[b]
    \centering
    \includegraphics[width=\columnwidth]{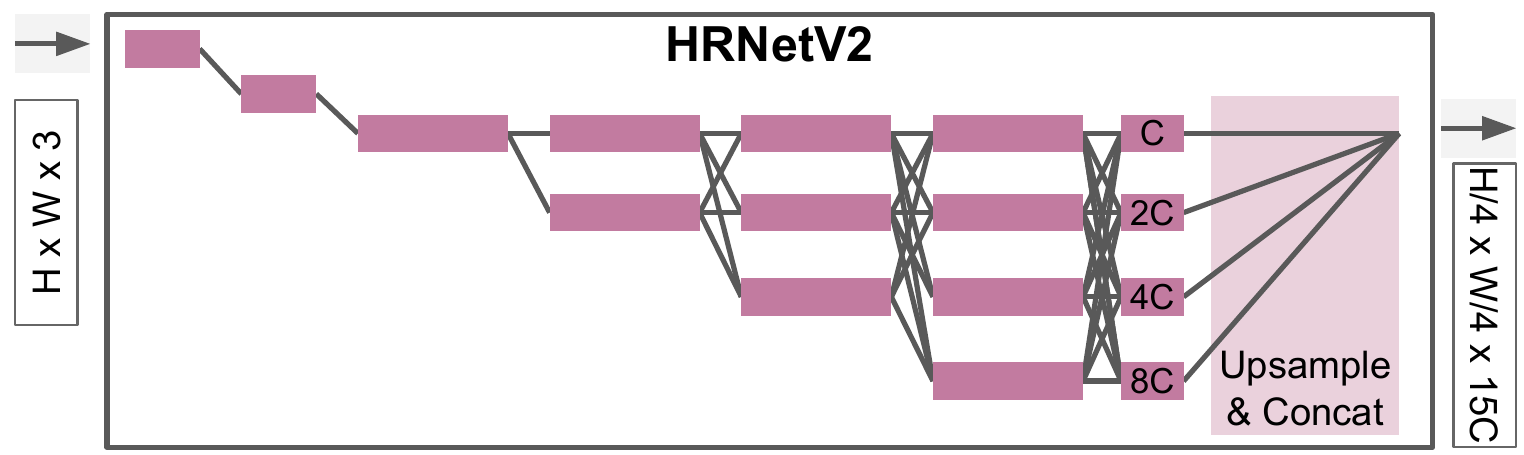}
    \caption{Schematic illustration of the HRNet backbone. Each horizontal layer depicts a resolution and each block represents a series of residual units. The blocks are fully connected using all resolutions with either downscaling or upsampling. The final representation is obtained by concatenating all upsampled layers to $H/4 \times W/4$.}
    \label{fig:network_backbone_app}
\end{figure}

\para{Architecture.}
The HRNetV2 backbone extends the ResNet architecture to preserve high-resolution spatial features.
A regular ResNet consists of blocks organised in four `stages', each reducing the image resolution by factor 2. HRNetV2 follows this design, but after each stage it \emph{also} keeps a parallel high-resolution branch (\autoref{fig:network_backbone_app}).
All branches from one stage are fed into the next, so this next stage incorporates information from the representations at different resolutions.
The backbone produces 4 feature maps, each with different resolution and number of channels. These are combined into a single feature map by upscaling and concatenating them along the channel dimension. This results in a final output feature map of shape $W/4 \times H/4 \times 15C$, where $W$ and $H$ are the original image width and height, and $C=48$ is the number of channels in the highest resolution output layer. Hence the final feature map has $15 \times 48 = 720$ channels.

\para{Training.}
\begin{figure}
    \centering
    \begin{subfigure}[b]{\columnwidth}
        \includegraphics[width=\columnwidth]{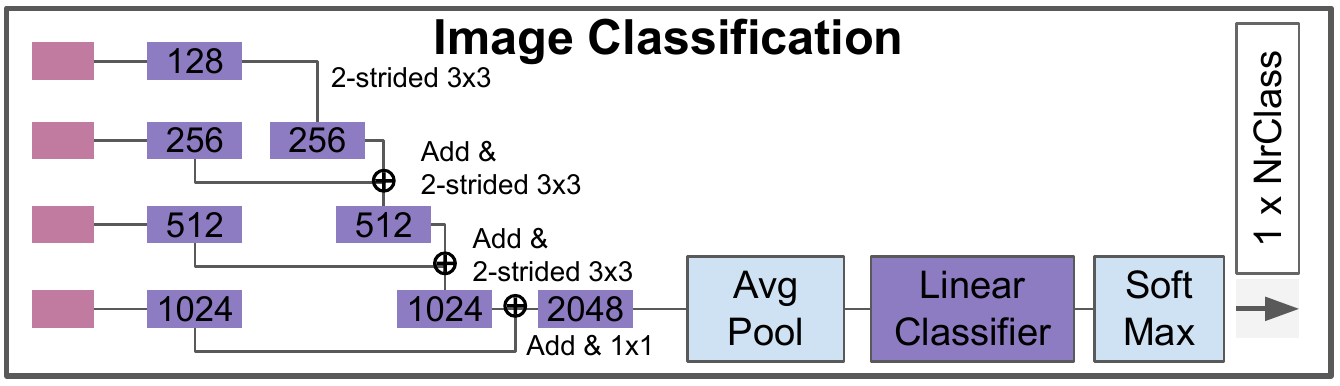}
        \caption{}
        \label{fig:network_classification_head_app}
    \end{subfigure}
    
    \begin{subtable}[b]{\columnwidth}
        \centering
        \begin{tabular}{lc}\toprule
            Description                 & Top-1 Acc\\\midrule
            ResNet-50\cite{wang20pami}  & 76.9      \\
            HRNetV2-W44\cite{wang20pami}& 77.0      \\\midrule
            \textbf{ours} HRNetV2-W48   & 79.5      \\\bottomrule
        \end{tabular}
        \caption{}
        \label{tab:results_classification_head_app}
    \end{subtable}
    \caption{Illustration of the classification head architecture (\ref{fig:network_classification_head_app}) and table with top-1 accuracies for different networks and backbones on the ILSVRC'12 validation set (\ref{tab:results_classification_head_app}). 
    }
    \label{fig:classification_head_app}
\end{figure}

In order to validate our backbone architecture we use image classification on ILSVRC'12.
We extend the backbone with the classification head proposed in~\cite{wang20pami} (\autoref{fig:network_classification_head_app}).
This head increases the number of channels, while reducing the resolution to $H/32 \times W/32$, and then uses an average pooling layer with a linear classifier.
The network is trained for 200 epochs, with SGD using momentum and a large batch-size of 4096, learning-rate decay after 60, 120, and 180 epochs.

\para{Evaluation.}
Performance is evaluated using Top-1 accuracy on the ILSVRC'12 validation set.

\para{Results.}
In \autoref{tab:results_classification_head_app} we show the results of our network compared to the ResNet-50 and HRNetV2-W44 architectures from~\cite{wang20pami}. 
Our model performs competitively, reaching $79.5\%$ Top-1 accuracy, validating our choice of architecture.

\subsection{Semantic Segmentation}
\label{sec:appendix_single_task_semseg}
\begin{figure}
    \centering
    \begin{subfigure}[b]{\columnwidth}
        \centering
        \includegraphics[width=.8\columnwidth]{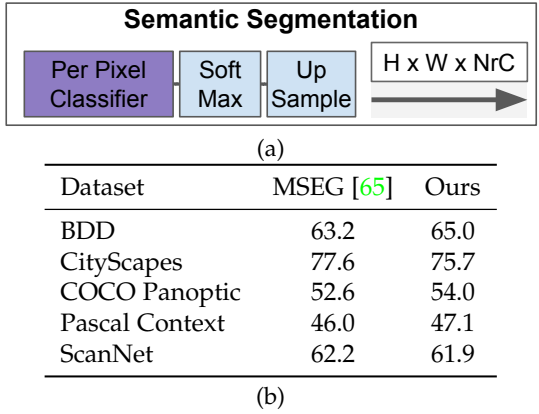}
        \caption{}
        \label{fig:network_semseg_head_app}
    \end{subfigure}
    
    \begin{subtable}[b]{\columnwidth}
        \centering
        \begin{tabular}{lcc}\toprule
            Dataset         & MSEG \cite{lambert20cvpr}  & Ours\\\midrule
            BDD             & 63.2          & 65.0\\
            CityScapes      & 77.6          & 75.7\\
            COCO Panoptic   & 52.6          & 54.0\\
            Pascal Context  & 46.0          & 47.1\\
            ScanNet         & 62.2          & 61.9\\\bottomrule
        \end{tabular}
        \caption{}
        \label{tab:results_semseg_head_app}
    \end{subtable}
    \caption{%
    Semantic segmentation model and performance. 
    \ref{fig:network_semseg_head_app} shows an illustration of the head architecture, where purple blocks indicate trainable parameters (in total $(720 + 1) * NrC$) and blue blocks have no trainable parameters.
    \ref{tab:results_semseg_head_app} compares our performance against that of the MSEG setup~\cite{lambert20cvpr} on several datasets of the MSEG collection. 
    Our setup is validated by the fact we perform on par with~\cite{lambert20cvpr}.
    }
    \label{fig:task_semseg_app}
\end{figure}

\para{Architecture.}
For semantic segmentation, we adopt the network head proposed in~\cite{wang20pami} (\autoref{fig:network_semseg_head_app}).
It consists of three stages:
(1) a 1x1 convolution changing the dimensionality of the backbone output to the number of classes $K$. This implements a linear classifier at each pixel;
(2) a softmax non-linearity;
(3) a bi-linear up sampling layer to produce the final predicted segmentation map $H \times W \times K$.

\para{Training.}
The network is trained using the cross-entropy loss at each pixel, ignoring the background class:
\begin{align}
    L_{\textrm{log}} &= \frac{1}{N} \sum_{k=1}^{N} \delta(y_k \ne \textrm{bg}) \log p(y_k | x_k),
\end{align}
where the $\delta$ denotes the Dirac-delta function to return 1 iff the pixel depicts a not background class, and $p(y_k | x_k)$ is computed using the softmax function.

Starting from ILSVRC'12 pre-trained weights, we train on the source dataset using SGD with momentum. We use multiple Google Cloud TPU-v3-8 accelerators using synchronized batch norm and a batch size of 32.
For almost all sources datasets we use a stepwise learning rate decay, and optimize per source the starting learning rate, the number of steps after which the learning rate is lowered (by a fixed factor of 10), and the total number of training steps. 
However, we found performance to be unstable for the small SUIM and KITTI datasets. 
For these we found that switching to a ``poly'' learning rate policy $(\mathrm{base\_lr} \times (1 - \frac{\mathrm{curr\_step}}{\mathrm{max\_steps}})^p$, $p=0.9)$ stabilizes training, as also done in~\cite{chen18pami,lambert20cvpr}.

\para{Evaluation.}
Performance is evaluated by the Intersection-over-Union (IoU), averaged over classes~\cite{everingham15ijcv}.
During evaluation we process complete images at a single resolution only. 
We scale each image to match the resolution used during training.

\para{Results.}
To validate our training setup, we compare the performance of our networks on a subset of the MSEG dataset collection~\cite{lambert20cvpr}. 
While these datasets are also part of our collection, for this experiment we use the annotation provided in MSEG, which has fewer classes than in our transfer setting setup, \cf \mainpaperref{sec:transfer_experiments}.
As \autoref{tab:results_semseg_head_app} shows, our models perform on par with those of~\cite{lambert20cvpr}, which also use a HRNetV2 backbone and linear semantic segmentation layer, but have different normalisation and data augmentation steps and a different implementation.

\subsection{Object Detection model architecture details}\label{sec:appendix_networks_od}
\begin{figure}
    \centering
    \begin{subfigure}[b]{\columnwidth}
        \centering
        \includegraphics[width=.9\columnwidth]{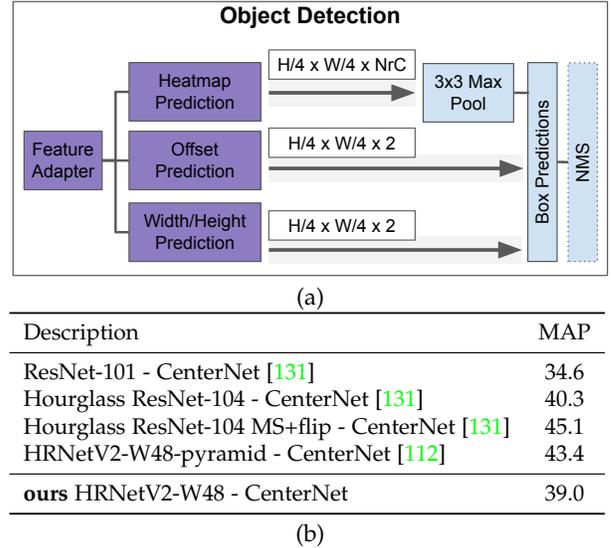}
        \caption{}
        \label{fig:network_detection_head_app}
    \end{subfigure}
    ~
    \begin{subtable}[b]{\columnwidth}
        \centering
        \resizebox{.9\columnwidth}{!}{%
        \begin{tabular}{lc}\toprule
            Description                                         & MAP\\\midrule 
            ResNet-101 - CenterNet~\cite{zhou19arxiv}           & 34.6 \\
            Hourglass ResNet-104 - CenterNet~\cite{zhou19arxiv} & 40.3 \\
            Hourglass ResNet-104 MS+flip - CenterNet~\cite{zhou19arxiv} & 45.1 \\
            HRNetV2-W48-pyramid - CenterNet~\cite{wang20pami}       & 43.4\\\midrule
            \textbf{ours} HRNetV2-W48 - CenterNet               & 39.0\\\bottomrule
        \end{tabular}
        }
        \caption{}
        \label{tab:results_detection_head_app}
    \end{subtable}
    \caption{%
    Object detection model and performance. 
    \ref{fig:network_detection_head_app} illustrates our CenterNet-based~\cite{zhou19arxiv} object detection head.
    \ref{tab:results_detection_head_app} compares the performance to recent works. Our model performs close to the CenterNet Hourglass ResNet-14~\cite{zhou19arxiv} under the same conditions: no data augmentation during evaluation (neither Multi-Scale (MS) nor horizontal flipping (flip)). This demonstrates that we base our transfer learning experiments on a strong modern object detection framework.
    }
    \label{fig:task_detection_app}
\end{figure}

\para{Architecture.}
We follow the CenterNet\footnote{With CenterNet we denote a family of object detection models which predicts boxes via their center points with fully convolutational architectures~\cite{duan19iccv,zhou19arxiv}. The paper of~\cite{duan19iccv} is called CenterNet while the GitHub of~\cite{zhou19arxiv} is also called CenterNet \url{https://github.com/xingyizhou/CenterNet}. We mostly follow the approach of~\cite{zhou19arxiv}.}
approach of Zhou et al.~\cite{zhou19arxiv} (\autoref{fig:network_detection_head_app}).
For each pixel in the feature map and for each class, a binary classifier evaluates whether this pixel is the center point of an object bounding box.
Additionally, at these points we predict a translation correcting offset (in x- and y- coordinates) and the size of the bounding box (width and height). Using center points to predict bounding boxes results in a simpler model than a two-stage architecture like Faster-RCNN~\cite{ren15nips}, while being about as accurate~\cite{duan19iccv,zhou19arxiv}.

Following~\cite{zhou19arxiv, wang20pami} we add a feature adapter (\emph{neck}) between the backbone and the head. 
This neck consists of 1x1 convolution followed by a 3x3 convolution, batch-normalization~\cite{ioffe15icml} and ReLu. 
The detection CenterNet head outputs three prediction maps, each is the result of a 3x3 convolution, ReLu, and another 3x3 convolution.
The three pixel-wise prediction maps are:
\begin{itemize}
    \item [M1] a $K$-channel heatmap $\hat{Y}_{xyc}$, one per class. Each pixel is assigned the likelihood to be the center on an object of a certain class computed using Gaussian kernel.
    \item [M2] a two-channel offset map $\hat{O}_{p}$ representing x- and y-offsets of the point to the real center of the bounding box $p$; and
    \item [M3] a two-channel width/height map $\hat{S}_{p}$, which predicts the width and height of an object centered at the pixel $p$.
\end{itemize}
Following~\cite{zhou19arxiv} the final box predictions are obtained by combining the top $T=100$ highest scoring pixels from the class-specific heatmaps, with their respective x/y offsets to define the box-centers. The box dimensions are extracted from the width/height prediction map.
Then, optionally, non-maximum suppression (NMS) is applied.

\para{Training.}

During training, three losses are computed, one for each head~\cite{zhou19arxiv}:
\begin{itemize}
    \item [L1] a logistic regression focal loss~\cite{law18eccv,lin17iccv} for the class heatmap, acting on all pixels:
    \begin{align}
        L_{\textrm{cls}} &= -\frac{1}{N} \sum_{xyc} \begin{cases} (1 - \hat{Y}_{xyc})^{\alpha}\log(\hat{Y}_{xyc}),&\quad Y_{xyc} = 1, \\
        (1 - Y_{xyc})^{\beta}(\hat{Y}_{xyc})^{\alpha}\log(1 - \hat{Y}_{xyc}),&\quad Y_{xyc} \ne 1 \end{cases},\label{eq:focal_loss_od_app}
    \end{align}
    Here $Y_{xyc}$ is a sum of Gaussians around from all ground-truth boxes object centers, projected to low-resolution feature map. For $\alpha = 2$ and $\beta = 4$ we use default parameters provided in~\cite{zhou19arxiv}
    \item [L2] a L1-loss for offset prediction, acting only on $N$ pixels that are close to center points of objects in the ground-truth (those where $Y_{xyc}=1$); here $o_k$ are the ground-truth offsets:
    \begin{align}
        L_{\textrm{off}} &= \frac{1}{N} \sum_{k=1}^N |\hat{O}_{p_k} - o_k |
        \label{eq:l1_loss_offsets_od_app}
    \end{align}
    \item [L3] a L1-loss for width and height prediction, again acting only on $N$ pixels close to ground-truth centers; $s_k$ are ground-truth box sizes:
    \begin{align}
        L_{\textrm{size}} &= \frac{1}{N} \sum_{k=1}^N | \hat{S}_{p_k} - s_k |
        \label{eq:l1_loss_width_height}
    \end{align}
\end{itemize}
The final training objective is $L = L_{\textrm{cls}} + \lambda_{\textrm{off}}L_{\textrm{off}} + \lambda_{\textrm{size}}L_{\textrm{size}}$.
We train using the same procedure as for semantic segmentation, but using a batch size of $128$ and the Adam optimizer~\cite{kingma15iclr} until convergence.

\para{Evaluation.}
We use the COCO definition of mean Average Precision (mAP)~\cite{lin14eccv}. They evaluate mAP at various IoU thresholds and average them. For COCO we do not use Non Maximum Suppression, since this did not improve results on this dataset as also observed by~\cite{zhou19arxiv} (on other datasets NMS was useful, see details in~\appendixsecref{sec:appendix_results}). 

\para{Results.}
We validate our setup by training on the COCO training set and evaluating performance on its validation set. We compare to results reported in~\cite{zhou19arxiv} and for a HRNetV2-W48 backbone with a feature pyramid using the CenterNet variant~\cite{wang20pami}.

As can be seen in~\autoref{tab:results_detection_head_app}, the best results are obtained by using use data augmentation during evaluation (multi-scale and horizontal flip) or by using a feature pyramid.
Without these enhancements, our implementation performs close to the Hourglass ResNet-104 results of~\cite{zhou19arxiv} (39.0 mAP and 40.3 mAP respectively).
Hence we conclude that we can base our transfer learning experiments on a strong, modern object detection framework.

\subsection{Keypoint Detection}
\begin{figure}
    \centering
    \begin{subfigure}[b]{\columnwidth}
        \centering
        \includegraphics[width=.9\columnwidth]{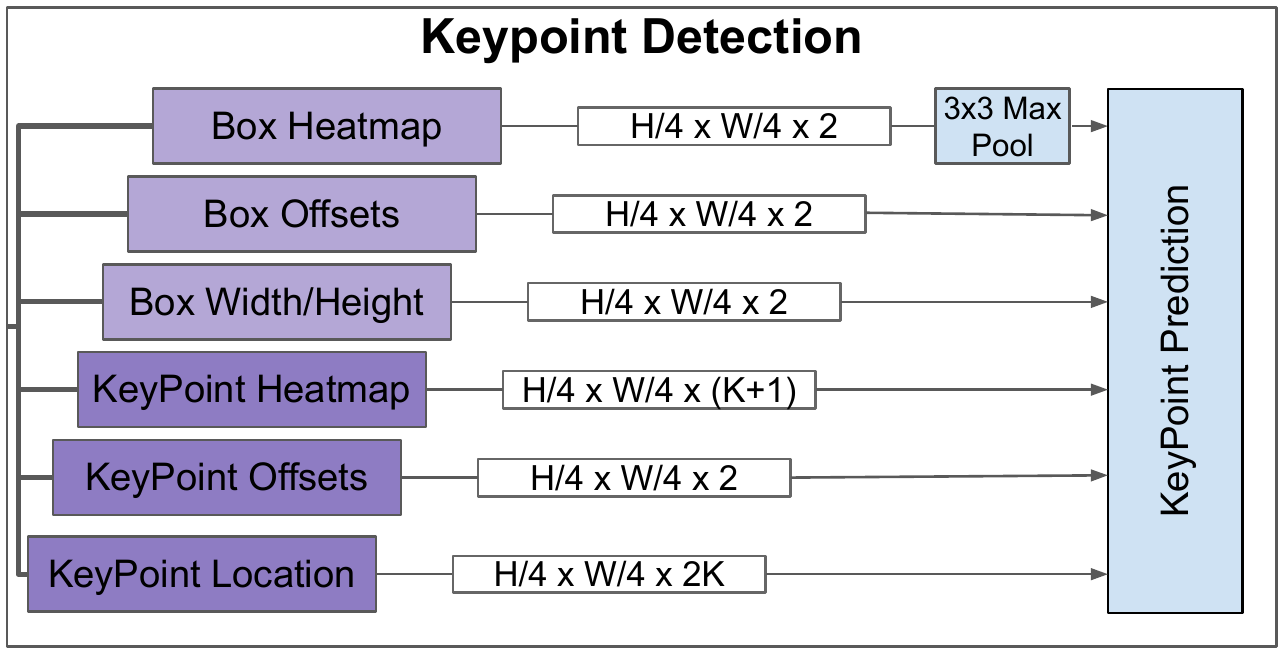}
        \caption{}
        \label{fig:network_keypoints_head_app}
    \end{subfigure}
    
    \begin{subtable}[b]{\columnwidth}
        \centering
        \resizebox{.9\columnwidth}{!}{%
        \begin{tabular}{lc}\toprule
            Description                                             & AP50\\\midrule
            Hourglass-104 - CenterNet - single stage~\cite{zhou19arxiv} & 84.2\\ %
            HRNetV1-W44 - two stage cascade~\cite{wang20pami}& 90.8\\\midrule
            \textbf{ours} HRNetV2-W48 - CenterNet - single stage    & 81.1\\\bottomrule
        \end{tabular}
        }
        \caption{}
        \label{tab:results_keypoints_head_app}
    \end{subtable}\vspace{-2mm}
    \caption{
        Keypoint Detection model and performance.
        \ref{fig:network_keypoints_head_app} illustrates our model head architecture.
        \ref{tab:results_keypoints_head_app} provides a performance comparison.
        The two stage cascade model first detects persons, crops out their image pixels, and uses a separate network to predict keypoints. We follow the simpler 
        single stage approach used in \eg~\cite{zhou19arxiv}.
        }
    \label{fig:results_keypoints_app}
\end{figure}

\para{Architecture.}
We again follow the CenterNet approach of~\cite{zhou19arxiv} (\autoref{fig:network_keypoints_head_app}). The keypoint head architecture resembles the object detection architecture, yet with some differences.

We create 6 pixelwise prediction heads, each composed of a 3x3 convolution, ReLu, and another 3x3 convolution.
We do not use a neck, but directly feed the feature maps produced by the backbone into these heads.
The maps output by the first three heads are analogous to the object detection ones, but for a single class only, to detect the box around the person (or dog depending on the dataset).
The remaining three maps are:
\begin{enumerate}
    \item[M4] a $K$-channel keypoint heatmap, where each channel is the output of a binary classifier predicting one of the $K$ keypoints;
    \item[M5] a keypoint offset map: a two-channel x/y offset map for a translation correction of the keypoint;
    \item[M6] a keypoint allocation map: a $2 \times K$-channel map marking the displacement from the center of an object at this pixel to each of the $K$ keypoints belonging to that object.
\end{enumerate}

We obtain keypoints for a image by processing these maps: 
First, a bounding box around the person/dog is obtained by combining the box centers (M1), with the offsets (M2) and the box dimensions (M3).
Then, M4 and M5 are combined to create a set of candidate keypoint locations with accompanying confidence scores, within the person/dog bounding box. 
Finally, M6 is used to associate each candidate keypoint to the person/dog object it belongs to.

\para{Training.}
For M1-M3 we use the same losses L1-L3 as for object detection. For M4-M6 we add the following~\cite{zhou19arxiv}:
\begin{enumerate}
   \item [L4] a focal loss is used for multi-class joint classification, analogous to object detection loss L1 as described by \autoref{eq:focal_loss_od_app};
   \item [L5] a L1-loss is used for offset prediction, analogous to object detection loss L2 as described by~\autoref{eq:l1_loss_offsets_od_app};
   \item [L6] a L1-loss is used to directly regress the initial keypoint locations w.r.t. the object centers predicted by M1+M2.
\end{enumerate}   

We largely follow the training setup for object detection, using a batch size of 128 and the Adam optimizer~\cite{kingma15iclr}. We set the starting learning rate and number of steps per dataset.

\para{Evaluation.}
We report the mean Average Precision at 0.5 Object Keypoint Similarity (OKS), as defined by the COCO challenge~\cite{coco-challenge}. 

\para{Results.}
To validate our setup we train for keypoint prediction on the COCO dataset. In~\autoref{tab:results_keypoints_head_app} we report the performance of our network architecture and compare to~\cite{zhou19arxiv}. They use as backbone a stacked hourglass~\cite{newell16eccv} consisting of two consecutive hourglass models, which they call Hourglass-104. Their model yields 84.2 AP50.\footnote{
Result obtained via correspondence with the authors: The public available model in the zoo obtaining 64.0 AP50 on \texttt{coco-test}, obtains 84.2 mAP on \texttt{coco-val}, when evaluated without flipping,  see:\newline\url{https://github.com/xingyizhou/CenterNet}
} %
For completeness, we include the results of a more complex two stage cascade keypoint detection approach~\cite{wang20pami}, which obtains very good performance.
In this paper we use the HRNetV2 backbone with the simpler single-stage CenterNet heads.
Our model yields 81.1 mAP, which is near~\cite{zhou19arxiv} and thus strong enough for our transfer learning exploration.

\subsection{Depth Estimation}
\begin{figure}
    \centering
    \begin{subfigure}[b]{\columnwidth}
        \centering
        \includegraphics[width=.8\columnwidth]{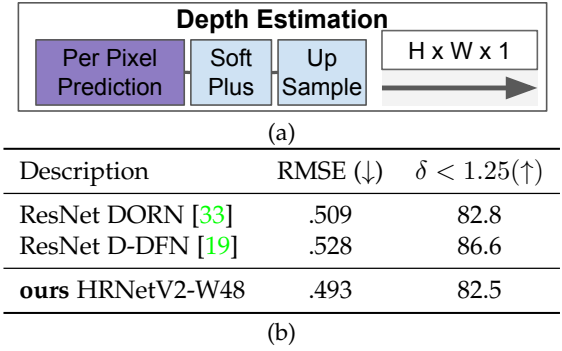}
        \caption{}
        \label{fig:network_depth_head_app}
    \end{subfigure}
    
    \begin{subtable}[b]{\columnwidth}
        \centering
        \begin{tabular}{lcc}\toprule
            Description                         & RMSE ($\downarrow$)   & $\delta < 1.25 (\uparrow)$\\\midrule
            ResNet DORN \cite{fu18cvpr}         & .509                  & 82.8                      \\
            ResNet D-DFN\cite{chen193dv}        & .528                  & 86.6                      \\\midrule
            \textbf{ours} HRNetV2-W48           & .493                  & 82.5                      \\\bottomrule
        \end{tabular}
        \caption{}
        \label{tab:results_depth_head_app}
    \end{subtable}\vspace{-2mm}
    \caption{%
    Monocular Depth Estimation.
    \ref{fig:network_depth_head_app} illustrates the task head architecture.
    \ref{tab:results_depth_head_app} shows a comparison of performance on NYUDepth- V2~\cite{silberman12eccv} dataset using a RMSE and $\delta < 1.25$ accuracy as performance measures.
    While our architecture features a light-weighted head, it still outperforms depth-specific architectures in terms of RMSE and is on par with $\delta < 1.25$.
    }
    \label{fig:task_depth_app}
\end{figure}

\para{Architecture.}
For monocular depth estimation, we modify the architecture for semantic segmentation. We propose to add a single regression layer on top of the output of the backbone, followed by a soft-plus layer and a bi-linear upsampling layer to match the original resolution (\autoref{fig:network_depth_head_app}). 
The softplus activation converts the logit value to depth: $\log(\exp(x) + 1)$ and is also used in~\cite{gordon19iccv, zhu19iccv}.
This is a differentiable clipping function which ensures that depth predictions are positive.

\para{Training.}
To train the network, we combine two losses following~\cite{chen193dv,li17iccv}:
\begin{itemize}
    \item a L1-loss between the predicted depth map and the ground truth map at each pixel;
    \begin{equation}
        L_{\textrm{L1}} = \frac{1}{N} \sum_{k=1}^N |y_k - \hat{y}_k|
    \end{equation}
    \item a smoothness loss which encourages differences in neighbouring depth pixels to be the same in the predicted depth map and the ground-truth. More precisely, we take the derivative in x-direction for both the predicted and ground-truth maps, and compare them with an L1-loss. We repeat this for the y-direction. Resulting in:
    \begin{equation}
        L_{\textrm{smooth}} = \frac{1}{N} \sum_{k=1}^N | (\nabla_x y)_k - (\nabla_x \hat{y})_k| + | (\nabla_y y)_k - (\nabla_y \hat{y})_k|,
    \end{equation}
\end{itemize}
where $y$ denotes the ground-truth depth map, $\hat{y}$ the predicted depth map and $N$ the number of pixels.

For training we use a batch size of 64 and the Adam optimizer~\cite{kingma15iclr}.

\para{Evaluation.}
We validate on the NYUDepthV2~\cite{silberman12eccv} dataset, using the root mean squared error (RMSE) metric and the $\delta < 1.25$ accuracy, where $\delta = \max(\tfrac{\hat{z}}{z},\tfrac{z}{\hat{z}})$ is a measure of relative accuracy defined in~\cite{ladicky14cpr}.

\para{Results.}
We compare our proposed network to two recent ResNet based models: 
\cite{fu18cvpr} which uses depth-specific losses based on ordinal regression; and
\cite{chen193dv} which uses the same losses as we do, but with a depth-specific network architecture on top of the ResNet backbone.
The results are presented in \autoref{tab:results_depth_head_app}. 
We observe that our model is slightly better than both others, when measured using the root mean squared error (RMSE), while slightly behind on the $\delta < 1.25$ measure. 
Hence, we conclude that our light-weight depth prediction head is well suited for monocular depth estimation.

\section{Additional Scenarios}
\label{sec:detailed_generalization_experiments}
\label{sec:appendix_generalization}

\begin{table}[tb]
    \includegraphics[height=75mm]{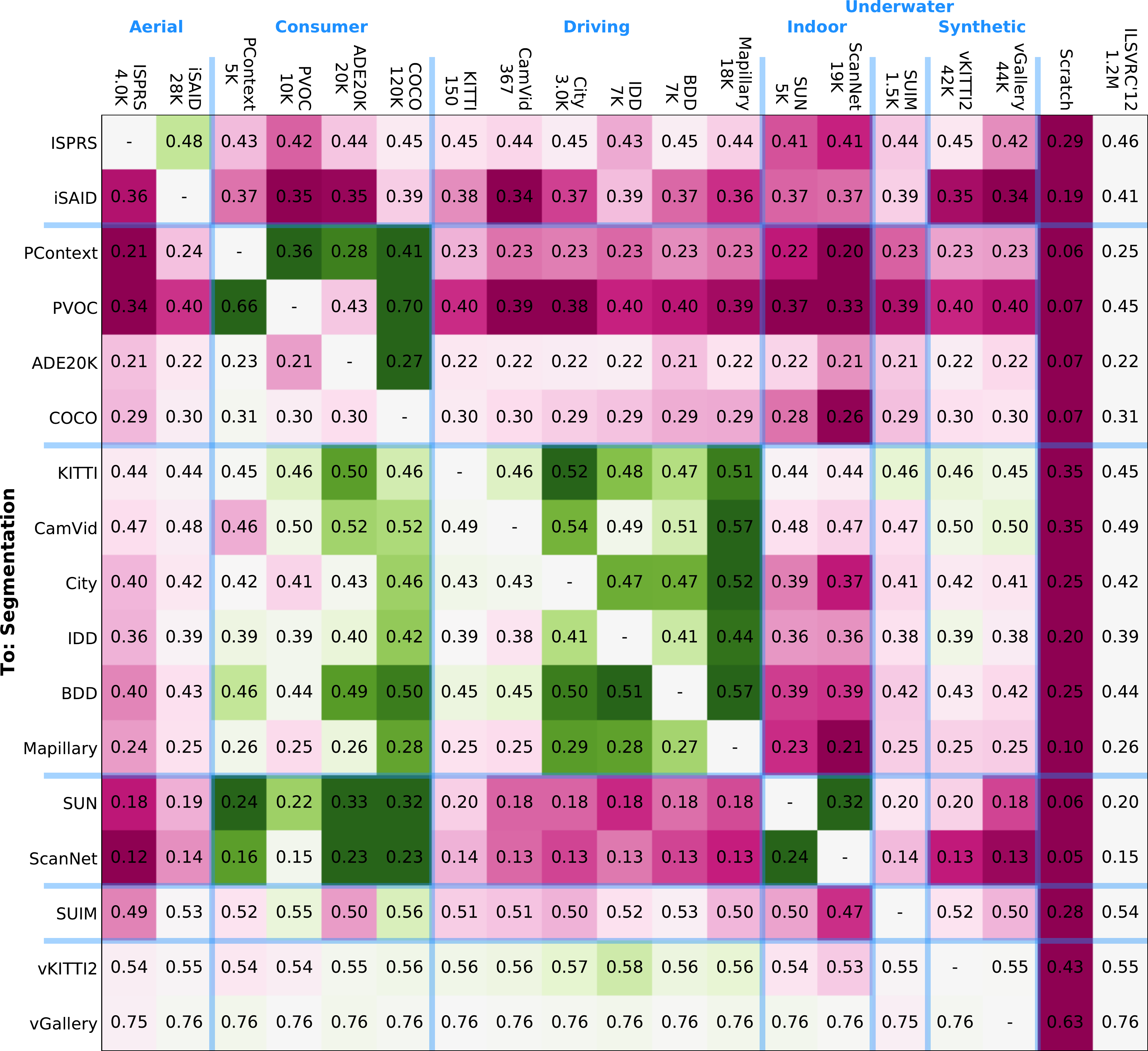}
    \includegraphics[height=75mm]{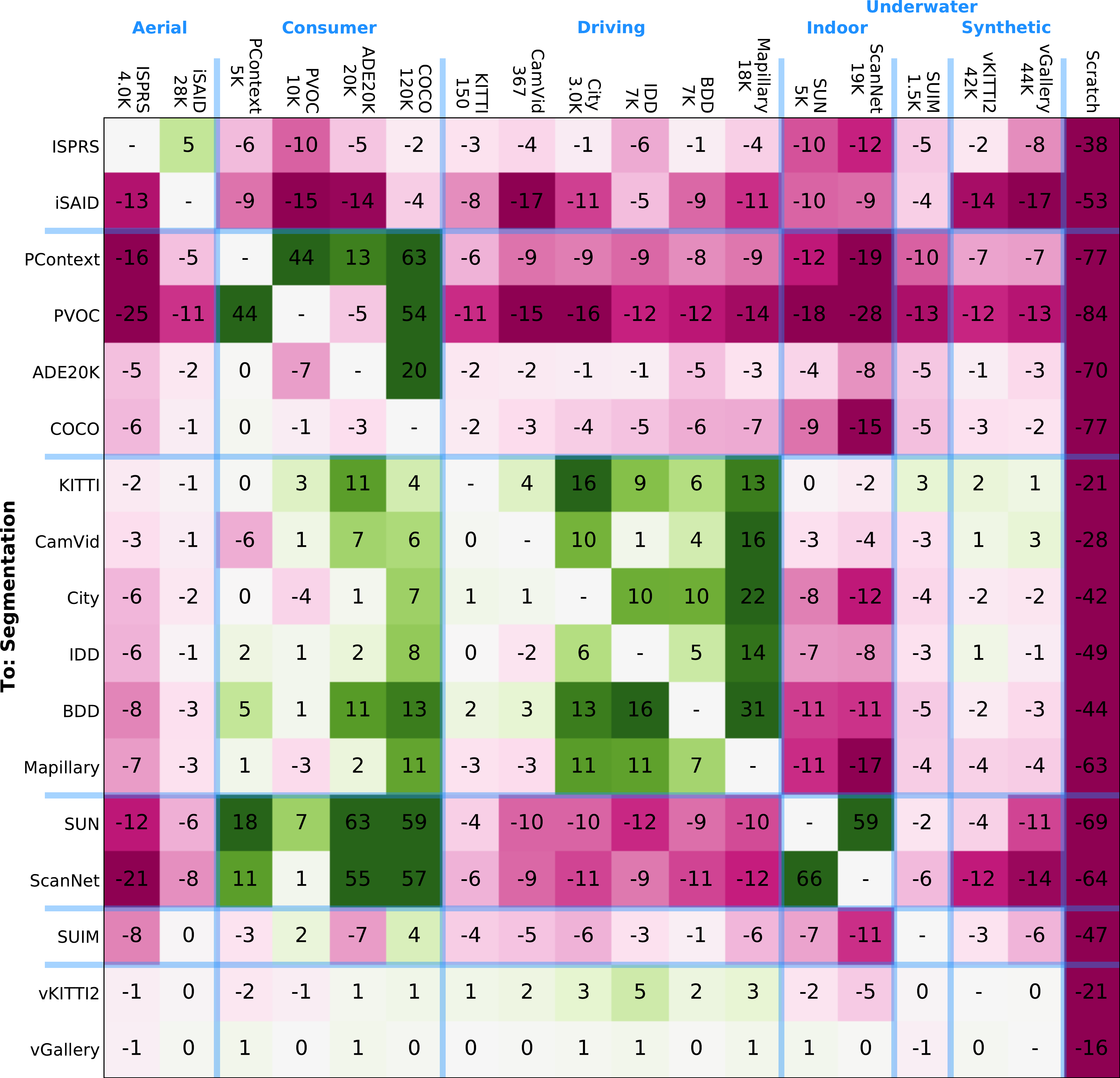}
    \caption{
    HRNetV2 backbone using a fixed crop ($713 \times 713$) instead of dataset specific crops.
    Absolute performance and the corresponding relative transfer gains for semantic segmentation as both a source and target task type. 
    Top table: mean Intersection-over-Union.
    Bottom table: the corresponding relative transfer gains.
    }
    \label{tab:tt_segmentation_raw_within_fixed_crop}
\end{table}

\begin{table*}[p]
    \centering
    \begin{subtable}[t]{\textwidth}
        \centering
        \includegraphics[height=80mm]{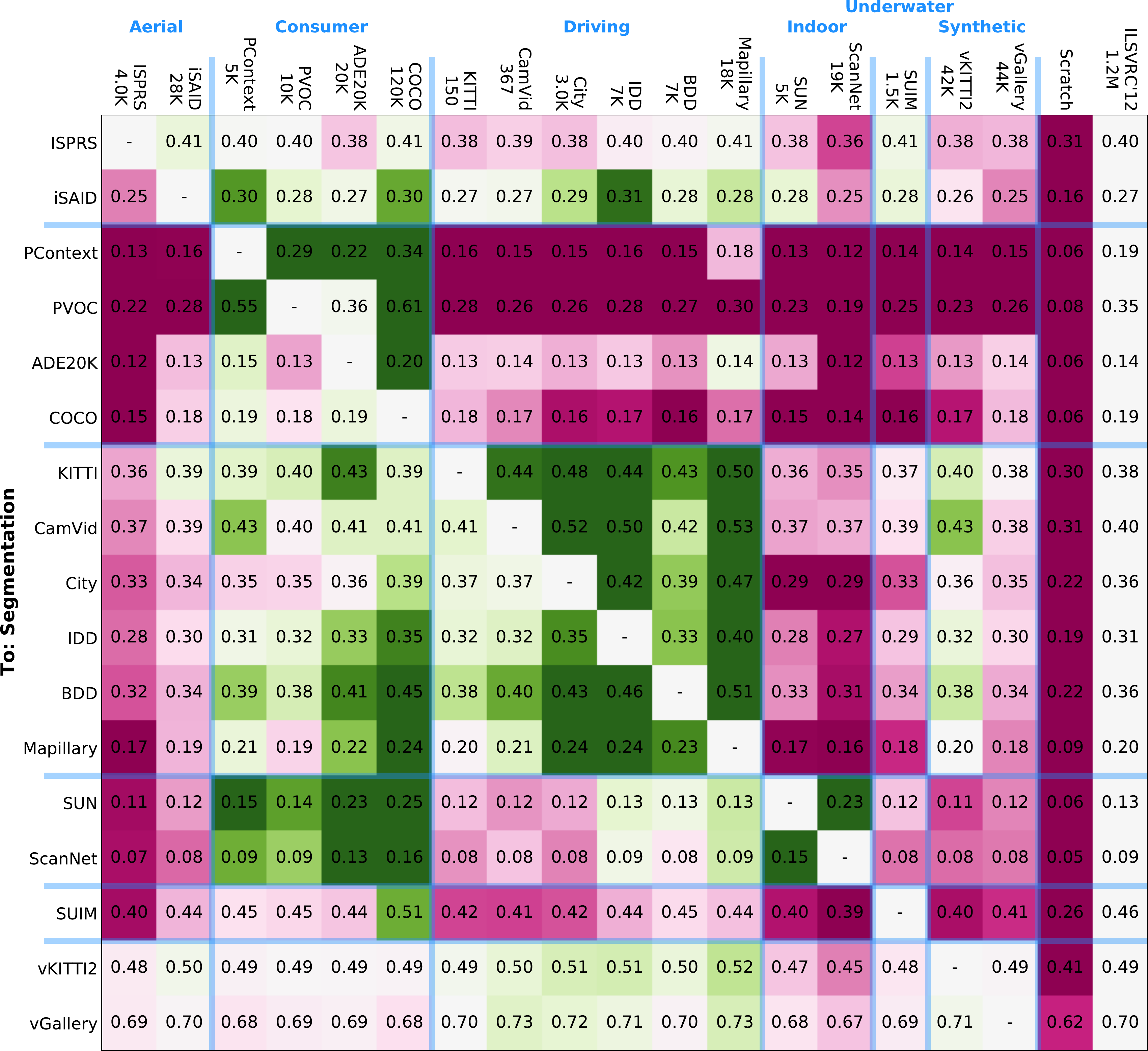}\quad 
        \includegraphics[height=80mm]{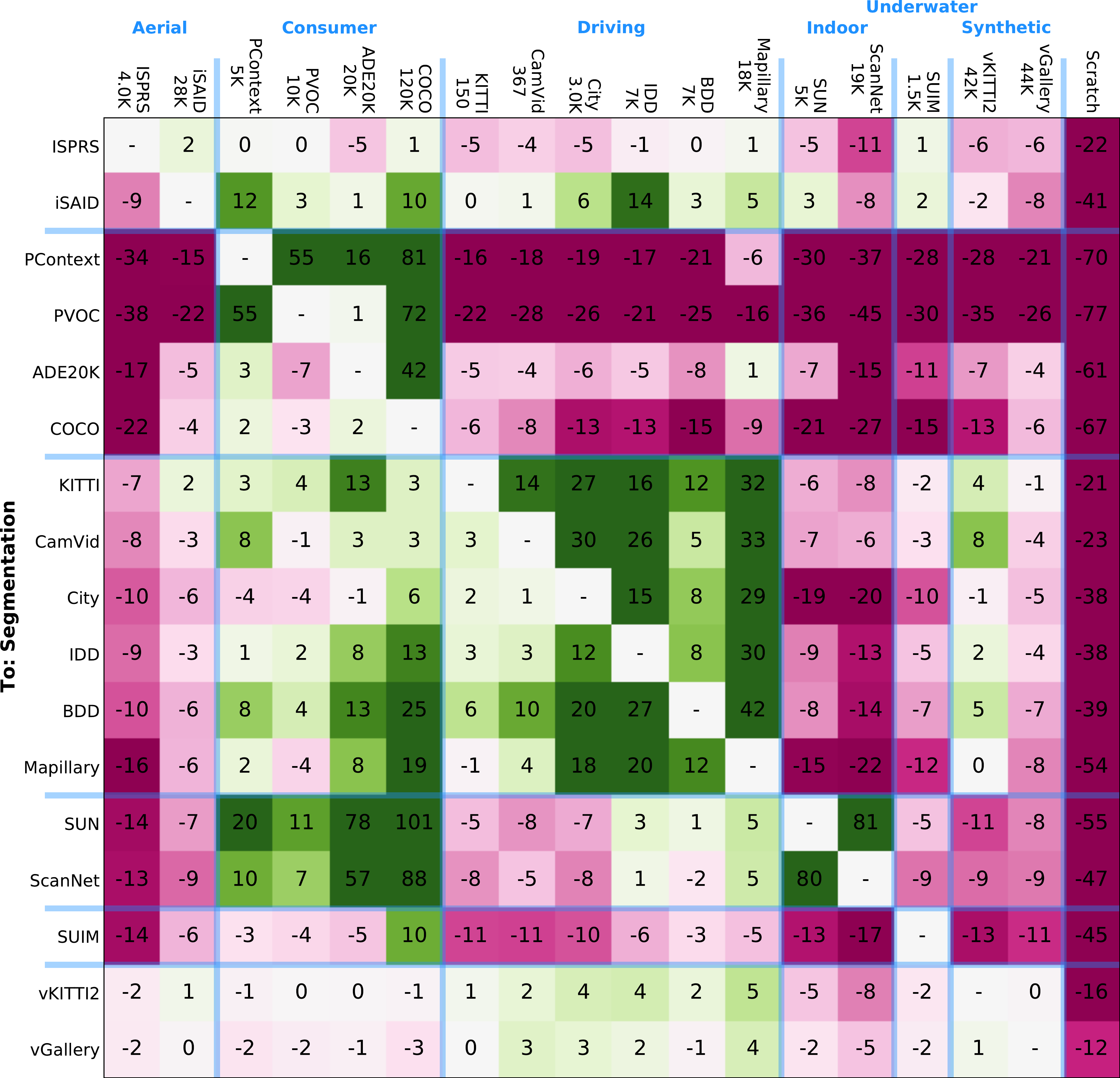}
        \caption{ResNet50 Classification on ILSVRC'12}
        \label{tab:tt_segmentation_raw_within_resnet_full}    
    \end{subtable}
    
    \begin{subtable}[t]{\textwidth}
        \centering
        \includegraphics[height=80mm]{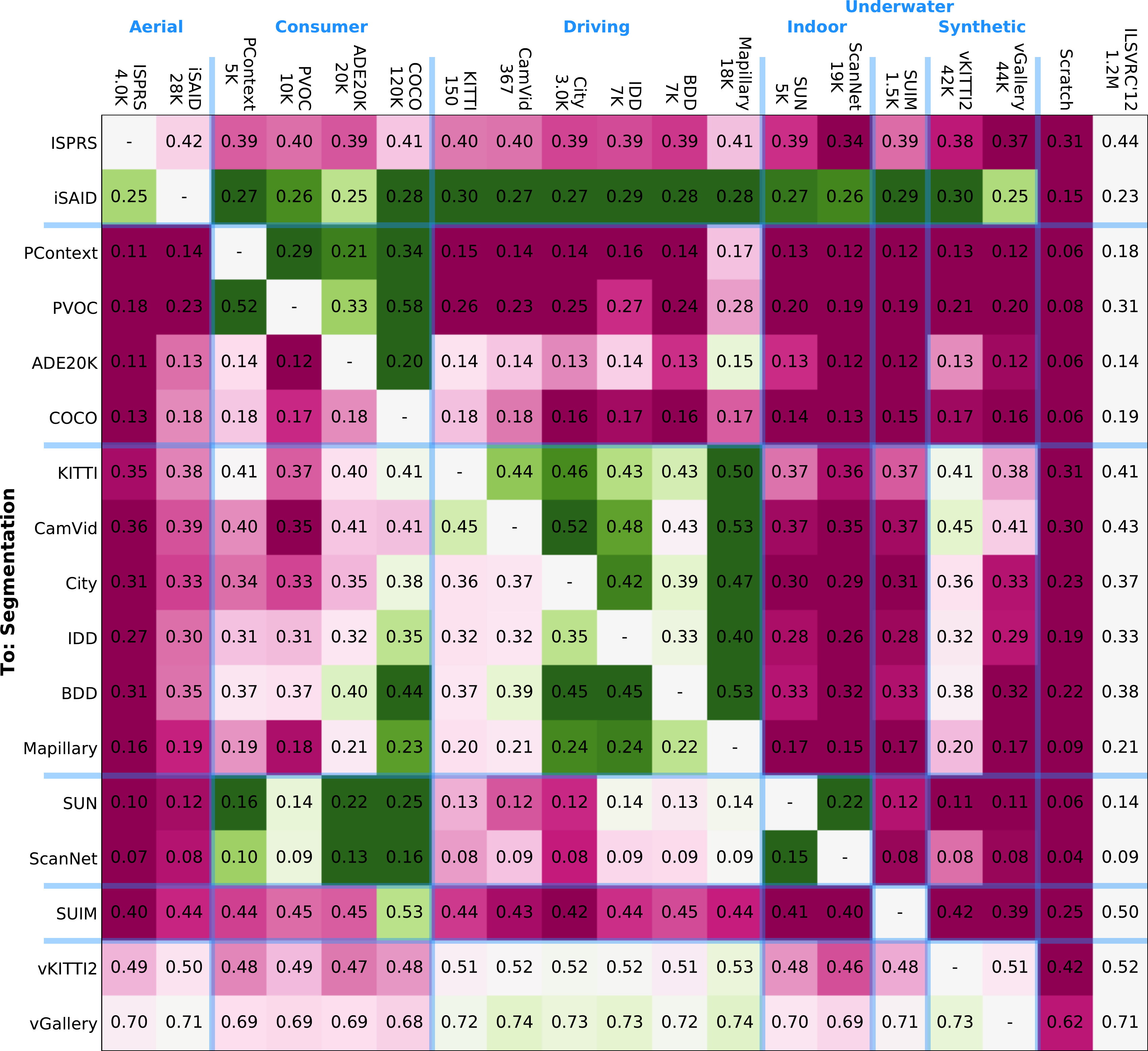}\quad 
        \includegraphics[height=80mm]{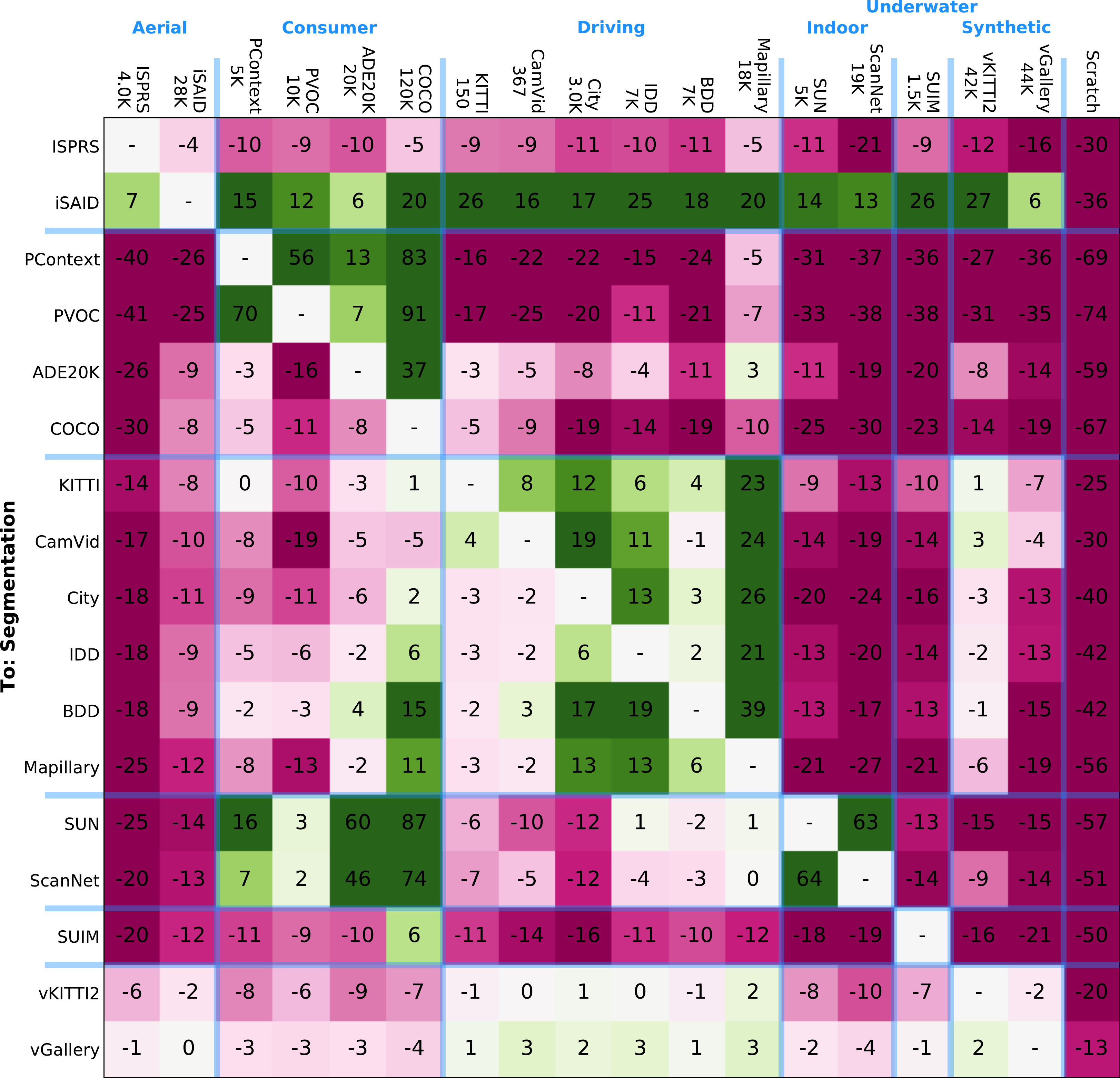}
        \caption{ResNet50 Self Supervised on ILSVRC'12}
        \label{tab:tt_segmentation_raw_within_resnet_ss}    
    \end{subtable}
    \caption{Comparison of using different network architectures and pre-training task: (a) ResNet50 pre-trained on ILSVRC'12 classification; (b) ResNet50 trained self-supervised on ILSVRC'12 training set~\cite{chen20nips}.
    Absolute performance and the corresponding relative transfer gains for semantic segmentation as both a source and target task type. 
    Left tables: mean Intersection-over-Union.
    Right tables: the corresponding relative transfer gains.
    }
    \label{tab:tt_segmentation_raw_within_resnet}    
\end{table*}

\begin{table*}[p]
    \centering
    \begin{subtable}[t]{.5\textwidth}
        \includegraphics[height=32mm]{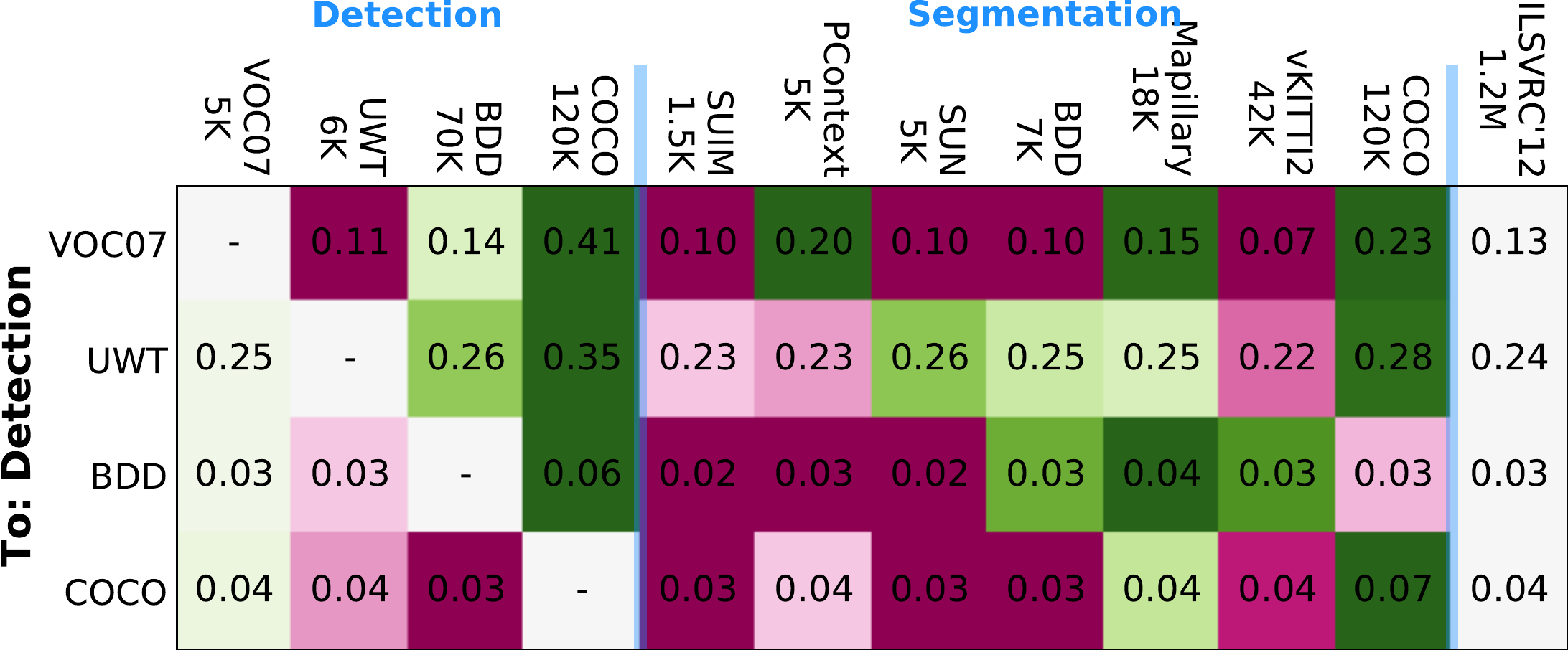}
        \caption{Average Precision}
    \end{subtable}
    ~
    \begin{subtable}[t]{.45\textwidth}
        \includegraphics[height=32mm]{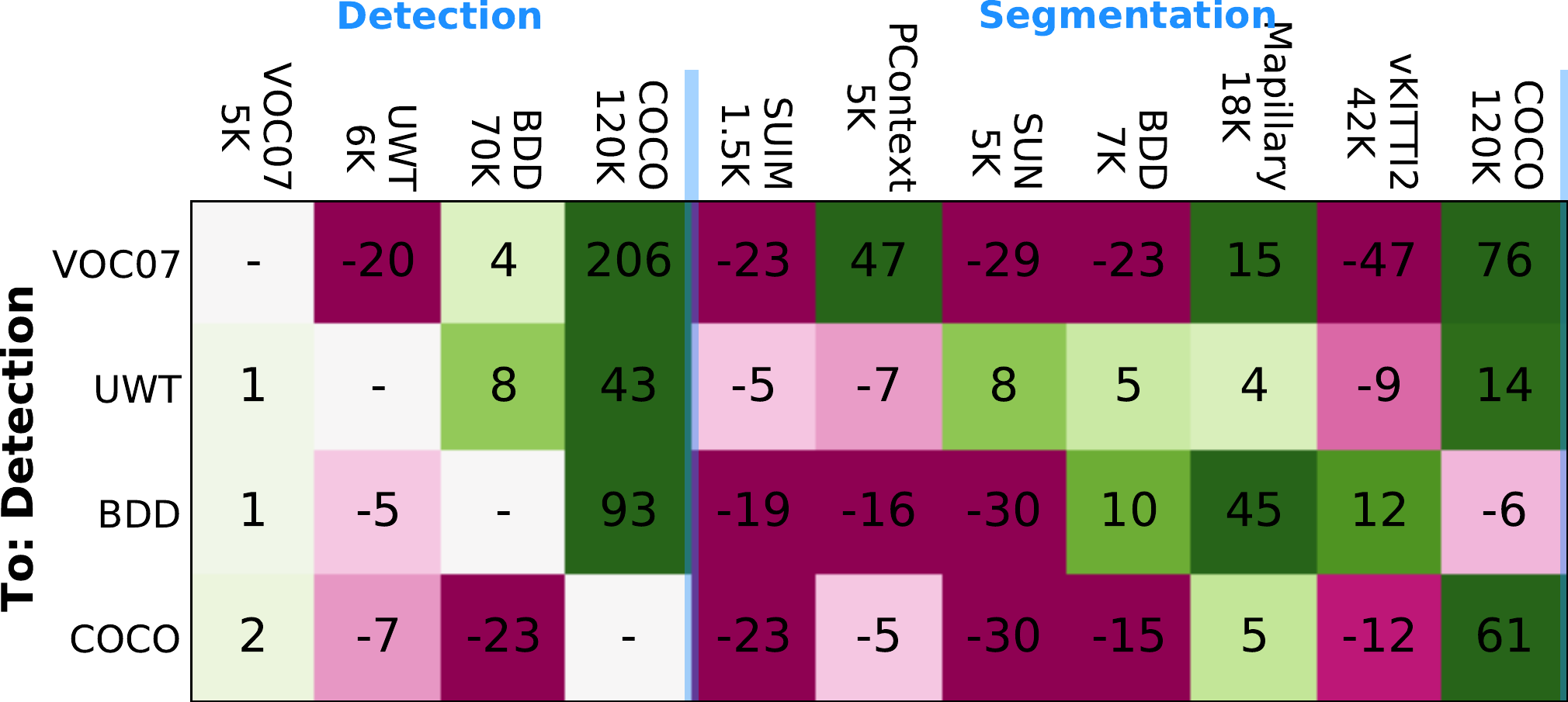}
        \caption{Relative transfer gain}
    \end{subtable}
    \caption{Absolute metrics and the corresponding relative transfer gains for object detection as a target task type \textit{when using ResNet50 as backbone.}
    }
    \label{tab:tt_detection_raw_resnet}
\end{table*}

\begin{table}[t]
    \centering
    \includegraphics[height=120mm]{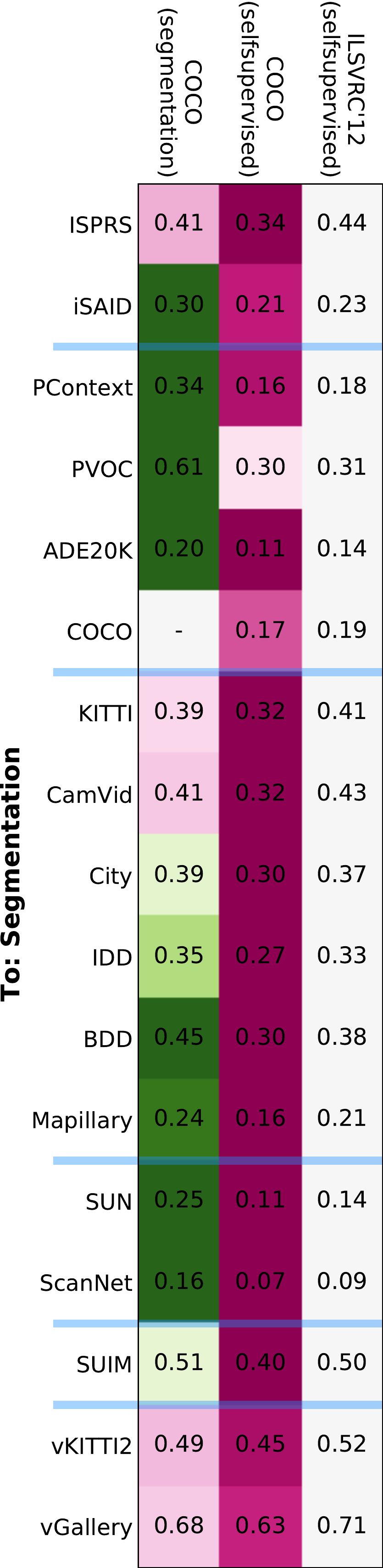}
    \includegraphics[height=120mm]{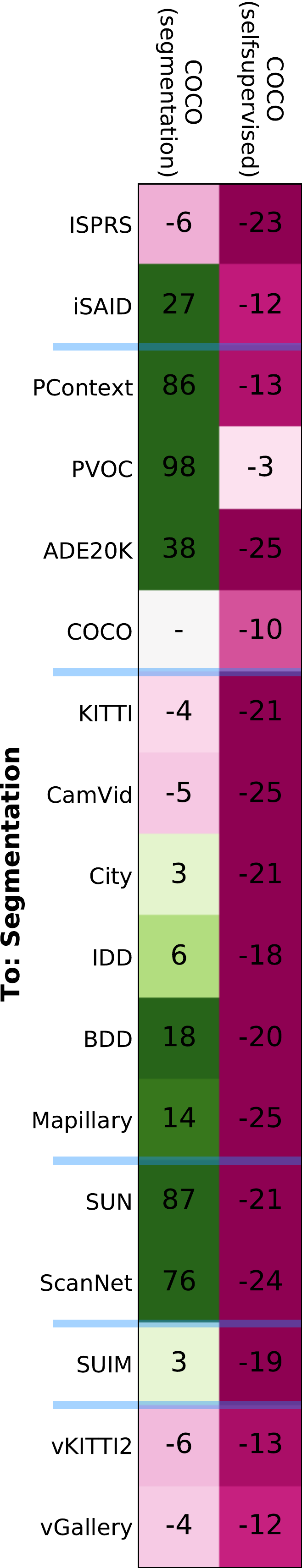}
    \caption{Self-supervised Transfer Chain. Mean IoU (left), Relative (right).}
    \label{tab:tt_segmentation_raw_within_resnet_ss_chain}    
\end{table}
In this section we perform several additional experiments to verify whether our work generalizes to different scenarios.
In particular, while we optimized image resolution per dataset, \cf~\mainpaperref{sec:data_normalization}, we explore what happens when we fix the image resolution across all datasets. In addition, we partially redo experiments using ResNet50~\cite{he16cvpr} as backbone. Finally, we explore self-supervision in our transfer learning chain.

\para{Fixed image resolution.} We redo the segmentation experiments in the small target training setting (i.e. \mainpaperref{tab:few_shot_semseg_within}) but now fixing the image resolution to be $713 \times 713$ pixels across all datasets. Results are shown in~\autoref{tab:tt_segmentation_raw_within_fixed_crop}. The average accuracy across all experiments goes down by -0.01 IoU. However, when comparing the results tables (optimized vs fixed resolution) we qualitatively observe that all transfer patterns remain very similar.

Quantitatively, for 88\% of the experiments with significant positive or negative gains in ~\mainpaperref{tab:few_shot_semseg_within}, similar significant gains are observed at a fixed image resolution in~\autoref{tab:tt_segmentation_raw_within_fixed_crop}. More importantly, for all experiments with significant positive transfer gains, we find that the best source remains the best source. The only exception is for SUN RBG-D as target: the best three sources are the same but have a different order. All these sources yield high transfer gains. We conclude that our analysis (\mainpaperref{sec:analysis}) holds when fixing the image resolution.

\para{ResNet50.} We change the backbone to ResNet50~\cite{he16cvpr} and redo the experiments in the small target training setting for segmentation (\ie redo~\mainpaperref{tab:few_shot_semseg_within}) and partially for detection (\ie redo~\mainpaperref{tab:few_shot_detection}). Results are shown in~\autoref{tab:tt_segmentation_raw_within_resnet_full}.

For segmentation, on average results are significantly worse (-0.11 IoU) compared to the HRNetV2 backbone used in our main experiments. But again, most transfer patterns stay the same. Quantitatively, for 81\% of the experiments with significant positive or negative gains with HRNetV2 backbone (\mainpaperref{tab:few_shot_semseg_within}), similar gains are observed for ResNet50. In fact, gains are stronger for 69\% of all experiments, both in positive and negative direction. There are some difference though: 
generally for ResNet50 there are more source-target combinations with significant positive transfer gains. This is for iSAID as a target dataset, vKITTI2 as a source for \emph{driving}, and for \emph{consumer} as a source for \emph{consumer, driving}, and \emph{indoor}. Measured quantitatively, in 13\% of the experiments where we before saw negative transfer gains, we now see positive transfer gains. This suggests that ResNet50 benefits more from transfer learning. We hypothesize this is because its activations are at a lower resolution, and training on a segmentation source dataset makes the network better suited for localized predictions.

For detection (~\autoref{tab:tt_detection_raw_resnet}) there are a few more changes when using ResNet50. Within detection, COCO remains a good source, VOC07 becomes less good (but not negative), while BDD now yields positive transfer instead of negative for VOC07 and Underwater Trash as a target. When using segmentation as a source, all but one sources which had positive transfer gains using HRNetV2 still yield positive transfer. More interestingly, the number of source-target combinations which yield positive transfer gains are doubled (from 6 to 12). Again, this suggests training on segmentation makes the ResNet50 model better suited for localized predictions.

To conclude, ResNet50 models perform less than HRNetV2 in terms of the absolute performance, however ResNet50 generally benefits more from transfer learning, possibly due to improving its ability for localized predictions. At the same time, the overall trends which we observed in~\mainpaperref{sec:analysis} still hold.

\para{Self-supervised ILSVRC'12 training.} We now redo the experiments with ResNet50, but instead we start from a checkpoint which was obtained by using self-supervised learning on ILSVRC'12 using the publicly available SimCLR V2 implementation~\cite{chen20icml}. Starting from these weights, we train the sources fully supervised as before. Results are shown in~\autoref{tab:tt_segmentation_raw_within_resnet_ss}.

We compare results with our other ResNet50 experiments (\ie~\mainpaperref{tab:tt_segmentation_raw_within_resnet_full}). We see that the best or top 3 sources remain the same for all experiments with positive transfer gains, while most previously observed patterns hold. However, there are noticeably fewer positive gains overall, some even turning into negative gains. This is especially visible for \emph{consumer} as source and \emph{driving} as target. But a closer look reveals that the most significant changes happens on the ILSVRC'12 baseline: results starting from self-supervised weights are on average 0.01 IoU higher than starting from fully supervised classification weights, while results for the full transfer chains are comparable. This manifests itself as lower measured transfer gains. The only exception is iSAID, where the self-supervised ILSVRC'12 baseline is 0.03 IoU lower, while results for the full transfer chain are comparable. Hence here we measure higher transfer gains.

To conclude, the main observation is that the self-supervised ILSVRC'12 pre-training generally yields slightly better segmentation results than fully supervised classification pre-training, which is in line with~\cite{ericsson21cvpr,zoph20nips}. Still, overall patterns remain the same and the overall conclusions in~\mainpaperref{sec:analysis} remain unaltered.

\para{Self-supervised Transfer Chain.}
We do a single experiment where we use SimCLR~\cite{chen20icml} to create a self-supervised transfer chain: ILSVRC'12 self-supervised $\rightarrow$ COCO self-supervised $\rightarrow$ target. We first verify whether the self-supervision works for COCO image classification as a target following a standard self-supervised evaluation protocol~\cite{chen20icml,chen21cvpr,grill20nips}: we take the trained backbone, freeze its weights, attach a linear classifier, and train on COCO image classification. We do this for both ILSVRC'12 weights and ILSVRC'12 $\rightarrow$ COCO weights. We find that the additional self-supervised training on COCO leads to a small improvement in classification accuracy of 0.4\%.

We now test the transfer chain on segmentation for the small target training setting. Results are shown in~\autoref{tab:tt_segmentation_raw_within_resnet_ss_chain}. 
Interestingly, we find that this chain is worse than directly training from ILSVRC'12 self-supervised weights for \emph{all} target datasets. On average it is 0.07 IoU worse, while even for COCO as a target results are 0.02 IoU worse.

This result suggests that self-supervised pre-training using SimCLR~\cite{chen20icml} is biased towards image classification. Furthermore, image classification results on self-supervised models are not very predictive of performance on other tasks, as also shown in the dedicated study of~\cite{ericsson21cvpr}.

\section{Complete Result Tables}\label{sec:appendix_results}

\changed{
This~\thisappendix reports more extensive results for all of our experiments, \ie including both absolute performance of the task type specific metric as well as relative transfer gain calculated \wrt the ILSVRC'12 column of the absolute performance, \cf \mainpaperref{eq:relative_transfer_gain}.
The experiments and metrics are summarized as follows:
}

\begin{itemize}
\item For semantic segmentation we use Intersection-over-Union (IoU), averaged over classes~\cite{everingham15ijcv}. See \autoref{tab:tt_segmentation_raw_within} and \autoref{tab:tt_segmentation_raw_cross}.
\item For semantic segmentation we also include experiments using ResNet50 as backone, both when pre-trained on the ILSVRC'12 image classification task, and when self-supervisedly trained on the ILSVRC'12 training set. The results are in~\autoref{tab:tt_segmentation_raw_within_resnet}. For self-supervised training we follow the SimCLR~\cite{chen20nips, chen20icml} approach. For both ResNet models we use the publicly available checkpoints from the SimCLR \href{https://github.com/google-research/simclr}{repo}.
\item For semantic segmentation we also include experiments using the HRNetV2 backbone, but with a fixed image cropping instead of the dataset specific cropping used in the main experiments (\cf \mainpaperref{sec:data_normalization}).
The results are in~\autoref{tab:tt_segmentation_raw_within_fixed_crop}.
\item For object detection, we use mean Average Precision (mAP) at a threshold of 0.5 (box-based) IoU~\cite{everingham15ijcv} for Pascal VOC and Underwater Trash. For COCO and BDD we use the COCO mAP variant~\cite{lin14eccv}, which evaluates mAP at multiple IoU thresholds and averages them.
As earlier observed in~\cite{zhou19arxiv}, the common post-processing stage of Non-Maximum Suppression (NMS) did not improve results on COCO or BDD, but we found it beneficial on Pascal VOC and SUIM. For the latter two dataset we use NMS at an IoU of 0.3.
Results are in \autoref{tab:tt_detection_raw}.
\item For object detection we have also used ResNet50 as a backbone, using pre-training on ILSVRC'12 image classification. The results are in~\autoref{tab:tt_detection_raw_resnet}.
\item For keypoint detection we report the mean Average Precision at 0.5 Object Keypoint Similarity (OKS), as defined by the COCO challenge~\cite{coco-challenge}. 
Results are in \autoref{tab:tt_keypoints_raw}.
\item For depth estimation we use the Root Mean Square Error (RMSE), linear version, as defined in~\cite{eigen14nips} (lower is better), and also the standard metric $\delta < 1.25$, where $\delta = \max(\tfrac{\hat{z}}{z},\tfrac{z}{\hat{z}})$ is a measure of relative accuracy as defined in~\cite{ladicky14cpr} (higher is better).
Results are in \autoref{tab:tt_depth_raw}.
\end{itemize}

\noindent%

\begin{table*}[p]
    \centering
    \begin{subtable}[t]{\textwidth}
        \centering
        \includegraphics[height=74mm]{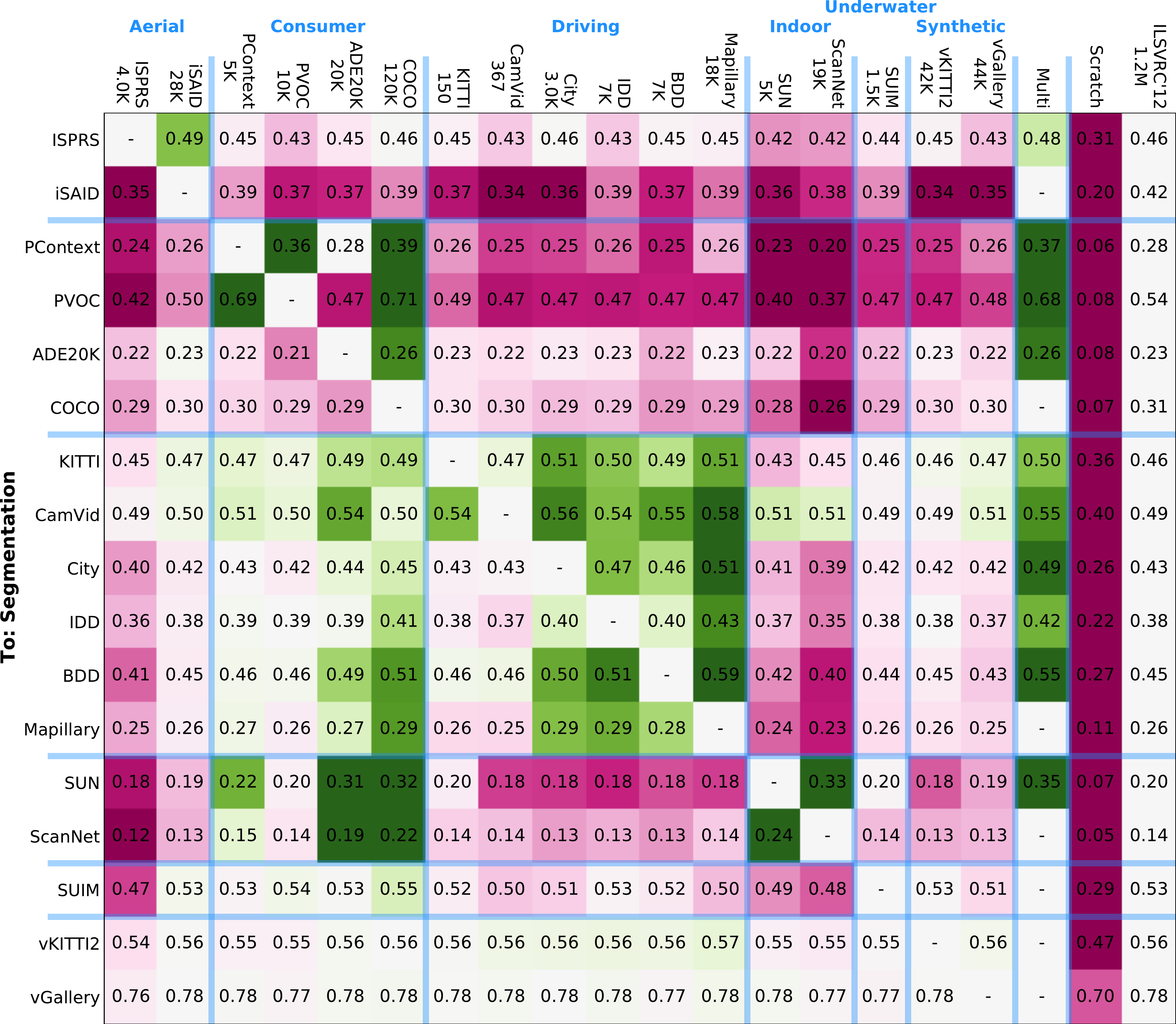}\quad 
        \includegraphics[height=74mm]{figures/raw_tables/semseg_within_Bstar_isprs_relative}
        \caption{Small target training set.}
        \label{tab:segmentation_within_b}
    \end{subtable}
    
    \begin{subtable}[t]{\textwidth}
        \centering
        \includegraphics[height=74mm]{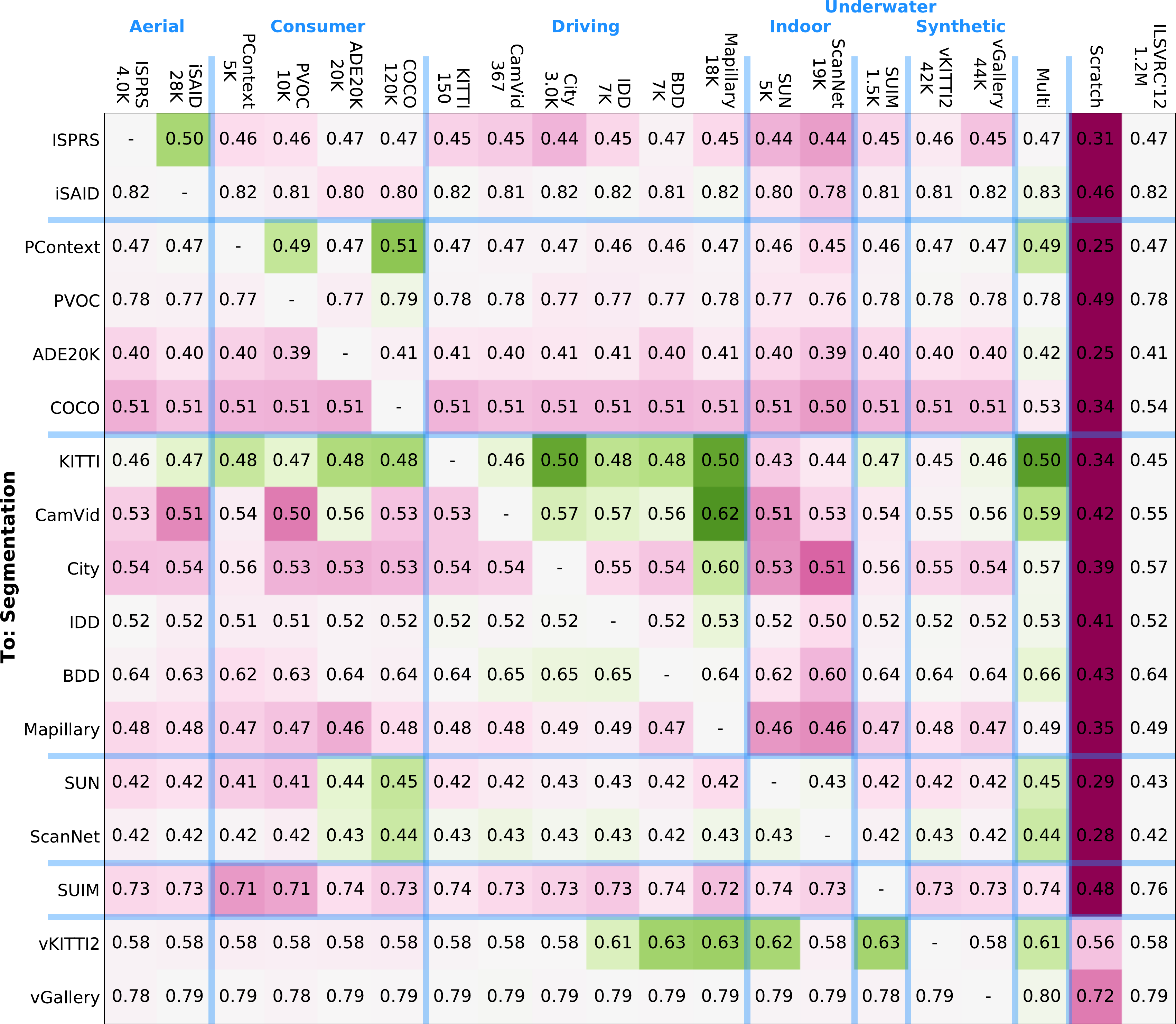}\quad 
        \includegraphics[height=74mm]{figures/raw_tables/semseg_within_A1_isprs_relative}
        \caption{Full target training set.}
        \label{tab:segmentation_a}
    \end{subtable}
    
    \begin{subtable}[t]{\textwidth}
        \centering
        \includegraphics[height=74mm]{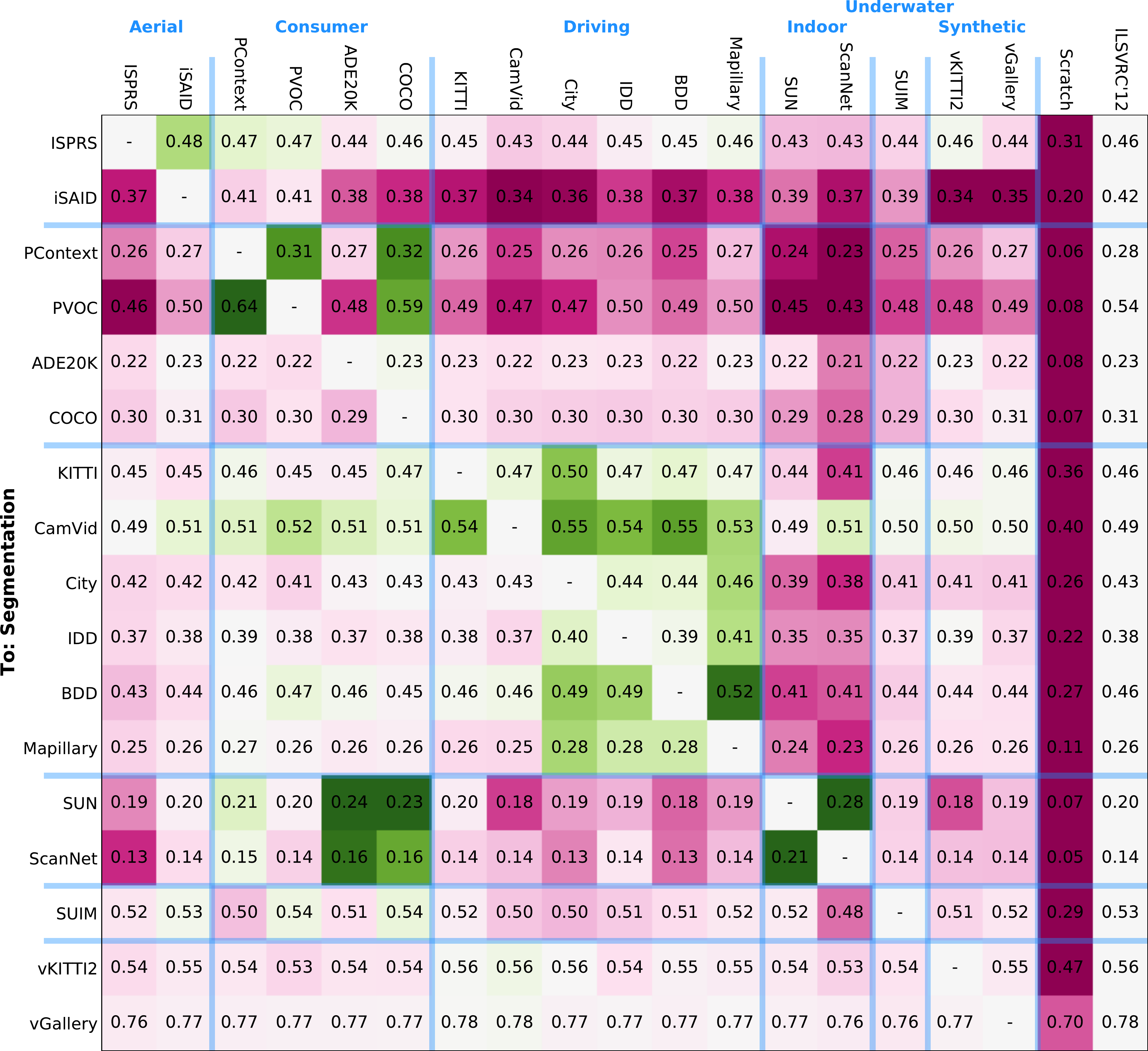}\quad 
        \includegraphics[height=74mm]{figures/raw_tables/semseg_within_Dstar_isprs_relative}
        \caption{Small source and small target training set.}
        \label{tab:segmentation_within_d}
    \end{subtable}

    \caption{Absolute performance and the corresponding relative transfer gains for semantic segmentation as both a source and target task type. Left tables: mean Intersection-over-Union. Right tables: relative transfer gains.}
    \label{tab:tt_segmentation_raw_within}
\end{table*}

\begin{table*}[pt]
    \centering
    \begin{subtable}[t]{\textwidth}
        \centering
        \includegraphics[height=60mm]{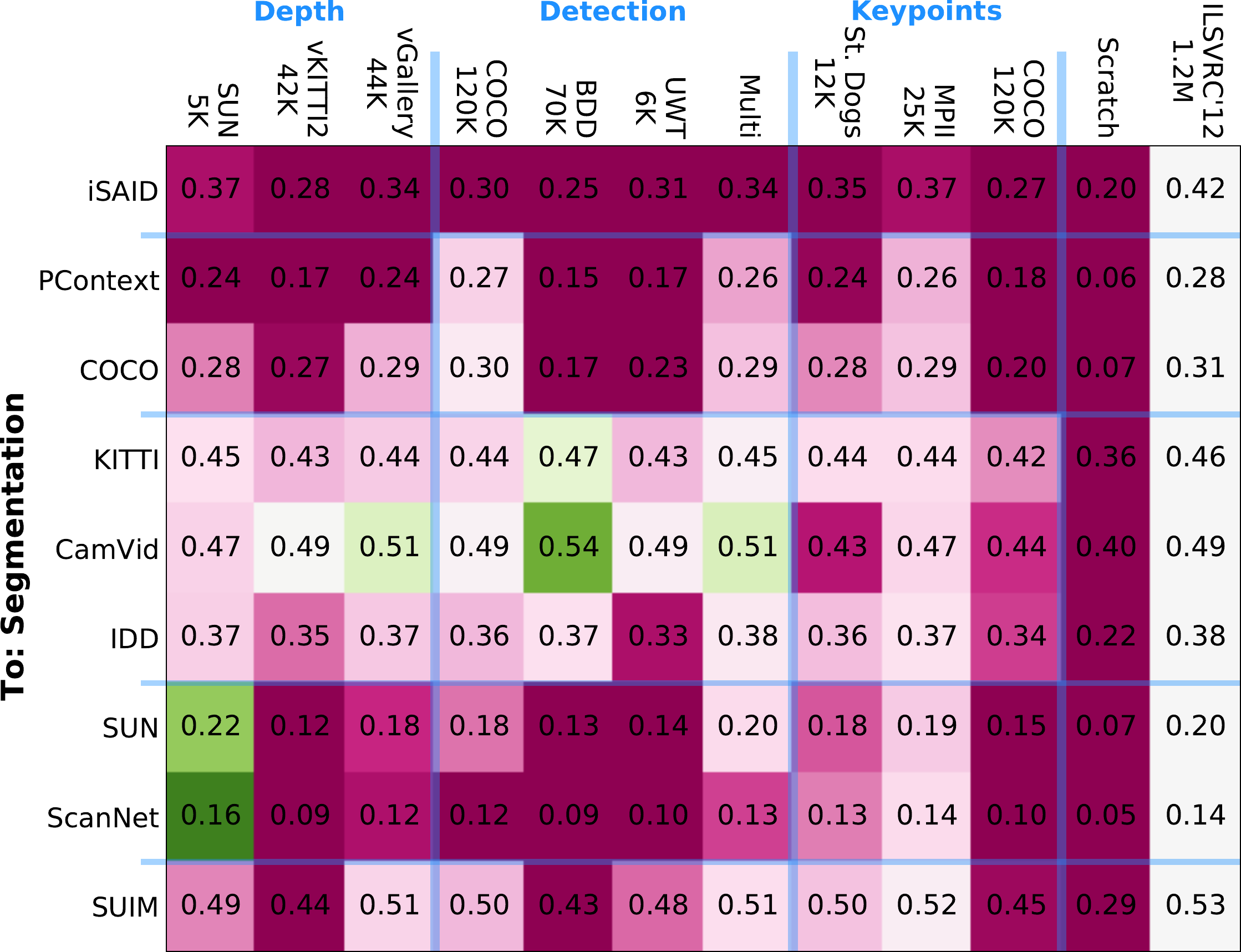}\quad 
        \includegraphics[height=60mm]{figures/raw_tables/semseg_cross_Bstar_relative}
    \end{subtable}
    \caption{Transfer from other task types to semantic segmentation, in the small target training setting. Left: absolute metric (Intersection-over-Union)  and the corresponding relative transfer gains (right).
    }
    \label{tab:tt_segmentation_raw_cross}
\end{table*}

\begin{table*}[pt]
    \centering
    \begin{subtable}[t]{.68\textwidth}
        \includegraphics[height=32mm]{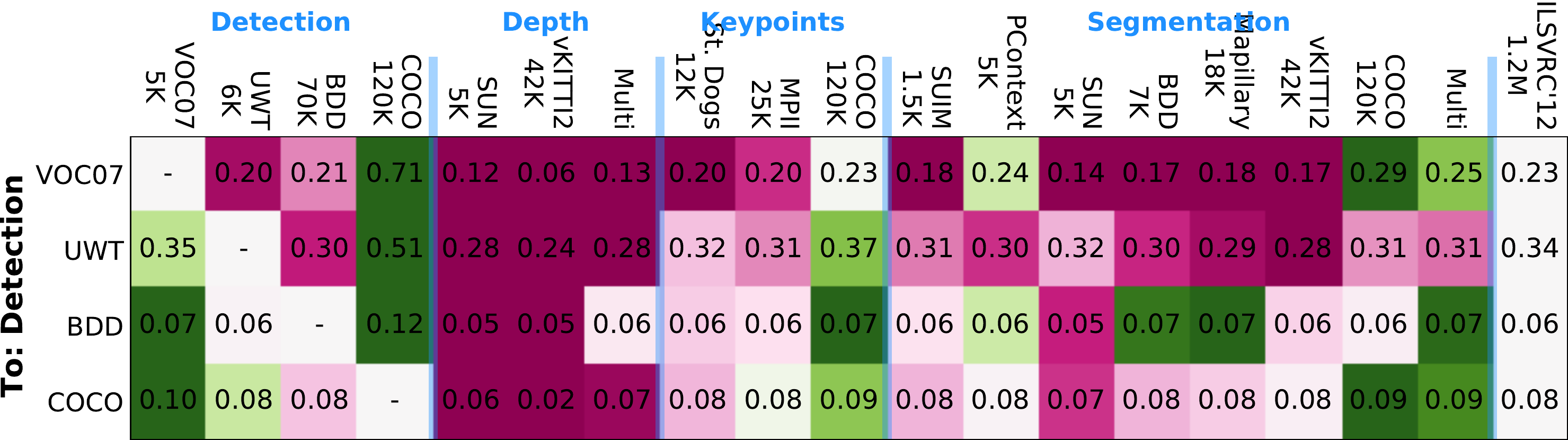}\\[5mm]
        \includegraphics[height=32mm]{figures/raw_tables/detection_all_Bstar_hrnet48_relative}
        \caption{Small target training set. Top: mean Average Precision. Bottom: relative transfer gain.}
        \label{tab:detection_all_b}
    \end{subtable}
    \begin{subtable}[t]{.68\textwidth}
        \vspace{1cm}
        \includegraphics[height=33mm]{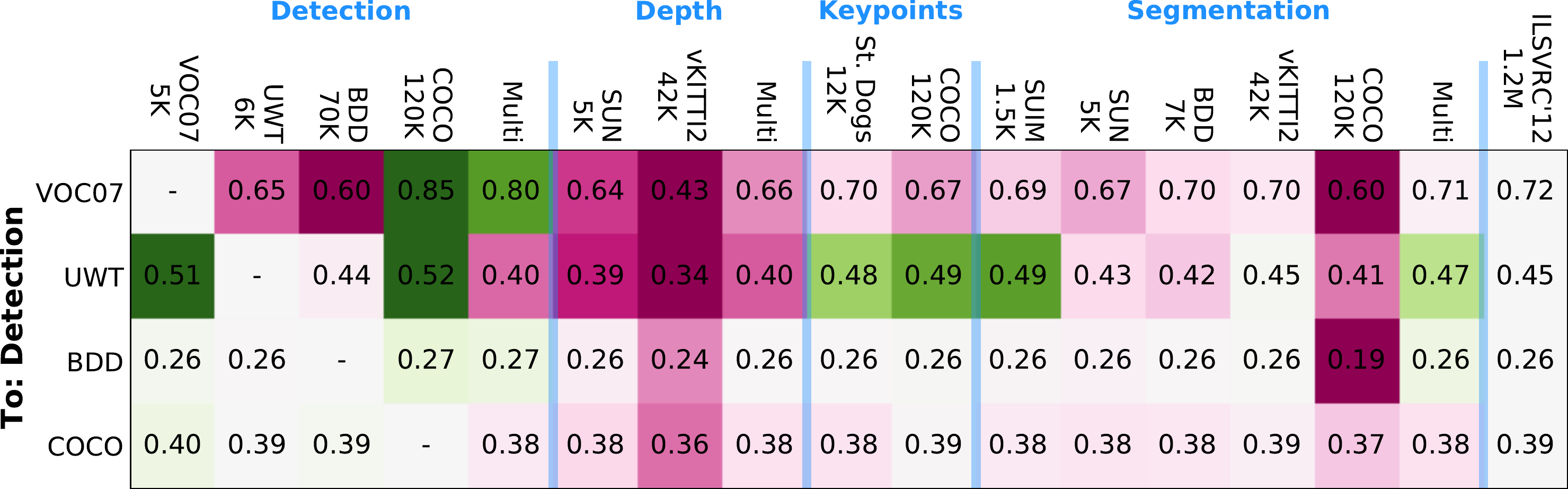}\\[5mm]
        \includegraphics[height=33mm]{figures/raw_tables/detection_all_A_relative}
        \caption{Full target training set. Top: mean Average Precision. Bottom: relative transfer gain.}
        \label{tab:detection_a}
    \end{subtable}
    \caption{Absolute metrics and the corresponding relative transfer gains for object detection as a target task type.
    }
    \label{tab:tt_detection_raw}
\end{table*}

\begin{table*}[tp]
    \centering
    \begin{subtable}[t]{.48\textwidth}
        \includegraphics[height=34mm]{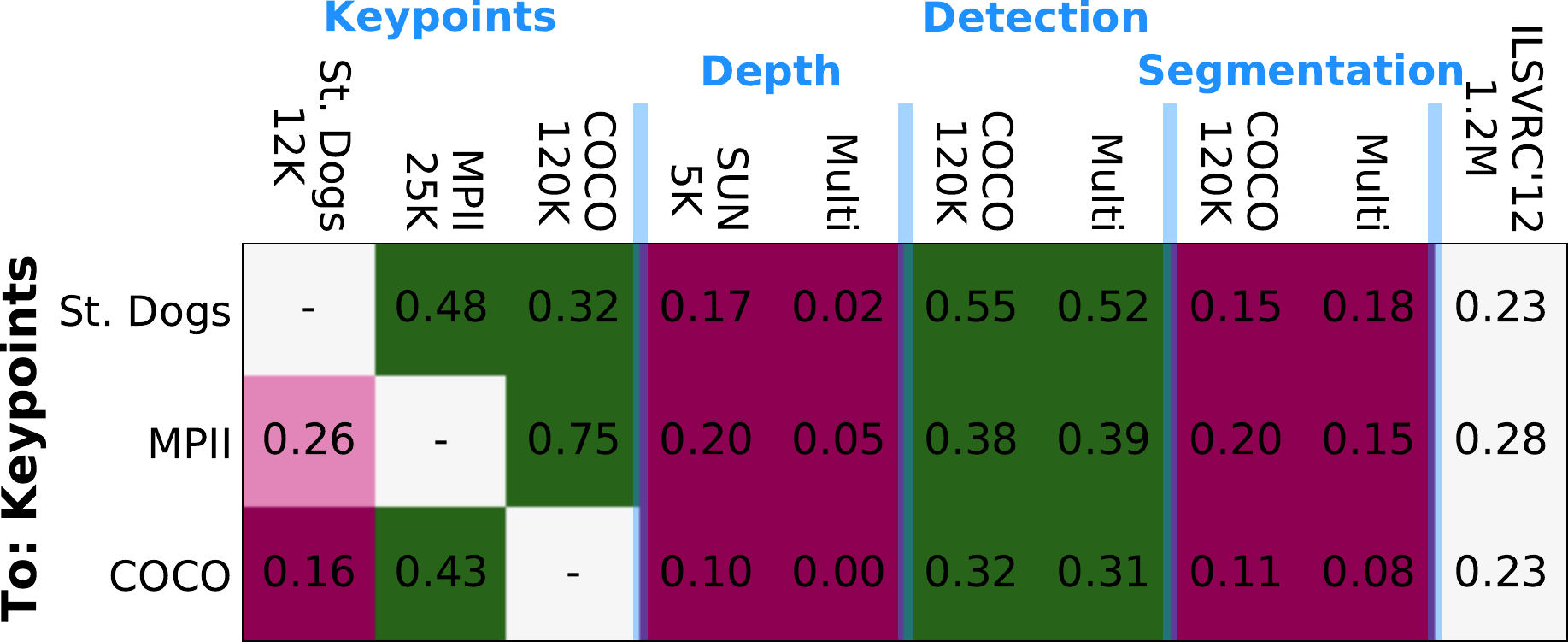}\\[5mm]
        \includegraphics[height=34mm]{figures/raw_tables/keypoint_all_b1_relative}
        \caption{Small target training set. Top: mean Average Precision at 0.5 Object Keypoint Similarity. Bottom: relative transfer gain.}
        \label{tab:keypoints_b}
    \end{subtable}
    \hspace{5mm}
    \begin{subtable}[t]{.48\textwidth}
        \includegraphics[height=34mm]{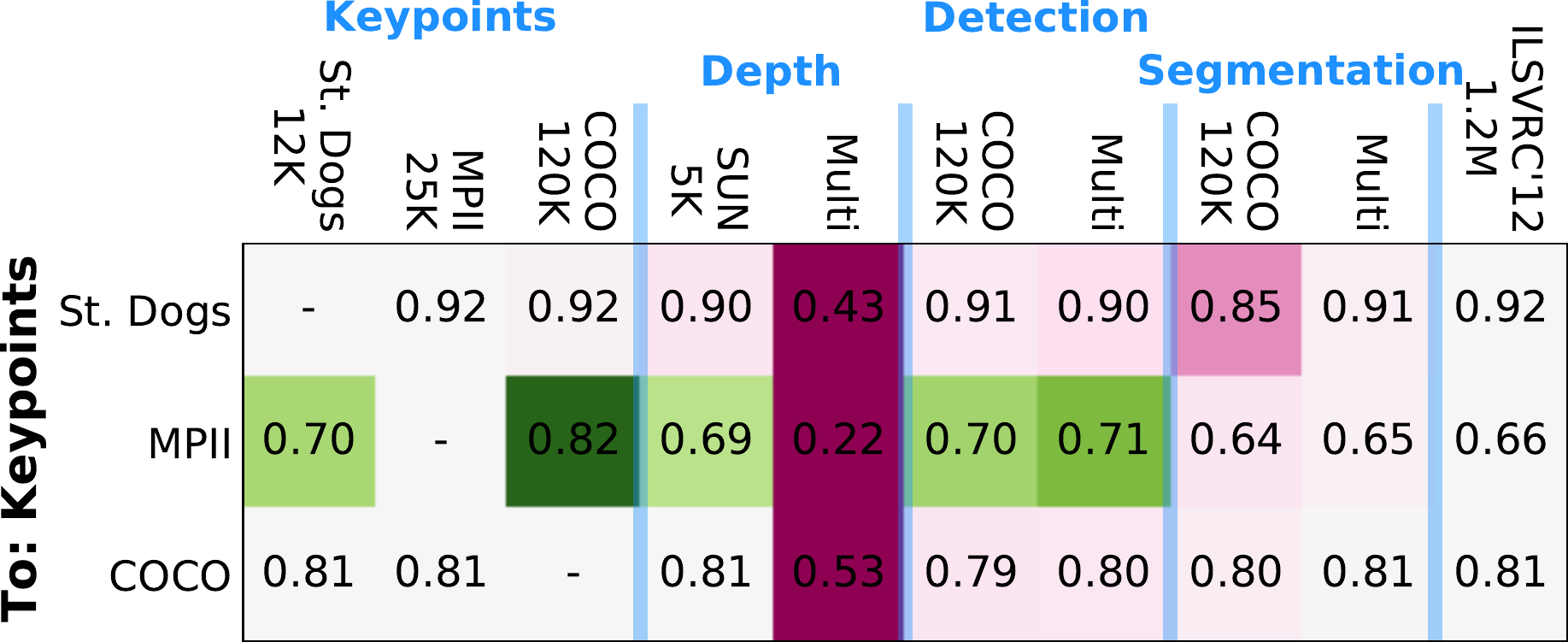}\\[5mm]
        \includegraphics[height=34mm]{figures/raw_tables/keypoint_all_a1_relative}
        \caption{Full target training set. Top: mean Average Precision at 0.5 Object Keypoint Similarity. Bottom: relative transfer gain.}
        \label{tab:keypoints_a}
    \end{subtable}
    \caption{Absolute metrics and the corresponding relative transfer gains for keypoint detection as target task type.}
    \label{tab:tt_keypoints_raw}
\end{table*}

\newcommand{\depthtableheight}{30mm}
\begin{table*}[tp]
    \begin{subtable}[t]{.64\textwidth}
        \includegraphics[height=\depthtableheight]{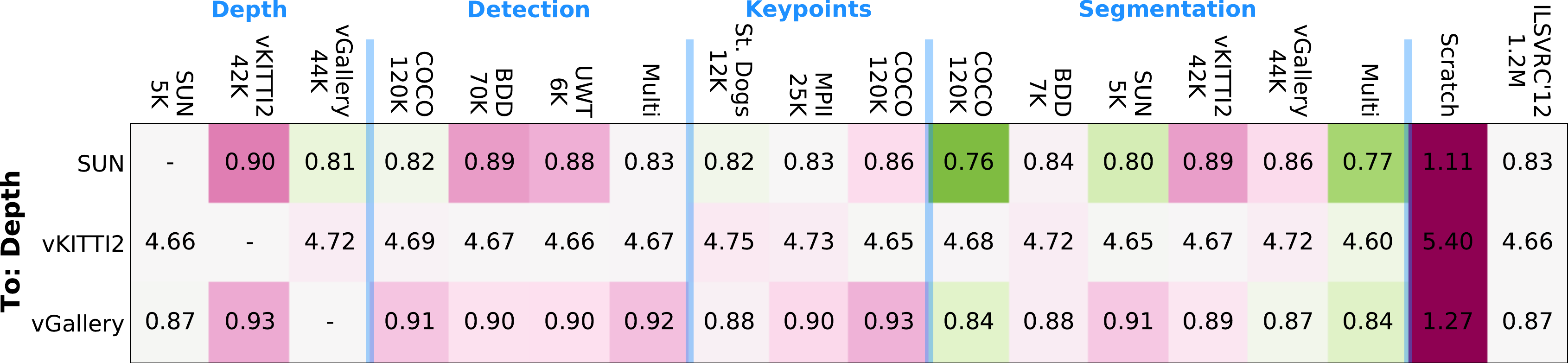}\\
        \includegraphics[height=\depthtableheight]{figures/raw_tables/depth_b_rmse_relative}
        \caption{Small target training set. Top: RMSE. Bottom: relative transfer gain.}
    \end{subtable}
    \hspace{1.5cm}
    \begin{subtable}[t]{.30\textwidth}
        \includegraphics[height=\depthtableheight]{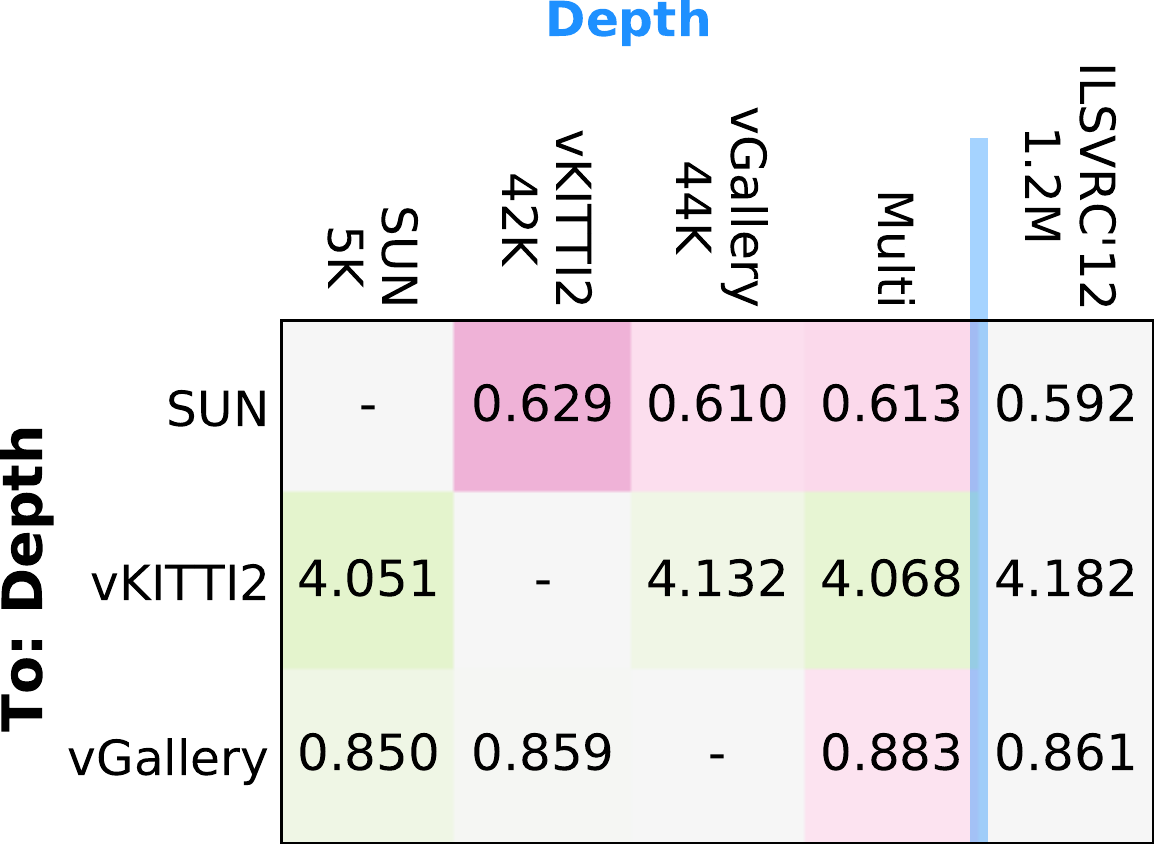}\\
        \includegraphics[height=\depthtableheight]{figures/raw_tables/depth_a_rmse_relative}
        \caption{Full target training set. Top: RMSE. Bottom: relative transfer gain.}
    \end{subtable}
    
    \begin{subtable}[t]{.64\textwidth}
        \includegraphics[height=\depthtableheight]{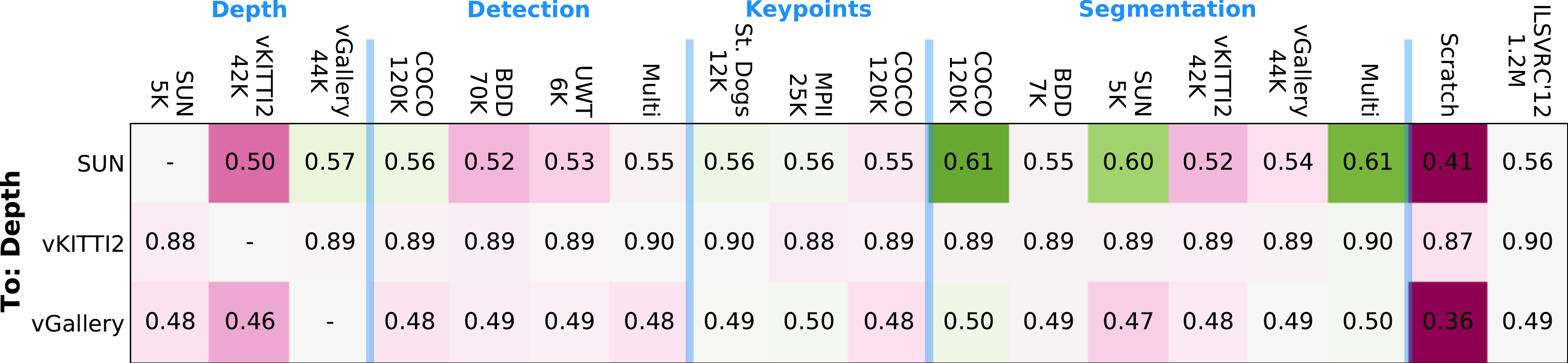}\\
        \includegraphics[height=\depthtableheight]{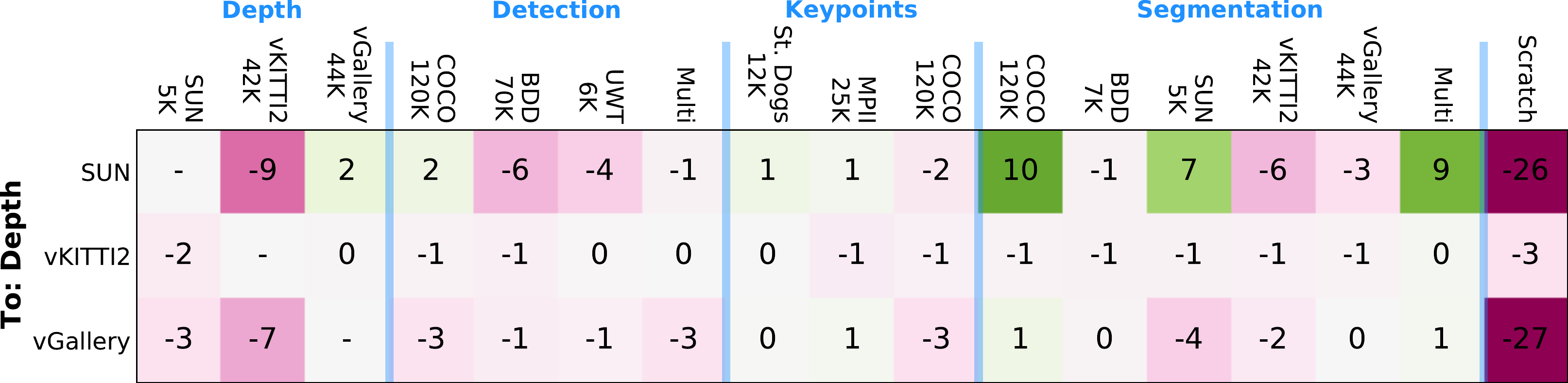}
        \caption{Small target training set. Top: $\delta < 1.25$. Bottom: relative transfer gain.}
    \end{subtable}        
    \hspace{1.2cm}
    \begin{subtable}[t]{.32\textwidth}
        \includegraphics[height=\depthtableheight]{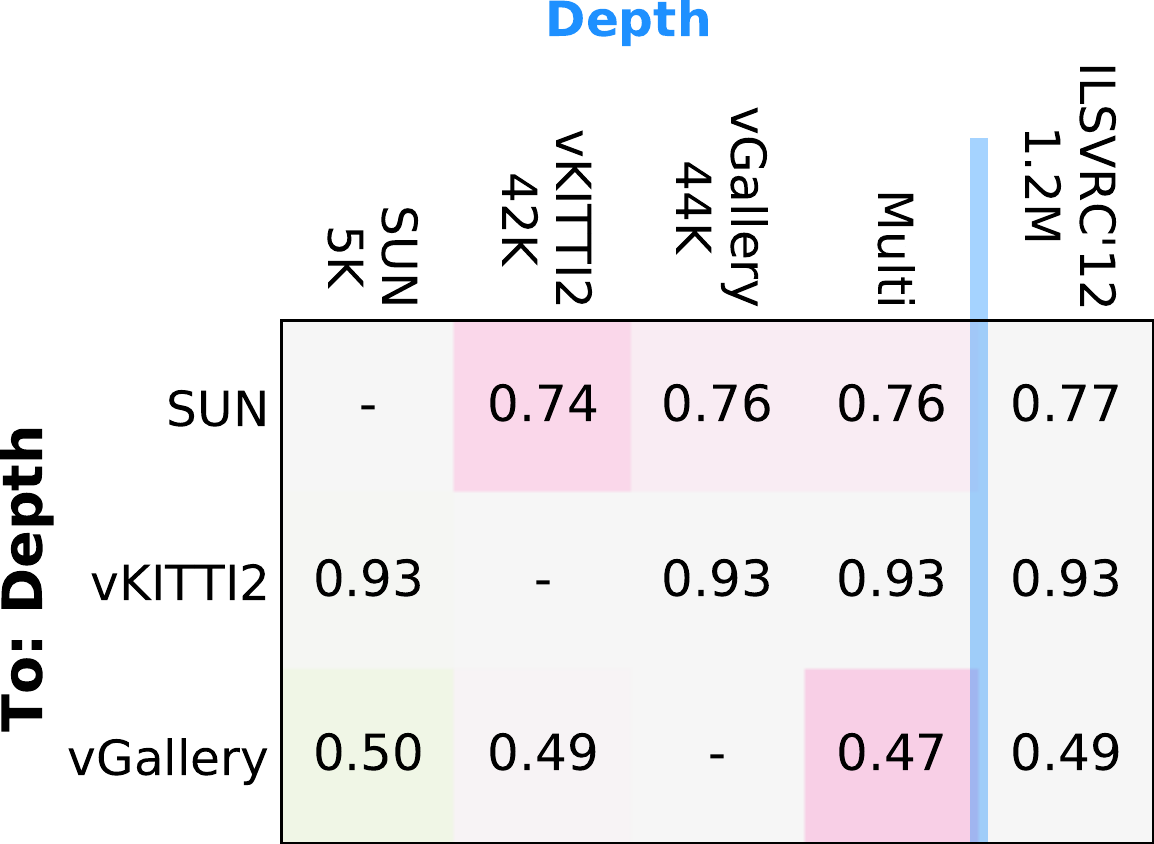}\\
        \includegraphics[height=\depthtableheight]{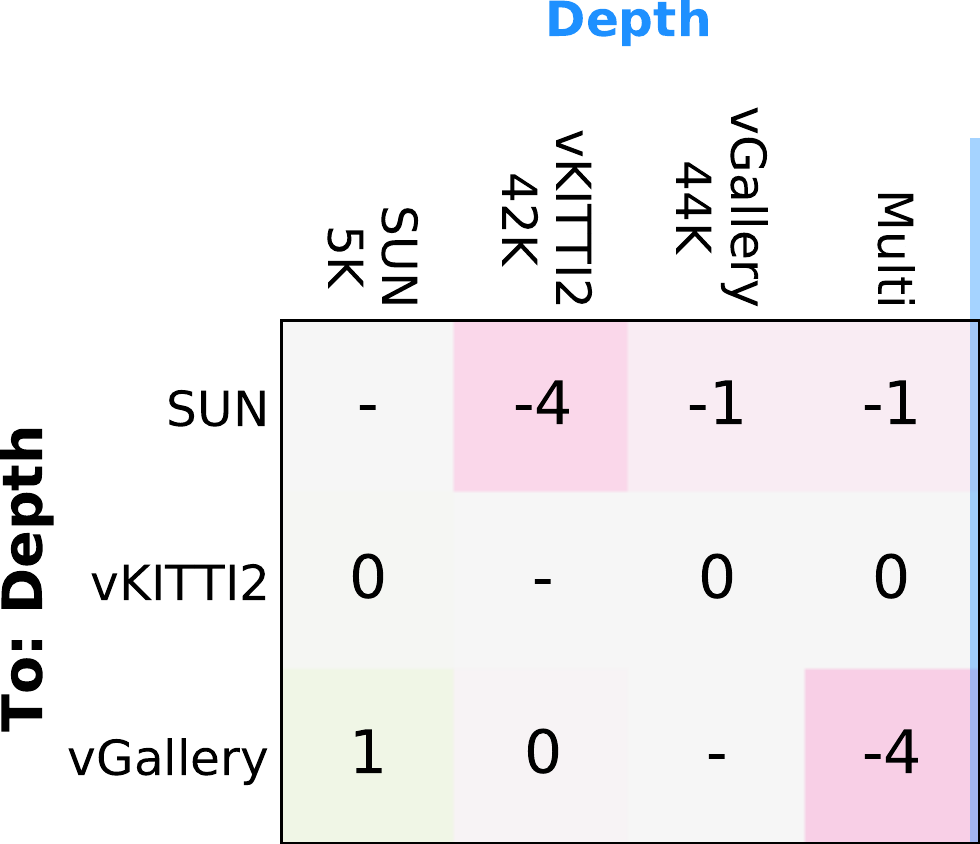}
        \caption{Full target training set. Top: $\delta < 1.25$. Bottom: relative transfer gain.}
    \end{subtable}
    \caption{Absolute metrics and the corresponding relative transfer gains for depth estimation as a target task type.}
    \label{tab:tt_depth_raw}
\end{table*}

\iftoggle{includesuppmat}{\newpage}{}
\section{Computational costs of transfer chains}
\label{sec:computational_cost}
For performed the experiments in this paper on Google Cloud TPU-v3 chips. If we calculate costs in terms of TPU-hours, defined as the number of TPUs multiplied with the computation time:

\begin{itemize}
    \item ILSVRC’12 pretraining (ImageNet) takes 364 hours. 
    \item Training a COCO segmentation source model (largest \emph{consumer} source) takes an additional 92 hours. 
    \item Training a Mapillary segmentation source model (\ie the largest \emph{driving} source) takes an additional 352 hours (due to higher resolution images). 
    \item Training a multi-source model takes 228 hours.
\end{itemize}

This means that pre-training costs for our transfer chains are increased by 25\%-97\%, compared to the standard practice of pre-training on ILSVRC’12. 

However, the source models are trained only once. The same source models can then be repeatedly used for transfer learning to many target tasks. This means that the relative additional costs of pre-training becomes increasingly small as more researchers reuse the same source model for new target tasks.

\newpage

\section{Dataset Overlap}
\label{sec:data_overlap}
When using 21 datasets there is always a risk that images in the test set of one dataset are used as training in another dataset, this is deemed undesirable for fair evaluation. 
Since we have not quantified such an overlap, we can not assure there is no overlap at all. 
Still, we do believe that the risk of such potential overlap significantly changes the results of our experiments to be minimal, for the following reasons:
\begin{enumerate}
    \item We evaluate structured prediction tasks, while the largest dataset (ILSVRC'12) had only classification annotation, thus even if an image has already been seen it had different type of annotation.
    \item Within the same annotation type datasets either differ in visual domain (\eg underwater and aerial are not expected to have any overlap), or they are acquired in different geographical regions (\eg BDD and IDD). 
    \item Since the collection consists of popular public benchmarks, major overlaps of these datasets would have been reported.
\end{enumerate}

}

\end{document}